\documentclass{article}

\PassOptionsToPackage{numbers, compress}{natbib}

\usepackage[main, final]{neurips_2026}

\usepackage[utf8]{inputenc} 
\usepackage[T1]{fontenc}    
\usepackage{hyperref}       
\usepackage{url}            
\usepackage{booktabs}       
\usepackage{amsfonts}       
\usepackage{nicefrac}       
\usepackage{microtype}      
\usepackage{xcolor}         

\usepackage{wrapfig}

\usepackage{amsthm}

\newtheorem{theorem}{Theorem}

\newtheorem{definition}{Definition}

\usepackage{booktabs}
\usepackage{subfigure}
\usepackage{multirow}
\usepackage{pifont}
\usepackage{bbding}
\usepackage{adjustbox}
\usepackage{threeparttable}
\usepackage{colortbl}
\usepackage{enumitem}

\definecolor{darkblue}{RGB}{0,51,102}
\hypersetup{
    colorlinks=true,
    linkcolor=darkblue,
    citecolor=darkblue,
    urlcolor=darkblue
}

\definecolor{COLOR_MEAN}{HTML}{f0f0f0}

\newcommand{\vpara}[1]{\vspace{0.05in}\noindent\textbf{#1 }}

\usepackage[edges]{forest}
\definecolor{level0}{HTML}{2C3E50} 
\definecolor{level1}{HTML}{2980B9} 
\definecolor{level2}{HTML}{16A085} 
\definecolor{level3}{HTML}{8E44AD} 

\usepackage{bbm}

\usepackage[most]{tcolorbox}

\title{Uncovering Semantic Hierarchies in Text-Attributed Graphs via Variational EM-based LLM-GNN Synergy}

\author{%
  Yunhui Liu$^{1}$, Xudong Jin$^{1}$, Qizhuo Xie$^{1}$, Chunhui Zhao$^{1}$ \\
  \bfseries Xiao Luo$^{2}$, Tao Zheng$^{1}$, Bin Chong$^{3}$, Tieke He$^{1}$\thanks{Corresponding author.} \\
  \mdseries\footnotesize $^{1}$State Key Laboratory for Novel Software Technology, Nanjing University \\
  \mdseries\footnotesize $^{2}$Department of Statistics, University of Wisconsin--Madison \\
  \mdseries\footnotesize $^{3}$National Engineering Laboratory for Big Data Analysis and Applications, Peking University \\
  \mdseries\footnotesize \texttt{lyhcloudy1225@gmail.com}, \texttt{xiao.luo@wisc.edu}, \texttt{chongbin@pku.edu.cn}, \texttt{hetieke@gmail.com}
}

\begin{document}

\maketitle

\begin{abstract}
While hierarchical structures are intrinsic to organizing human knowledge, current representation learning on text-attributed graphs predominantly operates on flat semantic spaces, overlooking the rich coarse-to-fine granularity inherent in real-world data. 
To bridge this gap, we propose \texttt{SHiFT}, which synergizes the reasoning power of LLMs with the structural representation capability of GNNs within an Expectation-Maximization (EM) paradigm. 
In the E-step, we harness LLMs as expert taxonomists to induce an explicit semantic tree, introducing a lightweight Mapping \& Diagnosis strategy that reduces inference costs by surgically updating the hierarchy only when anomalies are detected. 
In the M-step, we align the GNN encoder with this induced hierarchy from complementary perspectives, including clustering partition, topological skeleton, and semantic concepts. 
Theoretical analysis interprets \texttt{SHiFT} as a generalized Variational EM algorithm. 
Extensive experiments demonstrate that \texttt{SHiFT} not only achieves state-of-the-art performance but also uncovers high-quality, human-readable taxonomies.
Our code can be found at \url{https://github.com/Cloudy1225/SHiFT}.
\end{abstract}

\section{Introduction}\label{sec:intro}
Structuring knowledge into hierarchies is fundamental to human cognition~\cite{tenenbaum2011grow} and modern AI\begin{wrapfigure}{r}{0.42\textwidth}
    \centering
    \vspace{-0.8em}
    \includegraphics[width=0.40\textwidth]{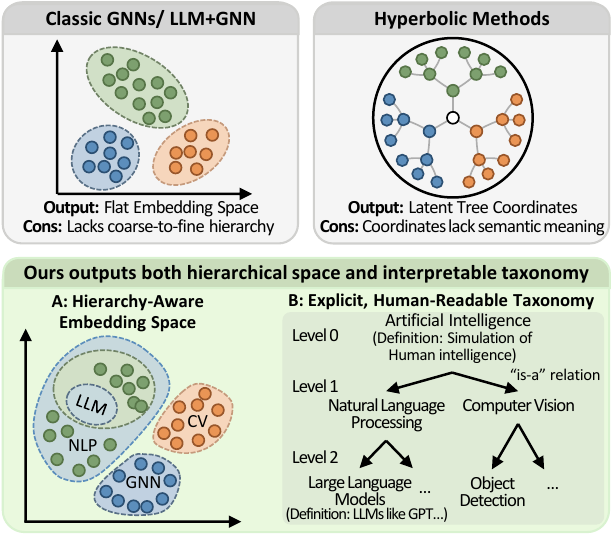}
    \caption{Existing paradigms vs. \texttt{SHiFT}.}
    \label{fig:motivation}
    \vspace{-1.2em}
\end{wrapfigure} systems, underpinning biomedical ontologies~\cite{BioOnto}, document taxonomies~\cite{DocTaxo}, and recent advances in retrieval-augmented generation~\cite{HiRAG} and agent memory~\cite{MemTree}.
Yet hierarchical modeling remains underexplored for Text-Attributed Graphs (TAGs), graphs where nodes carry rich textual content and edges encode semantic relations, which are pervasive in scientific literature, e-commerce, and social media~\cite{LLM4Graph}.
In these domains, data exhibits natural hierarchical semantics: for instance, in a citation network, \textit{Large Language Models} and \textit{Graph Machine Learning} are distinct sub-fields of \textit{Artificial Intelligence}, residing on different branches from \textit{Computer Vision}~\cite{SciHTC}.
An ideal embedding space should reflect this geometry, encoding node similarity across levels of semantic abstraction.

However, prevailing TAG methods, ranging from classic Graph Neural Networks (GNNs)~\cite{GCN} to recent interactions with Large Language Models (LLMs)~\cite{LLM4Graph}, predominantly operate on flat semantic structures, as shown in Figure~\ref{fig:motivation}. 
While they excel at capturing the graph topology, they often fail to disentangle the semantic hierarchy inherent in the node documents. 
Existing attempts to introduce hierarchy into graph learning, such as hyperbolic embeddings~\cite{HGCN, HyperGCL}, typically focus on geometric properties. 
These methods require complex optimization and, crucially, yield latent hierarchies that lack semantic interpretability, e.g., they provide coordinates, not concepts. 
Consequently, a critical question arises: \emph{How can we effectively uncover and model the explicit semantic hierarchy in TAGs when such taxonomies are absent or incomplete?}

Addressing this question is non-trivial due to the duality of the data: the hierarchy is partially encoded in textual semantics and partially implied by the graph topology, like community structures. 
Aligning these modalities presents a dilemma. 
While LLMs possess vast internalized knowledge to recognize semantic hierarchies~\cite{TaxoAdapt, TaxoBench}, they lack the ability to perceive the graph structures that reveal sub-fields. 
Conversely, GNNs excel at structural clustering~\cite{SASE, PyAGC, AGCSurvey} but lack the semantic capability to name and organize these clusters into a meaningful taxonomy. 
Furthermore, naive solutions, such as prompting an LLM to classify every node independently~\cite{LLMNodeBed}, are not only prohibitively expensive but also disregard the relational signals provided by the graph structure.

To bridge this gap, we propose \texttt{SHiFT} (\texttt{S}emantic \texttt{H}ierarchy-\texttt{i}nduced \texttt{F}ramework for \texttt{T}ext-attributed graphs), which synergizes the semantic reasoning of LLMs with the structural representation power of GNNs within an Expectation-Maximization (EM) style framework. 
In the \textbf{E-step}, we harness LLMs as expert taxonomists: hierarchical clustering on the embedding space selects representative coreset documents, from which the LLM induces a tree-structured taxonomy with explicit concept names and definitions; to mitigate inference cost, a lightweight \textit{Mapping \& Diagnosis} strategy invokes the LLM only upon detecting anomalies such as concept drift or low cluster cohesion. 
In the \textbf{M-step}, we optimize the GNN encoder via a unified objective that internalizes three facets of the induced taxonomy: \emph{Taxonomy Partition Alignment}, enforcing local cohesion by grouping nodes per cluster assignment; \emph{Taxonomy Skeleton Alignment}, imposing a global geometric constraint so that embedding distances mirror path distances in the hierarchy; and \emph{Taxonomy Concept Alignment}, anchoring node embeddings to LLM-generated concepts across multiple abstraction levels for semantic grounding.
This iterative process lets graph topology guide the LLM toward topologically consistent concepts, while the LLM guides the GNN toward a semantically organized space.

We further provide a formal theoretical analysis showing \texttt{SHiFT} instantiates a generalized Variational EM algorithm, offering a principled account of its convergence.
Experiments show that \texttt{SHiFT} not only achieves state-of-the-art performance in classification and clustering tasks but also produces high-quality, human-readable taxonomies that reveal the latent organization of the data.

\section{Related Work}\label{sec:related_work}

\subsection{Hierarchical Structure Learning}
Capturing hierarchical structures in data has demonstrated broad applicability across domains such as document classification~\cite{SciHTC, TELEClass}, knowledge graph completion~\cite{GS-Quant}, retrieval-augmented generation~\cite{RAPTOR, HiRAG}, and agent memory systems~\cite{MemTree}.
One line of research relies on hierarchical contrastive learning~\cite{HGCLR, HMCL} or hyperbolic embeddings~\cite{HyboNet, HiTs}.
However, purely embedding methods lack explicit semantic interpretability (e.g., dimensions in the hyperbolic Poincaré ball do not directly correspond to human-readable concepts).
Another line of research involves taxonomy induction~\cite{TaxoGen, TaxoBench} and hierarchical clustering~\cite{CHIME, HTCLLM}. 
While effective for corpus-level text analysis, these methods typically treat documents as independent instances, ignoring the rich relational signals provided by the graph topology~\cite{HG2M}. 
\texttt{SHiFT} bridges these gaps by leveraging graph topology to guide the induction process and utilizing LLMs to generate an explicit, human-readable hierarchy, rather than relying on static external taxonomies or opaque geometric coordinates.

\subsection{LLMs for Text-Attributed Graphs}
The integration of LLMs into TAG learning has gained significant attention~\cite{LLM4Graph, HTAG}. 
Existing works mainly fall into three paradigms as follows. 
(1) \textit{LLM as Encoder}: Using LLMs to generate semantically rich node embeddings, which serve as enhanced input features for downstream GNNs~\cite{TAPE, GAugLLM, SKETCH}. 
(2) \textit{LLM as Predictor}: Flattening graph structures into textual prompts, enabling LLMs to directly perform classification~\cite{GraphICL}, zero-shot prediction~\cite{DENSE}, or explanation generation~\cite{GraphNarrator}. 
(3) \textit{LLM-GNN Alignment}: Employing techniques like contrastive learning to align the latent spaces of GNNs and LLMs, facilitating mutual enhancement~\cite{GRENADE, LensGNN}.
However, most methods operate on a flat semantic plane, neglecting the semantic hierarchy inherent in TAGs~\cite{TIER}.
In contrast, \texttt{SHiFT} leverages LLMs not as flat predictors, but as efficient taxonomists to induce and refine a global hierarchical taxonomy, which in turn serves as a structural prior to regularize the representation learning process.

\section{Preliminaries}\label{sec:preliminaries}

In this section, we formally define the problem and provide theoretical motivation for hierarchy-aware embeddings.

\subsection{Problem Definition}
Let $\mathcal{G} = (\mathcal{V}, \mathcal{E}, \mathcal{S})$ be a TAG with $n$ nodes, edges $\mathcal{E} \subseteq \mathcal{V} \times \mathcal{V}$, adjacency matrix $\boldsymbol{A} \in \{0,1\}^{n \times n}$. Each node $v_i$ is associated with a raw textual document $s_i \in \mathcal{S}$. 
Following LLMNodeBed~\cite{LLMNodeBed}, we utilize a frozen pretrained language model (PLM), such as RoBERTa~\cite{RoBERTa}, to encode the raw texts into initial node features $\boldsymbol{X} = [\boldsymbol{x}_1, \dots, \boldsymbol{x}_n]^\top \in \mathbb{R}^{n \times d_{in}}$.
A semantic hierarchy is a rooted tree $\mathcal{T} = (\mathcal{C}, \mathcal{R})$, where each concept $c \in \mathcal{C}$ is characterized by a (name, definition) tuple, and parent-child edges $\mathcal{R}$ encode ``is-a'' relations.
Each graph node $v_i \in \mathcal{V}$ is assigned a unique root-to-leaf path $P(v_i) = \{c_{(0)}, \dots, c_{(L)}\}$ representing its semantic lineage from the global domain to its finest-grained concept.
Given $\mathcal{G}$, our goal is to jointly learn:
(1) a GNN encoder $f_\theta: (\mathcal{G}, \boldsymbol{X}) \rightarrow \boldsymbol{Z} \in \mathbb{R}^{n \times d}$ mapping nodes to a hierarchy-aware embedding space; and
(2) a latent taxonomy $\mathcal{T}$ with assignment $\psi: \mathcal{V} \rightarrow \mathrm{Leaves}(\mathcal{T})$ organizing nodes into semantically coherent multi-level clusters.

\subsection{Theoretical Motivation: Hierarchy-Aware vs.\ Flat Embeddings}
\label{sec:theory_motivation}

Real-world TAGs often exhibit intrinsic hierarchical organization, such as paper topics in citation networks, item categories in e-commerce graphs, and user interests in social networks.
We formalize this intuition and prove that hierarchy-aware embeddings are provably superior to flat embeddings:

\vpara{Setup.}
Let $\mathcal{X}$ be a set of data points whose latent semantic structure is governed by a rooted tree $\mathcal{T}^*$ with $K$ leaf concepts $\{c_1,\dots,c_K\}$ and depth $L$.
For a point $x$, let $\ell(x)\in\{1,\dots,K\}$ denote its ground-truth leaf concept.
For two independently sampled points $(x_i,x_j)$, define their latent semantic distance as $D^*_{ij}=d_{\mathcal{T}^*}(\ell(x_i),\ell(x_j))$,
where $d_{\mathcal{T}^*}(\cdot,\cdot)$ is the tree-path distance between leaves.
For an encoder $f:\mathcal{X}\to\mathbb{R}^d$, we measure how much hierarchy it preserves by the mutual information $\mathcal{I}_{\mathrm{hier}}(f) = I\!\left( D^*_{ij}; \|f(x_i)-f(x_j)\|_2 \mid \mathcal{T}^* \right)$.
This quantity captures whether embedding distances reveal sibling, cousin, or more distant semantic relations in the latent hierarchy.

\begin{theorem}[Hierarchy Preservation and Flat Non-identifiability]
\label{thm:hier_preservation}
Under the hierarchical data-generating process in Definition~\ref{def:hdgp}, if a hierarchy-aware encoder $f^{\mathrm{hier}}$ is aligned with $\mathcal{T}^*$ with bounded error $(\varepsilon,r,\zeta)$, sufficient margin $\Delta$ and $\varepsilon+2r<\Delta/2$, then
\begin{equation}
    \mathcal{I}_{\mathrm{hier}}(f^{\mathrm{hier}})
    \geq
    H(D^*_{ij}\mid\mathcal{T}^*)
    -
    h_2(2\zeta)
    -
    2\zeta\log(|\mathcal{D}|-1),
\end{equation}
where $\mathcal{D}$ is the set of possible tree distances and $h_2(\cdot)$ is the binary entropy function.
In contrast, there exists a flat embedding $f^{\mathrm{flat}}$ that perfectly separates all leaf concepts but fails to encode hierarchical relations:
\begin{equation}
    I\!\left(
    D^*_{ij};
    \|f^{\mathrm{flat}}(x_i)-f^{\mathrm{flat}}(x_j)\|_2
    \mid
    D^*_{ij}>0,\mathcal{T}^*
    \right)
    =
    0.
\end{equation}
\end{theorem}

The proof is provided in Appendix~\ref{sec:theory_motivation_appendix}.
Theorem~\ref{thm:hier_preservation} shows that flat embeddings may be sufficient for separating leaf clusters but are not identifiable with respect to the semantic skeleton: sibling and distant concepts can collapse to the same inter-cluster distance.
By contrast, hierarchy-aware embeddings preserve pairwise tree-distance information, making the learned space more suitable for semantic-rich TAG tasks that depend on the latent concept structure.
Therefore, modeling the latent semantic hierarchy is not merely a heuristic design choice but a theoretically motivated idea.

\section{Methodology}\label{sec:methodology}
\begin{figure*}[t]
    \centering
    \includegraphics[width=1.\textwidth]{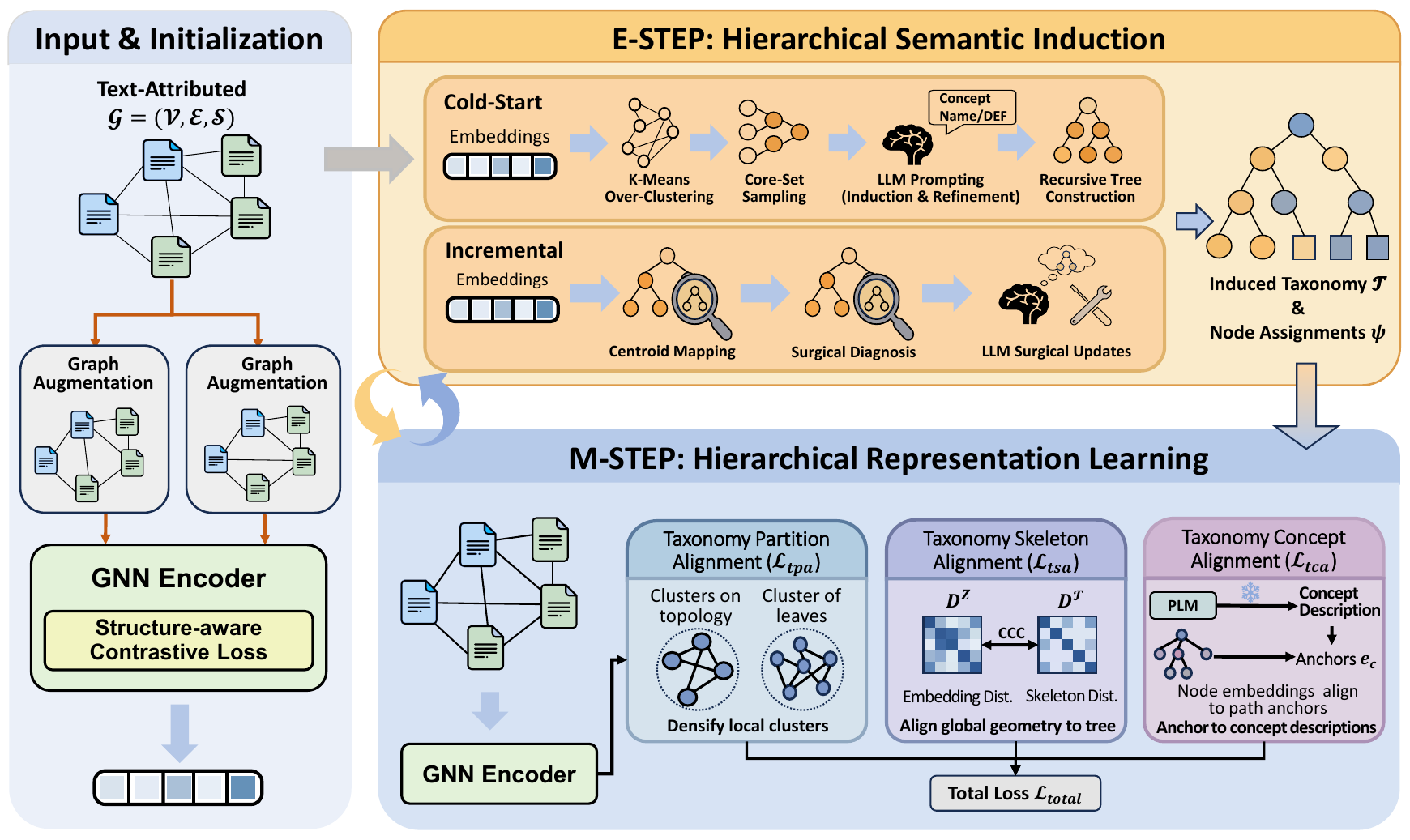}
    \caption{The overview of \texttt{SHiFT}. 
    (Left) The model is initialized via structure-aware pretraining.
    (Top, \textbf{E-Step}) The LLM acts as an expert taxonomist to induce or incrementally update the semantic hierarchy $\mathcal{T}$ based on current node embeddings.
    (Bottom, \textbf{M-Step}) The GNN encoder is optimized to internalize the induced hierarchy through three complementary objectives: Taxonomy Partition Enhancement, Taxonomy Skeleton Alignment, and Taxonomy Concept Alignment.}
    \label{fig:framework}
    \vspace{-1.2em}
\end{figure*}

In this section, we present \texttt{SHiFT} (Figure~\ref{fig:framework}), which jointly learns a hierarchy-aware graph encoder and an explicit semantic taxonomy within an Expectation-Maximization (EM) paradigm.

\subsection{Graph Contrastive Pretraining}
Since raw PLM features $\boldsymbol{X}$ lack structural awareness, we perform structure-aware pretraining to initialize the node embeddings~\cite{BLNN, APF}.
We follow standard graph contrastive learning pipelines~\cite{NS4GC, SSGE}:
given the graph $\mathcal{G}$, we first generate two augmented views $\mathcal{G}^1, \mathcal{G}^2$ via random edge dropping and feature masking.
The GNN encoder $f_\theta$ maps these views to $\ell_2$-normalized embeddings $\boldsymbol{Z}_1, \boldsymbol{Z}_2 \in \mathbb{R}^{n \times d}$.
We minimize a structure-aware contrastive loss:
\begin{equation}
\mathcal{L}_{gcp} = - \sum_{(i, j) \in \mathcal{P}} \boldsymbol{z}_1^i \cdot \boldsymbol{z}_2^j + \gamma \sum_{(i, j) \notin \mathcal{P}}(\boldsymbol{z}_1^i \cdot \boldsymbol{z}_2^j)^2,\label{eq:gcp}
\end{equation}
where $\mathcal{P} = \{(i, j) \mid \boldsymbol{A}_{ij} = 1\}$ denotes the set of positive pairs (including self-loops and connected neighbors) derived from the homophily assumption.
This yields structure-enhanced initial embeddings $\boldsymbol{Z}^{(0)}$ for the subsequent E-step.

\subsection{E-Step: Hierarchical Semantic Induction}
\label{sec:e_step}

Given the current node embeddings, the E-step estimates a latent semantic hierarchy $\mathcal{T}$ and node-to-leaf assignments $\psi$. 
Our goal is not only to partition nodes, but also to induce a human-readable taxonomy. 
We use embeddings to first cluster the graph into compact semantic groups, and then invoke the LLM on representative documents to construct or update the hierarchy.

\subsubsection{Cold-Start Taxonomy Initialization}
\label{sec:taxonomy_initialization}

We initialize the taxonomy in a top-down recursive manner. 
Starting from the root concept, which represents the global domain, suppose a parent concept $c_p$ contains a document set $D_p$. 
We first apply K-Means to the embeddings of nodes in $D_p$ with an intentionally large cluster number, e.g., $K_{init}=\sqrt{|D_p|/50}$. 
This over-clustering step favors high-purity micro-clusters, providing the LLM with localized and coherent evidence rather than a noisy global document collection.
Since directly feeding all documents in each cluster to an LLM is infeasible, we then select a small core set $\mathcal{S}_k$ of size $M$ from each micro-cluster $C_k$. 
The sampling score combines semantic centrality and structural importance $\mathrm{Score}(v) = a \cdot \cos(\boldsymbol{z}_v,\boldsymbol{\mu}_k) + (1-a) \cdot \mathrm{CCPR}(v \mid C_k)$,
where $\boldsymbol{\mu}_k$ is the centroid of $C_k$, and $\mathrm{CCPR}(v \mid C_k)$ is Cluster-Conditional PageRank computed on the subgraph induced by $C_k$. 
Thus, the selected documents are both semantically representative and structurally central.

Given the parent concept $c_p$ and each core set $\mathcal{S}_k$, the LLM acts as a domain taxonomist and generates a candidate child concept with a name, a definition, and a \texttt{split\_needed} flag indicating whether further refinement is necessary. 
Because over-clustering may produce redundant or fragmented candidates, we further use an LLM-based editor to refine all candidates under the same parent through three operations: \texttt{MERGE} synonymous or overly fine-grained concepts, \texttt{PRUNE} outliers irrelevant to the parent domain, and \texttt{KEEP} distinct meaningful concepts. 
The resulting children form a semantically coherent and human-readable expansion of $c_p$.
We recursively apply this procedure to each retained child concept until reaching the maximum depth $L_{max}$, the minimum cluster size $N_{min}$, or a negative \texttt{split\_needed} decision. 
This yields the initial taxonomy $\mathcal{T}^{(0)}$ and assignment function $\psi^{(0)}$, where each node is associated with a unique root-to-leaf semantic path.
Prompts are provided in Appendix~\ref{app:prompts}. 

\subsubsection{Incremental Taxonomy Updates}
\label{sec:taxonomy_updates}

In subsequent iterations, re-generating the whole taxonomy is unnecessary and costly, since most semantic regions remain stable across iterations. 
We instead propose a lightweight Mapping \& Diagnosis strategy that reuses the previous hierarchy $\mathcal{T}^{(t-1)}$ as a set of semantic anchors, first mapping nodes to existing concepts using the updated embeddings, and then invoking the LLM only when local evidence suggests that the current hierarchy no longer explains the data well.

Specifically, we update the centroid $\boldsymbol{\mu}_c$ of each concept $c \in \mathcal{T}^{(t-1)}$ using the embeddings of its currently assigned nodes. 
Each node is then reassigned by a top-down greedy search: starting from the root, it is mapped to the child concept whose centroid has the highest cosine similarity, and this process continues until a leaf is reached. 
If the maximum similarity at any level falls below a threshold $\delta$ (e.g., 0.5), the node is placed into an Unassigned Bucket. 
This centroid-based mapping produces a provisional assignment without requiring LLM calls.

After mapping, we diagnose the taxonomy for four types of semantic inconsistency. 
First, a \emph{Split} is triggered when a concept exhibits low cohesion, such as a sharp increase in intra-cluster variance or a substantial growth in cluster size; in this case, we rerun the recursive construction procedure within the affected concept to discover finer sub-concepts. 
Second, a \emph{Merge} is considered when two sibling concepts become highly redundant, e.g., when the cosine similarity between their centroids exceeds $\tau_{merge}$; the LLM is asked to verify whether they denote the same or strongly overlapping concept. 
Third, a \emph{Redefine} operation is invoked when a concept undergoes significant composition change, such as more than $50\%$ member turnover; the LLM receives the newly representative documents and revises the concept name or definition. 
Finally, for nodes in the Unassigned Bucket, a \emph{Reassign} operation asks the LLM to either attach them to the most appropriate existing concept or create a new concept if they form a coherent emerging group.

This incremental procedure updates only the semantically unstable parts of the hierarchy while keeping the rest unchanged. 
As a result, the E-step maintains an explicit and human-readable taxonomy that evolves with the GNN embeddings, but avoids repeatedly querying the LLM over the full graph. 
The updated hierarchy $\mathcal{T}^{(t)}$ and assignments $\psi^{(t)}$ are then passed to the M-step for hierarchy-aware representation learning.

By limiting LLM calls to these targeted scenarios, Mapping \& Diagnosis reduces LLM invocations by ${\approx}80\%$ in later epochs while keeping the taxonomy well-aligned with the evolving embedding space (see Appendix~\ref{app:efficiency}).
Full prompts for taxonomy induction are provided in Appendix~\ref{app:prompts}. 
We further present visualizations of induced taxonomies in Appendix~\ref{app:qualitative_analysis}.

\subsection{M-Step: Hierarchical Representation Learning}\label{sec:m_step}

In this step, we fix the taxonomy $\mathcal{T}$ and node assignments $\psi$ derived from the E-step to optimize the GNN encoder $f_\theta$.
The core objective is to align the continuous node embeddings with the discrete hierarchical structure induced by the LLM.
We decompose the taxonomy into three distinct facets: the clustering partition, the topological skeleton, and the semantic concepts.
Correspondingly, we propose three alignment objectives to internalize these facets into the embedding space: 

\subsubsection{Taxonomy Partition Alignment (TPA)}
The E-step partitions the graph nodes into disjoint semantic groups based on their assignment to leaf concepts.
Our first objective serves as a \textit{Local Consistency} constraint: the continuous embedding space must respect these discrete partitions.
Unlike prior graph pretraining which relies solely on edge connectivity, TPA explicitly pulls together topologically distant but semantically similar nodes.

We construct a partition-aware adjacency matrix $\boldsymbol{A}^+$, where $\boldsymbol{A}^+_{ij}=1$ if nodes $i$ and $j$ belong to the same leaf cluster (i.e., $\psi(v_i) = \psi(v_j)$) or are topologically connected.
We then employ a contrastive objective to pull intra-partition nodes together while pushing inter-partition nodes apart:
\begin{equation}
\mathcal{L}_{tpa} = - \sum_{(i, j) \in \mathcal{P^+}} \boldsymbol{z}_1^i \cdot \boldsymbol{z}_2^j + \gamma \sum_{(i, j) \notin \mathcal{P}^+}(\boldsymbol{z}_1^i \cdot \boldsymbol{z}_2^j)^2,\label{eq:tpa}
\end{equation}
where $\mathcal{P}^+ = \{(i,j) \mid \boldsymbol{A}^+_{ij}=1\}$.
By optimizing $\mathcal{L}_{tpa}$, the encoder respects the decision boundaries established by the taxonomy's finest granularity.

\subsubsection{Taxonomy Skeleton Alignment (TSA)}
While TPA ensures local cluster cohesion, it offers no guidance on how these clusters should be arranged relative to one another.
To address this, TSA imposes a \textit{Global Geometric} constraint based on the skeleton of the hierarchy tree $\mathcal{T}$.
Intuitively, the distances between cluster centroids in the vector space should preserve the path distances in the tree (e.g., sibling concepts should be geometrically closer than cousin concepts). 

We enforce this using the Cophenetic Correlation Coefficient (CCC)~\cite{CCC} as a differentiable objective. 
Let $\{\boldsymbol{\mu}_1, \dots, \boldsymbol{\mu}_K\}$ be the centroid embeddings of the $K$ leaf concepts, calculated as the mean of member node embeddings. 
We align two distance matrices: 
(1) The \textit{Skeleton Distance} $\boldsymbol{D}^{\mathcal{T}}$, defined as the shortest path length between concept nodes in $\mathcal{T}$; and
(2) The \textit{Embedding Distance} $\boldsymbol{D}^{\boldsymbol{Z}}$, defined as the Euclidean distance $\|\boldsymbol{\mu}_u - \boldsymbol{\mu}_v\|_2$. 
The alignment loss $\mathcal{L}_{tsa}$ maximizes the Pearson correlation between the upper triangular values of these matrices (i.e., CCC):
\begin{equation}
\mathcal{L}_{tsa} = - \frac{\text{Cov}(\text{vec}(\boldsymbol{D}^{\boldsymbol{Z}}), \text{vec}(\boldsymbol{D}^{\mathcal{T}}))}{\sigma_{\boldsymbol{D}^{\boldsymbol{Z}}} \cdot \sigma_{\boldsymbol{D}^{\mathcal{T}}}},
\end{equation}
where $\text{Cov}(\cdot)$ is the covariance, $\text{vec}(\cdot)$ flattens the upper triangle, and $\sigma$ is the standard deviation. 
This objective acts as a global regularizer, ensuring the graph embeddings explicitly encode the hierarchical relations (e.g., depth and proximity) inherent in the taxonomy skeleton.

\subsubsection{Taxonomy Concept Alignment (TCA)}
The previous two objectives focus on the structural alignment (partition and skeleton) but remain agnostic to the concept content.
To inject the concept semantics, TCA introduces a \textit{Semantic Grounding} constraint.
It ensures that the node embeddings are not just distinct vectors, but are interpretable representations aligned with the LLM-generated definitions.

For each concept $c \in \mathcal{C}$, we encode its text description (name + definition) into a semantic anchor vector $\boldsymbol{e}_c$ using the frozen PLM.
For a node $v_i$, let $P(v_i) = \{c_{(0)}, \dots, c_{(L)}\}$ be the path from the root to its assigned leaf.
We introduce a projection head $g_\phi$ to map node embeddings to the semantic space and minimize the distance to their conceptual anchors along the hierarchical path:
\begin{equation}
    \mathcal{L}_{tca} = \sum_{i \in \mathcal{V}} \sum_{l=1}^{L} \lambda^{l} \cdot \| g_\phi(\boldsymbol{z}_i) - \boldsymbol{e}_{c_{(l)}} \|_2^2.
\end{equation}
Here, we employ an exponential growth weight $w_l = \lambda^{l}$ (with $\lambda > 1$). This design assigns higher importance to finer-grained concepts ($c_{(L)}$) which possess stronger discriminative power, while maintaining a loose constraint on broader categories ($c_{(1)}$), thereby preserving the coarse-to-fine semantic structure.

\subsubsection{Optimization and Iteration}
The overall M-step objective aggregates these three complementary views:
\begin{equation}
    \mathcal{L}_{total} = \mathcal{L}_{tpa} + \alpha \mathcal{L}_{tsa} + \beta \mathcal{L}_{tca}.
\end{equation}
where $\alpha$ and $\beta$ are balancing hyperparameters.
Through this unified optimization, \texttt{SHiFT} ensures that the learned embeddings are locally partitioned, globally structured by the taxonomy skeleton, and semantically grounded in explicit concepts.
Upon convergence of the M-step, the updated embeddings $\boldsymbol{Z}$ are fed back into the E-step, where the Mapping \& Diagnosis strategy (as described in Sec~\ref{sec:taxonomy_updates}) adapts the hierarchy to the evolved embedding space.

\vpara{Efficiency Analysis.}
In Appendix~\ref{app:efficiency}, we analyze the efficiency of \texttt{SHiFT} by presenting its theoretical complexity and empirical runtime comparison against baselines.

\subsection{Theoretical Analysis}

We further establish a principled probabilistic foundation for \texttt{SHiFT}.
Let $\boldsymbol{H}{=}(\mathcal{T},\psi)$ denote the latent hierarchy and $\boldsymbol{Z}$ the continuous node representations (please kindly see Appendix~\ref{sec:theory} for details).
We posit that the observed adjacency $\boldsymbol{A}$ is generated via:
$p_\theta(\boldsymbol{A}) {=} \sum_{\psi,\mathcal{T}} \int p(\boldsymbol{A}|\boldsymbol{Z},\psi)\, p_\theta(\boldsymbol{Z}|\psi,\mathcal{T})\, p(\psi|\mathcal{T})\,p(\mathcal{T})\,\mathrm{d}\boldsymbol{Z}$,
where $p(\mathcal{T})$ encodes the semantic prior, and edges arise from both homophily and partition structure~\cite{SBM}. 
Under a mean-field approximation 
$q(\boldsymbol Z,\psi,\mathcal T)
= q_\theta(\boldsymbol Z)q(\psi,\mathcal T)$~\cite{VEM} with a Hard-EM approximation~\cite{Hard-EM1}, the ELBO is given by:
\begin{equation}
\mathcal{L}_{\mathrm{ELBO}} =
\underbrace{\mathbb{E}_q[\log p(\boldsymbol{A}|\boldsymbol{Z},\psi)]}
_{\text{(I) reconstruction}}
+\underbrace{\mathbb{E}_q[\log p_\theta(\boldsymbol{Z}|\psi,\mathcal{T})]}
_{\text{(II) embedding prior}}
-\underbrace{\mathrm{KL}(q(\psi,\mathcal{T})\,\|\,p(\psi,\mathcal{T}))}
_{\text{(III) hierarchy regularization}}
+\underbrace{H(q_\theta(\boldsymbol{Z}))}
_{\text{(IV) embed. entropy}}.
\label{eq:elbo_main}
\end{equation}
Our two steps perform alternating coordinate ascent on $\mathcal{L}_{\mathrm{ELBO}}$: the \textbf{E-step} minimizes Term~(III) by searching for $(\hat{\mathcal{T}},\hat{\psi})$ that maximizes the LLM-proxied semantic prior; the \textbf{M-step} maximizes Terms~(I)--(II), with a precise correspondence: 
$\mathcal{L}_{tpa}$ operationalizes Term~(I) via partition-augmented contrastive learning;
$\mathcal{L}_{tsa}$ maximizes Term~(II) under a metric-preserving structural prior 
$p(\boldsymbol{Z}|\mathcal{T}){\propto}\exp({-}\mathrm{dist}
(\boldsymbol{D}^{\boldsymbol{Z}},\boldsymbol{D}^{\mathcal{T}}))$;
$\mathcal{L}_{tca}$ maximizes Term~(III) under a Gaussian content prior
$p(\boldsymbol{z}_i|\hat{\psi}_i){=}\mathcal{N}(\boldsymbol{z}_i;\boldsymbol{e}_{\hat{\psi}_i},\sigma^2)$.
Our LLM-based discrete updates and non-convex neural objectives optimize surrogate terms associated with the same ELBO, leading empirically to stable convergence. The induced hierarchy and learned embeddings progressively become mutually consistent.
Full derivations are in Appendix~\ref{sec:theory}.

\section{Experiments}

\begin{table*}[!t]
\centering
\small
\caption{Node classification accuracy (\%). The \colorbox{orange!25}{\textbf{best}} and \colorbox{orange!10}{second-best} results are highlighted.}
\label{tab:node_classification}
\begin{tabular}{lcccccc}
\toprule
\textbf{Method} & \textbf{Cora} & \textbf{Citeseer} & \textbf{ArXiv} & \textbf{WikiCS} & \textbf{Computer} & \textbf{Avg.} \\
\midrule
\rowcolor{COLOR_MEAN} \multicolumn{7}{c}{\textbf{Classic/Hyperbolic GNNs and PLMs}} \\
GCN$_{\mathrm{SE}}$~\cite{GCN} & 82.30 \scriptsize{±0.19} & 70.55 \scriptsize{±0.32} & 71.39 \scriptsize{±0.28} & 79.86 \scriptsize{±0.19} & 71.44 \scriptsize{±1.19} & 75.11 \\
SAGE$_{\mathrm{SE}}$~\cite{SAGE} & 82.27 \scriptsize{±0.37} & 69.56 \scriptsize{±0.43} & 71.21 \scriptsize{±0.18} & 79.67 \scriptsize{±0.25} & 81.43 \scriptsize{±0.27} & 76.83 \\
GAT$_{\mathrm{SE}}$~\cite{GAT} & 81.30 \scriptsize{±0.67} & 69.94 \scriptsize{±0.74} & 71.57 \scriptsize{±0.25} & 79.99 \scriptsize{±0.65} & 83.39 \scriptsize{±0.22} & 77.24 \\
HGCN~\cite{HGCN} & 80.38 \scriptsize{±0.69} & 70.83 \scriptsize{±0.62} & 70.39 \scriptsize{±0.38} & 79.26 \scriptsize{±0.39} & 79.44 \scriptsize{±0.49} & 76.06 \\
HGCL~\cite{HGCL} & 82.37 \scriptsize{±0.47} & 72.11 \scriptsize{±0.64} & 71.40 \scriptsize{±0.68} & 80.16 \scriptsize{±0.35} & 80.28 \scriptsize{±0.55} & 77.26 \\
SenBERT~\cite{SentenceBERT} & 66.66 \scriptsize{±1.42} & 60.52 \scriptsize{±1.62} & 72.66 \scriptsize{±0.24} & 77.77 \scriptsize{±0.75} & 70.76 \scriptsize{±0.15} & 69.67 \\
RoBERTa~\cite{RoBERTa} & 72.24 \scriptsize{±1.14} & 66.68 \scriptsize{±2.03} & 74.12 \scriptsize{±0.12} & 76.81 \scriptsize{±1.04} & 72.31 \scriptsize{±0.37} & 72.43 \\
\midrule

\rowcolor{COLOR_MEAN} \multicolumn{7}{c}{\textbf{Hybrid PLM/LLM + GNN Methods}} \\
GIANT~\cite{GIANT} & 81.08 \scriptsize{±1.07} & 66.02 \scriptsize{±1.28} & 74.29 \scriptsize{±0.10} & 80.48 \scriptsize{±0.65} & 86.16 \scriptsize{±0.12} & 77.61 \\
GLEM~\cite{GLEM} & 81.30 \scriptsize{±0.88} & 68.80 \scriptsize{±2.46} & 73.55 \scriptsize{±0.22} & 76.43 \scriptsize{±0.55} & 80.46 \scriptsize{±1.45} & 76.11 \\
ENGINE~\cite{ENGINE} & \colorbox{orange!10}{84.22 \scriptsize{±0.46}} & \colorbox{orange!10}{72.14 \scriptsize{±0.74}} & 74.69 \scriptsize{±0.36} & 80.94 \scriptsize{±0.19} & 86.42 \scriptsize{±0.23} & 79.68 \\
GCN$_{\mathrm{LLM}}$~\cite{LLMNodeBed} & 83.33 \scriptsize{±0.75} & 71.39 \scriptsize{±0.90} & 74.39 \scriptsize{±0.31} & 80.94 \scriptsize{±0.16} & 88.22 \scriptsize{±0.16} & 79.65 \\
TAPE~\cite{TAPE} & 84.04 \scriptsize{±0.24} & 71.87 \scriptsize{±0.35} & 74.96 \scriptsize{±0.14} & \colorbox{orange!10}{81.94 \scriptsize{±0.16}} & \colorbox{orange!25}{\textbf{89.52 \scriptsize{±0.04}}} & \colorbox{orange!10}{80.47} \\
GAugLLM~\cite{GAugLLM} & 82.84 \scriptsize{±0.99} & 70.19 \scriptsize{±0.92} & 73.59 \scriptsize{±0.10} & 80.32 \scriptsize{±0.32} & 87.79 \scriptsize{±0.23} & 78.95 \\
GraphGPT~\cite{GraphGPT} & 64.72 \scriptsize{±1.50} & 64.58 \scriptsize{±1.55} & 75.15 \scriptsize{±0.14} & 70.71 \scriptsize{±0.37} & 80.10 \scriptsize{±0.76} & 71.05 \\
LLaGA~\cite{LLaGA} & 78.94 \scriptsize{±1.14} & 62.61 \scriptsize{±3.63} & 74.49 \scriptsize{±0.23} & 76.47 \scriptsize{±2.20} & 87.82 \scriptsize{±0.53} & 76.07 \\
LLM$_{\mathrm{IT}}$~\cite{LLMNodeBed} & 67.00 \scriptsize{±0.16} & 54.26 \scriptsize{±0.22} & \colorbox{orange!10}{76.08 \scriptsize{±0.00}} & 75.02 \scriptsize{±0.16} & 66.99 \scriptsize{±2.02} & 67.87 \\
\midrule

\textbf{\texttt{SHiFT} (Ours)} & \colorbox{orange!25}{\textbf{84.83 \scriptsize{±0.58}}} & \colorbox{orange!25}{\textbf{73.96 \scriptsize{±0.92}}} & \colorbox{orange!25}{\textbf{76.56 \scriptsize{±0.13}}} & \colorbox{orange!25}{\textbf{82.40 \scriptsize{±0.48}}} & \colorbox{orange!10}{89.26 \scriptsize{±0.34}} & \colorbox{orange!25}{\textbf{81.40}} \\
\bottomrule
\end{tabular}
\end{table*}

\subsection{Experimental Setup}
\label{sec:experimental_setup}

\vpara{Datasets \& Baselines.} 
We evaluate \texttt{SHiFT} on 5 datasets from the \texttt{LLMNodeBed} benchmark~\cite{LLMNodeBed}, spanning academic, e-commerce, and web domains, and cover a wide range of scales, from 2.7K to 169K nodes. 
We compare against a diverse set of baselines, from Classic \& Hyperbolic GNNs to Hybrid LLM+GNN Methods. 
Further details are provided in Appendix~\ref{app:setup_details}.

\vpara{Implementation Details.}
Following the \texttt{LLMNodeBed}~\cite{LLMNodeBed} benchmark, we employ RoBERTa-Large (355M)~\cite{RoBERTa} as the frozen PLM and GCN~\cite{GCN} as the encoder backbone; 
For the taxonomy induction, we prompt DeepSeek-V3~\cite{DeepSeekV3} via API. 
After representation learning, we attach a linear head and fine-tune for classification under the standard semi-supervised setting.
Accuracy is reported with the mean and standard deviation across 5 random seeds.
For node clustering, we apply K-Means to the learned embeddings, setting the cluster number to the ground-truth class count.
More details are presented in Appendix~\ref{app:implementation}.
Our code can be found at \url{https://github.com/Cloudy1225/SHiFT}.

\subsection{Evaluation of Node Representations}\label{sec:representation_evaluation}
\vpara{Node Classification.} 
As shown in Table~\ref{tab:node_classification}, \texttt{SHiFT} achieves the highest average accuracy, surpassing the strongest baseline TAPE by about 1\% and traditional GCN by over 6\%. 
We make three key observations:
(1) \emph{Explicit vs. Implicit Hierarchy.} \texttt{SHiFT} significantly outperforms hyperbolic methods (HGCN, HGCL). While hyperbolic spaces implicitly model hierarchy via geometry, \texttt{SHiFT} explicitly injects semantic structure through the taxonomy, proving that semantic grounding is crucial for TAGs.
\begin{wraptable}{r}{0.5\textwidth}
\vspace{-1.8em}
\centering
\caption{Node clustering performance (\%).}
\label{tab:node_clustering}
\resizebox{0.5\textwidth}{!}{
\begin{tabular}{lcccccc}
\toprule
\multirow{2}{*}{\textbf{Method}} 
 & \multicolumn{2}{c}{\textbf{Cora}} 
 & \multicolumn{2}{c}{\textbf{Citeseer}} 
 & \multicolumn{2}{c}{\textbf{WikiCS}} \\
\cmidrule(r){2-3} \cmidrule(r){4-5} \cmidrule(r){6-7}
 & \textbf{NMI} & \textbf{ARI} 
 & \textbf{NMI} & \textbf{ARI} 
 & \textbf{NMI} & \textbf{ARI} \\
\midrule
\rowcolor{COLOR_MEAN} \multicolumn{7}{c}{\textbf{GNN-based Methods}} \\
DMoN~\cite{DMoN} & 49.00 & 43.70 & 39.10 & 38.40 & 23.50 & 9.00 \\
MAGI~\cite{MAGI} & 52.90 & 46.40 & 41.10 & 39.60 & 46.70 & 38.10 \\
\midrule
\rowcolor{COLOR_MEAN} \multicolumn{7}{c}{\textbf{LLM+GNN Methods}} \\
GCLR~\cite{GCLR} & 50.87 & 59.20 & 43.30 & 42.49 & 32.51 & 36.49 \\
MARK~\cite{MARK} & 57.80 & 49.00 & 43.80 & 43.20 & 49.30 & 46.60 \\
\midrule
\textbf{\texttt{SHiFT} (Ours)} & \textbf{62.13} & \textbf{60.10} & \textbf{45.76} & \textbf{48.42} & \textbf{51.92} & \textbf{46.98} \\
\bottomrule
\end{tabular}
}
\vspace{-1em}
\end{wraptable}
(2) \emph{Synergy of GNN and LLM.} Compared to methods like RoBERTa and LLM$_{\mathrm{IT}}$, which struggle to capture complex graph topology, \texttt{SHiFT} effectively combines the reasoning power of LLMs with the structural propagation of GNNs.
(3) \emph{Benefit of Hierarchical Alignment.} Methods like TAPE and GAugLLM use LLMs solely for feature enhancement. By contrast, \texttt{SHiFT} aligns the embedding space with the latent taxonomy skeleton. The performance gain is particularly notable on larger datasets like ArXiv ($+1.6\%$ over TAPE), suggesting that the explicit taxonomic guidance helps the GNN navigate complex semantic landscapes better than flat 
strategies.

\vpara{Node Clustering.}
We report NMI and ARI scores in Table~\ref{tab:node_clustering}.
\texttt{SHiFT} demonstrates superior capability in disentangling semantic structures without supervision.
Notably, on Cora, \texttt{SHiFT} improves NMI by 4.33\% over the recent SOTA LLM+GNN method MARK. 
This confirms that enforcing local cluster cohesion and global skeleton alignment effectively pulls semantically synonymous nodes together, even when the graph topology is noisy.

\vpara{Visual Analysis.}
To intuitively verify that \texttt{SHiFT} captures the intrinsic hierarchy rather than just flat clusters, we visualize the pairwise cosine similarity matrix of the learned embeddings in Appendix~\ref{app:visualization_analysis}.

\subsection{Evaluation of Induced Taxonomy}
\label{sec:taxonomy_evaluation}
The benchmarks above provide flat labels only, so we assess the induced taxonomy $\mathcal{T}$ via intrinsic coherence and extrinsic utility, and test it against a known hierarchy in Appendix~\ref{app:patent}.
Detailed setups are provided in Appendix~\ref{app:taxo_eval_protocols}.

\vpara{Concept Coherence (Intrinsic).} 
We quantify the semantic consistency of the induced concepts using Normalized Pointwise Mutual Information (NPMI). 
Specifically, we extract top-10 distinct keywords for each concept using Class-based TF-IDF and calculate their pairwise co-occurrence in the corpus (see Appendix~\ref{app:concept_coherence_calculation}). 
A higher score thus reflects high semantic purity. 
As shown in Table~\ref{tab:npmi_results}, \texttt{SHiFT} ($\mathcal{T}_{\text{final}}$) consistently outperforms the purely text-based baseline ($\mathcal{T}_{\text{w/o } \mathcal{G}}$, induced solely by PLM embeddings) and the initial taxonomy ($\mathcal{T}_{\text{init}}$).
This confirms that (1) integrating graph structure aids in disambiguating concepts, and (2) our EM-style iteration successfully refines the semantic clusters.

\begin{table*}[h]
\vspace{-1em}
\centering
\begin{minipage}{0.38\linewidth}
    \centering
    \caption{Concept Coherence (NPMI) of taxonomies.}
    \label{tab:npmi_results}
    \small
    \setlength{\tabcolsep}{5pt}
    \begin{tabular}{l c c c}
    \toprule
    \textbf{Taxonomy} & \textbf{Cora} & \textbf{Citeseer} & \textbf{WikiCS} \\
    \midrule
    $\mathcal{T}_{\text{w/o } \mathcal{G}}$ & 0.1640 & 0.1602 & 0.2116 \\
    $\mathcal{T}_{\text{init}}$             & 0.1852 & 0.1642 & 0.2204 \\
    $\mathcal{T}_{\text{final}}$ (Ours)     & \textbf{0.1958} & \textbf{0.1856} & \textbf{0.2410} \\
    \bottomrule
    \end{tabular}
\end{minipage}
\hfill
\begin{minipage}{0.52\linewidth}
    \centering
    \caption{Human evaluation results (1--5 Likert scale, higher is better). Fleiss' $\kappa = 0.61$.}
    \label{tab:human_eval}
    \small
    \setlength{\tabcolsep}{2pt}
    \begin{tabular}{lccc}
    \toprule
    \textbf{Taxonomy} & \textbf{Readability} & \textbf{Coherence} & \textbf{HierarchyValidity} \\
\midrule
$\mathcal{T}_{\text{w/o }\mathcal{G}}$  & 3.72 & 3.15 & 2.87 \\
$\mathcal{T}_{\text{init}}$ & 4.08 & 3.83 & 3.65 \\
$\mathcal{T}_{\text{final}}$ (Ours)        & \textbf{4.35} & \textbf{4.22} & \textbf{4.10} \\
    \bottomrule
    \end{tabular}
\end{minipage}
\end{table*}

\vpara{Human Evaluation (Intrinsic).}
To complement NPMI with a judgment-based assessment, we conducted a small-scale human evaluation.
Three CS graduate students evaluated 20 randomly sampled concept nodes on Cora, each presented with the concept name, definition, and 5 representative documents.
Annotators rated each concept on three criteria using a 1--5 Likert scale: \textit{Readability} (clarity of concept name/definition), \textit{Coherence} (whether assigned documents genuinely belong to the concept), and \textit{Hierarchy Validity} (logical soundness of parent-child relations).
Inter-annotator agreement was substantial (Fleiss' $\kappa = 0.61$)~\cite{Fleiss}.
As shown in Table~\ref{tab:human_eval}, \texttt{SHiFT} consistently outperforms both the text-only baseline and cold-start taxonomy across all criteria.
Notably, the largest margin appears in \textit{Hierarchy Validity} ($+1.23$ over text-only $\mathcal{T}_{\text{w/o }\mathcal{G}}$), confirming that graph topology is essential for inducing logically sound hierarchical relations, a quality that NPMI alone cannot fully capture.
Full evaluation protocols are in Appendix~\ref{app:human_eval}.

\vpara{Taxonomy as Features (Extrinsic).}
To assess whether the taxonomy captures high-level semantics invisible to flat embeddings, we perform a feature augmentation experiment. 
We convert the hierarchical path of each node into a text string $\mathcal{P}$ and append it to the raw document $\mathcal{S}_{\text{raw}}$ as input for 
MLP and GCN classifiers (see Appendix~\ref{app:taxonomy_as_features}).
Table~\ref{tab:feature_aug} reveals two insights:
(1) \emph{Semantic Injection:} Appending \texttt{SHiFT}-induced paths $\mathcal{P}_{\text{final}}$ to an MLP, which\begin{wraptable}{r}{0.52\textwidth}
    \vspace{-1.8em}
    \centering
    \caption{Classification accuracy using taxonomy-augmented features. $\mathcal{P}$ is the path description.}
    \label{tab:feature_aug}
    \resizebox{0.52\textwidth}{!}{
    \begin{tabular}{l c c c}
    \toprule
    \textbf{Input Features} & \textbf{Cora} & \textbf{Citeseer} & \textbf{WikiCS} \\
    \midrule
    \multicolumn{4}{l}{\textit{Backbone: MLP (Structure-agnostic)}} \\
    $\mathcal{S}_{\text{raw}}$ & 76.05 \scriptsize{±0.41} & 67.22 \scriptsize{±0.49} & 75.79 \scriptsize{±0.18} \\
    $\mathcal{S}_{\text{raw}} + \mathcal{P}_{\text{w/o } \mathcal{G}}$ & 72.68 \scriptsize{±0.34} & 67.51 \scriptsize{±0.58} & 77.50 \scriptsize{±0.24} \\
    $\mathcal{S}_{\text{raw}} + \mathcal{P}_{\text{init}}$ & 80.59 \scriptsize{±0.18} & 71.05 \scriptsize{±0.55} & 78.06 \scriptsize{±0.37} \\
    $\mathcal{S}_{\text{raw}} + \mathcal{P}_{\text{final}}$ & \textbf{81.91 \scriptsize{±0.55}} & \textbf{72.83 \scriptsize{±0.73}} & \textbf{78.33 \scriptsize{±0.21}} \\
    \midrule
    \multicolumn{4}{l}{\textit{Backbone: GCN (Structure-aware)}} \\
    $\mathcal{S}_{\text{raw}}$ & 81.70 \scriptsize{±1.01} & 69.71 \scriptsize{±0.64} & 79.95 \scriptsize{±0.15} \\
    $\mathcal{S}_{\text{raw}} + \mathcal{P}_{\text{w/o } \mathcal{G}}$ & 82.02 \scriptsize{±0.54} & 71.54 \scriptsize{±0.50} & 79.56 \scriptsize{±0.33} \\
    $\mathcal{S}_{\text{raw}} + \mathcal{P}_{\text{init}}$ & 82.95 \scriptsize{±0.52} & 70.76 \scriptsize{±0.55} & 80.28 \scriptsize{±0.34} \\
    $\mathcal{S}_{\text{raw}} + \mathcal{P}_{\text{final}}$ & \textbf{83.41 \scriptsize{±0.44}} & \textbf{72.08 \scriptsize{±0.51}} & \textbf{80.78 \scriptsize{±0.18}} \\
    \bottomrule
    \end{tabular}
    }
    \vspace{-1em}
    \end{wraptable} ignores graph structure, boosts accuracy by $\approx 6\%$ on Cora. This implies our taxonomy successfully encodes structural information into explicit textual concepts, allowing the MLP to read the graph structure.
(2) \emph{Quality vs. Noise:} Using paths from the graph-agnostic baseline ($\mathcal{P}_{\text{w/o } \mathcal{G}}$) sometimes degrades performance, indicating that pure semantic clustering may create noisy, topologically inconsistent concepts. 
$\mathcal{P}_{\text{final}}$ consistently boosts GCN performance, proving that our hierarchical concepts provide informative signals complementary to the raw features and adjacency matrix of the graph data.

\vpara{Qualitative Analysis.}
While quantitative metrics reflect statistical cohesion, they may not fully capture structural rationality.
We \textit{visually} compare the taxonomies induced by the text-only baseline ($\mathcal{T}_{\text{w/o }\mathcal{G}}$) and \texttt{SHiFT} ($\mathcal{T}_{\text{init}}$, $\mathcal{T}_{\text{final}}$) in Appendix~\ref{app:qualitative_analysis}. 
The final taxonomy exhibits a logical coarse-to-fine progression, resolving semantic ambiguities present in the text-only case. 

\vpara{Verifiable Hierarchy.}
Appendix~\ref{app:patent} evaluates \texttt{SHiFT} on \textbf{Patent}, a graph of 10{,}090 patents with examiner-assigned CPC labels at three levels (7/55/162).
The gain over TAPE grows from the section level to the subclass level ($+1.3$ to $+2.6$ accuracy), and hierarchical F-score~\cite{Kiritchenko2006} rises by $2.4$ points.
The induced tree also tracks the official CPC geometry more closely than a text-only taxonomy.

\subsection{Ablation Study}
\label{sec:ablation}

We validate \texttt{SHiFT}'s core components against four variants (Table~\ref{tab:ablation}): w/o EM (using $\mathcal{T}_{\text{init}}$ without iterative refinement), and removing TPA (partition), TSA (skeleton), or TCA (concept) alignment.

\vpara{Impact of Iterative Refinement.}
Removing the EM mechanism (\emph{w/o EM}) leads to significant 
\begin{wraptable}{r}{0.52\textwidth}
\vspace{-1em}
\centering
\caption{Ablation study on node classification accuracy.}
\label{tab:ablation}
\resizebox{0.52\textwidth}{!}{
\begin{tabular}{l c c c | c}
    \toprule
    \textbf{Method} & \textbf{Cora} & \textbf{Citeseer} & \textbf{WikiCS} & \textbf{Avg Drop} \\
    \midrule
    \textbf{\texttt{SHiFT} (Full)} & \textbf{84.83 \scriptsize{±0.58}} & \textbf{73.96 \scriptsize{±0.92}} & \textbf{82.40 \scriptsize{±0.48}} & - \\
    \midrule
    w/o EM (One-pass) & 83.27 \scriptsize{±0.37} & 72.62 \scriptsize{±0.75} & 81.48 \scriptsize{±0.54} & $\downarrow$ 1.27 \\
    w/o TPA (Partition) & 83.46 \scriptsize{±0.35} & 72.84 \scriptsize{±0.76} & 81.18 \scriptsize{±0.22} & $\downarrow$ 1.24 \\
    w/o TSA (Skeleton) & 83.34 \scriptsize{±0.27} & 72.31 \scriptsize{±0.94} & 81.38 \scriptsize{±0.23} & $\downarrow$ 1.39 \\
    w/o TCA (Concept) & 84.14 \scriptsize{±0.60} & 73.56 \scriptsize{±0.83} & 82.05 \scriptsize{±0.44} & $\downarrow$ 0.48 \\
    \bottomrule
\end{tabular}
}
\vspace{-0.5em}
\end{wraptable}
degradation across all datasets (e.g., $-1.27\%$ avg. drop). This quantitative decline directly validates our qualitative findings in Sec~\ref{sec:taxonomy_evaluation}: the structural evolution from the initialized $\mathcal{T}_{\text{init}}$ to the final $\mathcal{T}_{\text{final}}$ is not merely cosmetic. The iterative refinement allows the GNN to learn from a topologically cleaner hierarchy, proving that the coarse-to-fine granularity is essential for classification accuracy.

\vpara{Impact of Alignment Objectives.}
Deconstructing the M-step reveals the necessity of all three constraints.
(1) \emph{Local Cohesion (w/o TPA):} Removing partition alignment consistently hurts performance ($\approx -1.2\%$), as the model loses the explicit guidance to maintain sharp decision boundaries at the leaf level.
(2) \emph{Global Structure (w/o TSA):} This variant suffers the most on sparse graphs like Citeseer ($-1.65\%$), confirming that global geometric constraints are critical for preserving inter-community distances in citation networks.
(3) \emph{Semantic Grounding (w/o TCA):} While the drop is smaller, it remains significant. Anchoring embeddings to LLM-generated definitions provides semantic robustness, preventing the model from overfitting to pure structural signals.

\vpara{Supplementary Analysis.}
Due to space constraints, we provide comprehensive details in the Appendices to support the reproducibility and analysis of our framework. 
First, Appendix~\ref{sec:theory_motivation_appendix} proves the information-theoretic superiority of hierarchy-aware embeddings; Appendix~\ref{sec:theory} offers the formal derivation of \texttt{SHiFT} as a generalized Variational EM algorithm.
Regarding experimental reproducibility, Appendices~\ref{app:setup_details}, \ref{app:prompts}, and \ref{app:taxo_eval_protocols} detail the dataset statistics, implementation specifics, full lists of LLM prompts, and the rigorous protocols used for taxonomy evaluation.
To validate model robustness, Appendices~\ref{app:efficiency}, \ref{app:llm_sensitivity}, and \ref{app:hyperparameter} present efficiency profiling on large-scale graphs, sensitivity analyses across different LLM backbones, and hyperparameter studies.
Finally, to enhance interpretability, Appendices~\ref{app:visualization_analysis} and \ref{app:qualitative_analysis} provide visual investigations of the embedding space and qualitative case studies of the induced taxonomies.
Appendix~\ref{app:patent} reports hierarchical classification and taxonomy recovery on a CPC-annotated patent graph.
Appendix~\ref{app:misc} presents the limitations and broader impacts of our work.

\section{Conclusion}
In this work, we introduce \texttt{SHiFT} to uncover and model the semantic hierarchy in Text-Attributed Graphs.
By synergizing the semantic reasoning of LLMs with the structural propagation of GNNs within an Expectation-Maximization paradigm, our approach effectively bridges the gap between flat topological embeddings and rich hierarchical concepts.
Specifically, it aligns the continuous embedding space with the discrete taxonomy through partition, skeleton, and concept constraints.
We further provide a formal theoretical analysis showing \texttt{SHiFT} instantiates a generalized Variational EM algorithm.
Empirical results demonstrate that \texttt{SHiFT} not only achieves remarkable performance on downstream tasks but also yields human-readable taxonomies.

\begin{ack}
This work was supported by the National Science and Technology Major Project (2026ZD16011200), and the Postgraduate Research \& Practice Innovation Program of Jiangsu Province (26CXJH0412), for the authors affiliated with Nanjing University.
\end{ack}

\bibliographystyle{plainnat}
\bibliography{main}

@article{tenenbaum2011grow,
    title={How to Grow a Mind: Statistics, Structure, and Abstraction},
    author={Tenenbaum, Joshua B and Kemp, Charles and Griffiths, Thomas L and Goodman, Noah D},
    journal={science},
    volume={331},
    number={6022},
    pages={1279--1285},
    year={2011},
    url={https://www.science.org/doi/10.1126/science.1192788},
    publisher={American Association for the Advancement of Science}
}

@inproceedings{DENSE,
title={Dynamic Bundling with Large Language Models for Zero-Shot Inference on Text-Attributed Graphs},
author={Yusheng Zhao and Qixin Zhang and Xiao Luo and Weizhi Zhang and Zhiping Xiao and Wei Ju and Philip S. Yu and Ming Zhang},
booktitle={The Thirty-ninth Annual Conference on Neural Information Processing Systems},
year={2025},
url={https://openreview.net/forum?id=1nSynwHvu2}
}

@inproceedings{GraphICL,
    title = "{G}raph{ICL}: Unlocking Graph Learning Potential in {LLM}s through Structured Prompt Design",
    author = "Sun, Yuanfu  and
      Ma, Zhengnan  and
      Fang, Yi  and
      Ma, Jing  and
      Tan, Qiaoyu",
    booktitle = "Findings of the Association for Computational Linguistics: NAACL 2025",
    month = apr,
    year = "2025",
    address = "Albuquerque, New Mexico",
    publisher = "Association for Computational Linguistics",
    url = "https://aclanthology.org/2025.findings-naacl.131/",
    doi = "10.18653/v1/2025.findings-naacl.131",
    pages = "2440--2459",
    ISBN = "979-8-89176-195-7"
}

@inproceedings{BioOnto,
    title = "Ontology alignment in the biomedical domain using entity definitions and context",
    author = "Wang, Lucy Lu  and
      Bhagavatula, Chandra  and
      Neumann, Mark  and
      Lo, Kyle  and
      Wilhelm, Chris  and
      Ammar, Waleed",
    booktitle = "Proceedings of the {B}io{NLP} 2018 workshop",
    year = "2018",
    publisher = "Association for Computational Linguistics",
    url = "https://aclanthology.org/W18-2306/",
    doi = "10.18653/v1/W18-2306",
    pages = "47--55"
}

@inproceedings{DocTaxo,
    title = "Comparing Taxonomies for Organising Collections of Documents",
    author = "Fernando, Samuel  and
      Hall, Mark  and
      Agirre, Eneko  and
      Soroa, Aitor  and
      Clough, Paul  and
      Stevenson, Mark",
    booktitle = "Proceedings of {COLING} 2012",
    month = dec,
    year = "2012",
    address = "Mumbai, India",
    url = "https://aclanthology.org/C12-1054/",
    pages = "879--894"
}

@inproceedings{HiRAG,
    title = "Retrieval-Augmented Generation with Hierarchical Knowledge",
    author = "Huang, Haoyu  and
      Huang, Yongfeng  and
      Junjie, Yang  and
      Pan, Zhenyu  and
      Chen, Yongqiang  and
      Ma, Kaili  and
      Chen, Hongzhi  and
      Cheng, James",
    booktitle = "Findings of the Association for Computational Linguistics: EMNLP 2025",
    month = nov,
    year = "2025",
    address = "Suzhou, China",
    publisher = "Association for Computational Linguistics",
    url = "https://aclanthology.org/2025.findings-emnlp.321/",
    doi = "10.18653/v1/2025.findings-emnlp.321",
    pages = "6044--6060",
    ISBN = "979-8-89176-335-7"
}

@inproceedings{MemTree,
title={From Isolated Conversations to Hierarchical Schemas: Dynamic Tree Memory Representation for {LLM}s},
author={Alireza Rezazadeh and Zichao Li and Wei Wei and Yujia Bao},
booktitle={The Thirteenth International Conference on Learning Representations},
year={2025},
url={https://openreview.net/forum?id=moXtEmCleY}
}

@article{LLM4Graph,
  title={Large Language Models on Graphs: A Comprehensive Survey},
  author={Jin, Bowen and Liu, Gang and Han, Chi and Jiang, Meng and Ji, Heng and Han, Jiawei},
  journal={IEEE Transactions on Knowledge and Data Engineering},
  year={2024},
  doi={10.1109/TKDE.2024.3469578},
  publisher={IEEE}
}

@inproceedings{GCN,
title={Semi-Supervised Classification with Graph Convolutional Networks},
author={Thomas N. Kipf and Max Welling},
booktitle={International Conference on Learning Representations},
year={2017}
}

@inproceedings{GAT,
title={Graph Attention Networks},
author={Petar Veličković and Guillem Cucurull and Arantxa Casanova and Adriana Romero and Pietro Liò and Yoshua Bengio},
booktitle={International Conference on Learning Representations},
year={2018}
}

@article{SAGE,
  title={Inductive representation learning on large graphs},
  author={Hamilton, Will and Ying, Zhitao and Leskovec, Jure},
  journal={Advances in neural information processing systems},
  year={2017}
}

@article{HGCN,
  title={Hyperbolic graph convolutional neural networks},
  author={Chami, Ines and Ying, Zhitao and R{\'e}, Christopher and Leskovec, Jure},
  journal={Advances in neural information processing systems},
  volume={32},
  year={2019}
}

@misc{HGCL,
      title={Enhancing Hyperbolic Graph Embeddings via Contrastive Learning}, 
      author={Jiahong Liu and Menglin Yang and Min Zhou and Shanshan Feng and Philippe Fournier-Viger},
      year={2022},
      eprint={2201.08554},
      archivePrefix={arXiv},
      primaryClass={cs.LG},
      url={https://arxiv.org/abs/2201.08554}, 
}

@inproceedings{HyperGCL,
  title={Understanding and mitigating hyperbolic dimensional collapse in graph contrastive learning},
  author={Zhang, Yifei and Zhu, Hao and Yang, Menglin and Liu, Jiahong and Ying, Rex and King, Irwin and Koniusz, Piotr},
  booktitle={Proceedings of the 31st ACM SIGKDD Conference on Knowledge Discovery and Data Mining V. 1},
  pages={1984--1995},
  year={2025}
}

@inproceedings{TaxoAdapt,
    title = "{T}axo{A}dapt: Aligning {LLM}-Based Multidimensional Taxonomy Construction to Evolving Research Corpora",
    author = "Kargupta, Priyanka  and
      Zhang, Nan  and
      Zhang, Yunyi  and
      Zhang, Rui  and
      Mitra, Prasenjit  and
      Han, Jiawei",
    booktitle = "Proceedings of the 63rd Annual Meeting of the Association for Computational Linguistics (Volume 1: Long Papers)",
    month = jul,
    year = "2025",
    address = "Vienna, Austria",
    publisher = "Association for Computational Linguistics",
    url = "https://aclanthology.org/2025.acl-long.1442/",
    doi = "10.18653/v1/2025.acl-long.1442",
    pages = "29834--29850",
    ISBN = "979-8-89176-251-0"
}

@inproceedings{TaxoBench,
    title = "Context-Aware Hierarchical Taxonomy Generation for Scientific Papers via {LLM}-Guided Multi-Aspect Clustering",
    author = "Zhu, Kun  and
      Liao, Lizi  and
      Gu, Yuxuan  and
      Huang, Lei  and
      Feng, Xiaocheng  and
      Qin, Bing",
    booktitle = "Proceedings of the 2025 Conference on Empirical Methods in Natural Language Processing",
    month = nov,
    year = "2025",
    address = "Suzhou, China",
    publisher = "Association for Computational Linguistics",
    url = "https://aclanthology.org/2025.emnlp-main.788/",
    doi = "10.18653/v1/2025.emnlp-main.788",
    pages = "15627--15645",
    ISBN = "979-8-89176-332-6"
}

@inproceedings{MARK,
    title = "{MARK}: Multi-agent Collaboration with Ranking Guidance for Text-attributed Graph Clustering",
    author = "Fu, Yiwei  and
      Zhang, Yuxing  and
      Chen, Chunchun  and
      Ma, Jianwen  and
      Yuan, Quan  and
      Tu, Rong-Cheng  and
      Huang, Xinli  and
      Ye, Wei  and
      Luo, Xiao  and
      Deng, Minghua",
    booktitle = "Findings of the Association for Computational Linguistics: ACL 2025",
    month = jul,
    year = "2025",
    address = "Vienna, Austria",
    publisher = "Association for Computational Linguistics",
    url = "https://aclanthology.org/2025.findings-acl.314/",
    doi = "10.18653/v1/2025.findings-acl.314",
    pages = "6057--6072",
    ISBN = "979-8-89176-256-5"
}

@inproceedings{HyboNet,
    title = "Fully Hyperbolic Neural Networks",
    author = "Chen, Weize  and
      Han, Xu  and
      Lin, Yankai  and
      Zhao, Hexu  and
      Liu, Zhiyuan  and
      Li, Peng  and
      Sun, Maosong  and
      Zhou, Jie",
    booktitle = "Proceedings of the 60th Annual Meeting of the Association for Computational Linguistics (Volume 1: Long Papers)",
    month = may,
    year = "2022",
    address = "Dublin, Ireland",
    publisher = "Association for Computational Linguistics",
    url = "https://aclanthology.org/2022.acl-long.389/",
    doi = "10.18653/v1/2022.acl-long.389",
    pages = "5672--5686"
}

@inproceedings{HTCLLM,
    title = "Improving Hierarchical Text Clustering with {LLM}-guided Multi-view Cluster Representation",
    author = "Pattnaik, Anup  and
      George, Cijo  and
      Tripathi, Rishabh Kumar  and
      Vutla, Sasanka  and
      Vepa, Jithendra",
    booktitle = "Proceedings of the 2024 Conference on Empirical Methods in Natural Language Processing: Industry Track",
    month = nov,
    year = "2024",
    publisher = "Association for Computational Linguistics",
    url = "https://aclanthology.org/2024.emnlp-industry.54/",
    doi = "10.18653/v1/2024.emnlp-industry.54",
    pages = "719--727"
}

@inproceedings{TaxoGen,
  title={Taxogen: Unsupervised topic taxonomy construction by adaptive term embedding and clustering},
  author={Zhang, Chao and Tao, Fangbo and Chen, Xiusi and Shen, Jiaming and Jiang, Meng and Sadler, Brian and Vanni, Michelle and Han, Jiawei},
  booktitle={Proceedings of the 24th ACM SIGKDD International Conference on Knowledge Discovery \& Data Mining},
  pages={2701--2709},
  year={2018}
}

@inproceedings{CHIME,
    title = "{CHIME}: {LLM}-Assisted Hierarchical Organization of Scientific Studies for Literature Review Support",
    author = "Hsu, Chao-Chun  and
      Bransom, Erin  and
      Sparks, Jenna  and
      Kuehl, Bailey  and
      Tan, Chenhao  and
      Wadden, David  and
      Wang, Lucy  and
      Naik, Aakanksha",
    booktitle = "Findings of the Association for Computational Linguistics: ACL 2024",
    month = aug,
    year = "2024",
    address = "Bangkok, Thailand",
    publisher = "Association for Computational Linguistics",
    url = "https://aclanthology.org/2024.findings-acl.8/",
    doi = "10.18653/v1/2024.findings-acl.8",
    pages = "118--132"
}

@inproceedings{TELEClass,
  title={Teleclass: Taxonomy enrichment and llm-enhanced hierarchical text classification with minimal supervision},
  author={Zhang, Yunyi and Yang, Ruozhen and Xu, Xueqiang and Li, Rui and Xiao, Jinfeng and Shen, Jiaming and Han, Jiawei},
  booktitle={Proceedings of the ACM on Web Conference 2025},
  pages={2032--2042},
  year={2025}
}

@inproceedings{HGCLR,
    title = "Incorporating Hierarchy into Text Encoder: a Contrastive Learning Approach for Hierarchical Text Classification",
    author = "Wang, Zihan  and
      Wang, Peiyi  and
      Huang, Lianzhe  and
      Sun, Xin  and
      Wang, Houfeng",
    booktitle = "Proceedings of the 60th Annual Meeting of the Association for Computational Linguistics (Volume 1: Long Papers)",
    month = may,
    year = "2022",
    address = "Dublin, Ireland",
    publisher = "Association for Computational Linguistics",
    url = "https://aclanthology.org/2022.acl-long.491/",
    doi = "10.18653/v1/2022.acl-long.491",
    pages = "7109--7119"
}

@inproceedings{HMCL,
    title = "{HMCL}: Task-Optimal Text Representation Adaptation through Hierarchical Contrastive Learning",
    author = "Wang, Zhenyi  and
      Jia, Yapeng  and
      Ning, Haiyan  and
      Wang, Peng  and
      Wang, Dan  and
      Cao, Yitao",
    booktitle = "Findings of the Association for Computational Linguistics: EMNLP 2025",
    month = nov,
    year = "2025",
    address = "Suzhou, China",
    publisher = "Association for Computational Linguistics",
    url = "https://aclanthology.org/2025.findings-emnlp.727/",
    doi = "10.18653/v1/2025.findings-emnlp.727",
    pages = "13495--13518",
    ISBN = "979-8-89176-335-7"
}

@inproceedings{HiTs,
title={Language Models as Hierarchy Encoders},
author={Yuan He and Moy Yuan and Jiaoyan Chen and Ian Horrocks},
booktitle={The Thirty-eighth Annual Conference on Neural Information Processing Systems},
year={2024},
url={https://openreview.net/forum?id=GJMYvWzjE1}
}

@inproceedings{RAPTOR,
title={{RAPTOR}: Recursive Abstractive Processing for Tree-Organized Retrieval},
author={Parth Sarthi and Salman Abdullah and Aditi Tuli and Shubh Khanna and Anna Goldie and Christopher D Manning},
booktitle={The Twelfth International Conference on Learning Representations},
year={2024},
url={https://openreview.net/forum?id=GN921JHCRw}
}

@inproceedings{SciHTC,
    title = "Hierarchical Multi-Label Classification of Scientific Documents",
    author = "Sadat, Mobashir  and
      Caragea, Cornelia",
    booktitle = "Proceedings of the 2022 Conference on Empirical Methods in Natural Language Processing",
    month = dec,
    year = "2022",
    publisher = "Association for Computational Linguistics",
    url = "https://aclanthology.org/2022.emnlp-main.610/",
    doi = "10.18653/v1/2022.emnlp-main.610",
    pages = "8923--8937"
}

@article{NS4GC,
  title={Reliable node similarity matrix guided contrastive graph clustering},
  author={Liu, Yunhui and Gao, Xinyi and He, Tieke and Zheng, Tao and Zhao, Jianhua and Yin, Hongzhi},
  journal={IEEE Transactions on Knowledge and Data Engineering},
  year={2024},
  publisher={IEEE}
}

@article{DMoN,
  author  = {Tsitsulin, Anton and Palowitch, John and Perozzi, Bryan and M{\"u}ller, Emmanuel},
  title   = {Graph Clustering with Graph Neural Networks},
  journal = {Journal of Machine Learning Research},
  year    = {2023},
  volume  = {24},
  number  = {127},
  pages   = {1--21},
  url     = {http://jmlr.org/papers/v24/20-998.html}
}

@inproceedings{MAGI,
  title={Revisiting modularity maximization for graph clustering: A contrastive learning perspective},
  author={Liu, Yunfei and Li, Jintang and Chen, Yuehe and Wu, Ruofan and Wang, Ericbk and Zhou, Jing and Tian, Sheng and Shen, Shuheng and Fu, Xing and Meng, Changhua and others},
  booktitle={Proceedings of the 30th ACM SIGKDD Conference on Knowledge Discovery and Data Mining},
  pages={1968--1979},
  year={2024}
}

@inproceedings{GCLR,
title={Large Language Model Guided Graph Clustering},
author={Puja Trivedi and Nurendra Choudhary and Edward W Huang and Vassilis N. Ioannidis and Karthik Subbian and Danai Koutra},
booktitle={The Third Learning on Graphs Conference},
year={2024},
url={https://openreview.net/forum?id=CLyhlb5DG5}
}

@inproceedings{BLNN,
  title={Bootstrap latents of nodes and neighbors for graph self-supervised learning},
  author={Liu, Yunhui and Zhang, Huaisong and He, Tieke and Zheng, Tao and Zhao, Jianhua},
  booktitle={Joint European Conference on Machine Learning and Knowledge Discovery in Databases},
  pages={76--92},
  year={2024},
  organization={Springer}
}

@article{SSGE,
  title={Negative-Free Self-Supervised Gaussian Embedding of Graphs},
  author={Liu, Yunhui and He, Tieke and Zheng, Tao and Zhao, Jianhua},
  journal={Neural Networks},
  year={2024}
}

@misc{VGAE,
      title={Variational Graph Auto-Encoders}, 
      author={Thomas N. Kipf and Max Welling},
      year={2016},
      eprint={1611.07308},
      archivePrefix={arXiv},
      primaryClass={stat.ML},
      url={https://arxiv.org/abs/1611.07308}, 
}

@article{Homophily,
  title={Birds of a feather: Homophily in social networks},
  author={McPherson, Miller and Smith-Lovin, Lynn and Cook, James M},
  journal={Annual review of sociology},
  volume={27},
  number={1},
  pages={415--444},
  year={2001},
  publisher={Annual Reviews 4139 El Camino Way, PO Box 10139, Palo Alto, CA 94303-0139, USA}
}

@article{SBM,
  title={Community detection and stochastic block models: recent developments},
  author={Abbe, Emmanuel},
  journal={Journal of Machine Learning Research},
  volume={18},
  number={177},
  pages={1--86},
  year={2018}
}

@misc{DeepSeekV3,
      title={DeepSeek-V3 Technical Report}, 
      author={DeepSeek-AI and Aixin Liu and Bei Feng and Bing Xue and Bingxuan Wang and Bochao Wu and Chengda Lu and Chenggang Zhao and Chengqi Deng and Chenyu Zhang and Chong Ruan and Damai Dai and Daya Guo and Dejian Yang and Deli Chen and Dongjie Ji and Erhang Li and Fangyun Lin and Fucong Dai and Fuli Luo and Guangbo Hao and Guanting Chen and Guowei Li and H. Zhang and Han Bao and Hanwei Xu and Haocheng Wang and Haowei Zhang and Honghui Ding and Huajian Xin and Huazuo Gao and Hui Li and Hui Qu and J. L. Cai and Jian Liang and Jianzhong Guo and Jiaqi Ni and Jiashi Li and Jiawei Wang and Jin Chen and Jingchang Chen and Jingyang Yuan and Junjie Qiu and Junlong Li and Junxiao Song and Kai Dong and Kai Hu and Kaige Gao and Kang Guan and Kexin Huang and Kuai Yu and Lean Wang and Lecong Zhang and Lei Xu and Leyi Xia and Liang Zhao and Litong Wang and Liyue Zhang and Meng Li and Miaojun Wang and Mingchuan Zhang and Minghua Zhang and Minghui Tang and Mingming Li and Ning Tian and Panpan Huang and Peiyi Wang and Peng Zhang and Qiancheng Wang and Qihao Zhu and Qinyu Chen and Qiushi Du and R. J. Chen and R. L. Jin and Ruiqi Ge and Ruisong Zhang and Ruizhe Pan and Runji Wang and Runxin Xu and Ruoyu Zhang and Ruyi Chen and S. S. Li and Shanghao Lu and Shangyan Zhou and Shanhuang Chen and Shaoqing Wu and Shengfeng Ye and Shengfeng Ye and Shirong Ma and Shiyu Wang and Shuang Zhou and Shuiping Yu and Shunfeng Zhou and Shuting Pan and T. Wang and Tao Yun and Tian Pei and Tianyu Sun and W. L. Xiao and Wangding Zeng and Wanjia Zhao and Wei An and Wen Liu and Wenfeng Liang and Wenjun Gao and Wenqin Yu and Wentao Zhang and X. Q. Li and Xiangyue Jin and Xianzu Wang and Xiao Bi and Xiaodong Liu and Xiaohan Wang and Xiaojin Shen and Xiaokang Chen and Xiaokang Zhang and Xiaosha Chen and Xiaotao Nie and Xiaowen Sun and Xiaoxiang Wang and Xin Cheng and Xin Liu and Xin Xie and Xingchao Liu and Xingkai Yu and Xinnan Song and Xinxia Shan and Xinyi Zhou and Xinyu Yang and Xinyuan Li and Xuecheng Su and Xuheng Lin and Y. K. Li and Y. Q. Wang and Y. X. Wei and Y. X. Zhu and Yang Zhang and Yanhong Xu and Yanhong Xu and Yanping Huang and Yao Li and Yao Zhao and Yaofeng Sun and Yaohui Li and Yaohui Wang and Yi Yu and Yi Zheng and Yichao Zhang and Yifan Shi and Yiliang Xiong and Ying He and Ying Tang and Yishi Piao and Yisong Wang and Yixuan Tan and Yiyang Ma and Yiyuan Liu and Yongqiang Guo and Yu Wu and Yuan Ou and Yuchen Zhu and Yuduan Wang and Yue Gong and Yuheng Zou and Yujia He and Yukun Zha and Yunfan Xiong and Yunxian Ma and Yuting Yan and Yuxiang Luo and Yuxiang You and Yuxuan Liu and Yuyang Zhou and Z. F. Wu and Z. Z. Ren and Zehui Ren and Zhangli Sha and Zhe Fu and Zhean Xu and Zhen Huang and Zhen Zhang and Zhenda Xie and Zhengyan Zhang and Zhewen Hao and Zhibin Gou and Zhicheng Ma and Zhigang Yan and Zhihong Shao and Zhipeng Xu and Zhiyu Wu and Zhongyu Zhang and Zhuoshu Li and Zihui Gu and Zijia Zhu and Zijun Liu and Zilin Li and Ziwei Xie and Ziyang Song and Ziyi Gao and Zizheng Pan},
      year={2025},
      eprint={2412.19437},
      archivePrefix={arXiv},
      primaryClass={cs.CL},
      url={https://arxiv.org/abs/2412.19437}, 
}

@misc{Qwen3,
      title={Qwen3 Technical Report}, 
      author={An Yang and Anfeng Li and Baosong Yang and Beichen Zhang and Binyuan Hui and Bo Zheng and Bowen Yu and Chang Gao and Chengen Huang and Chenxu Lv and Chujie Zheng and Dayiheng Liu and Fan Zhou and Fei Huang and Feng Hu and Hao Ge and Haoran Wei and Huan Lin and Jialong Tang and Jian Yang and Jianhong Tu and Jianwei Zhang and Jianxin Yang and Jiaxi Yang and Jing Zhou and Jingren Zhou and Junyang Lin and Kai Dang and Keqin Bao and Kexin Yang and Le Yu and Lianghao Deng and Mei Li and Mingfeng Xue and Mingze Li and Pei Zhang and Peng Wang and Qin Zhu and Rui Men and Ruize Gao and Shixuan Liu and Shuang Luo and Tianhao Li and Tianyi Tang and Wenbiao Yin and Xingzhang Ren and Xinyu Wang and Xinyu Zhang and Xuancheng Ren and Yang Fan and Yang Su and Yichang Zhang and Yinger Zhang and Yu Wan and Yuqiong Liu and Zekun Wang and Zeyu Cui and Zhenru Zhang and Zhipeng Zhou and Zihan Qiu},
      year={2025},
      eprint={2505.09388},
      archivePrefix={arXiv},
      primaryClass={cs.CL},
      url={https://arxiv.org/abs/2505.09388}, 
}

@inproceedings{APF,
  title={Towards anomaly-aware pre-training and fine-tuning for graph anomaly detection},
  author={Liu, Yunhui and Cheng, Jiashun and Lin, Yiqing and Xie, Qizhuo and Li, Jia and Tsung, Fugee and Yin, Hongzhi and Zheng, Tao and Zhao, Jianhua and He, Tieke},
  booktitle={International Conference on Learning Representations},
  volume={2026},
  pages={136630--136663},
  year={2026}
}

@article{LLM4Graph3,
  title={Exploring the Potential of Large Language Models (LLMs) in Learning on Graphs},
  author={Chen, Zhikai and Mao, Haitao and Li, Hang and Jin, Wei and Wen, Hongzhi and Wei, Xiaochi and Wang, Shuaiqiang and Yin, Dawei and Fan, Wenqi and Liu, Hui and others},
  journal={ACM SIGKDD Explorations Newsletter},
  volume={25},
  number={2},
  pages={42--61},
  year={2024}
}

@inproceedings{LLMNodeBed,
title={When Do {LLM}s Help With Node Classification? A Comprehensive Analysis},
author={Xixi Wu and Yifei Shen and Fangzhou Ge and Caihua Shan and Yizhu Jiao and Xiangguo Sun and Hong Cheng},
booktitle={Forty-second International Conference on Machine Learning},
year={2025}
}

@misc{WikiCS,
  title={Wiki-CS: A Wikipedia-Based Benchmark for Graph Neural Networks}, 
  author={Péter Mernyei and Cătălina Cangea},
  year={2022},
  eprint={2007.02901},
  archivePrefix={arXiv},
  primaryClass={cs.LG},
  url={https://arxiv.org/abs/2007.02901}, 
}

@article{OGB,
  title={Open graph benchmark: Datasets for machine learning on graphs},
  author={Hu, Weihua and Fey, Matthias and Zitnik, Marinka and Dong, Yuxiao and Ren, Hongyu and Liu, Bowen and Catasta, Michele and Leskovec, Jure},
  journal={Advances in neural information processing systems},
  volume={33},
  pages={22118--22133},
  year={2020}
}

@inproceedings{Amazon,
    title = "Justifying Recommendations using Distantly-Labeled Reviews and Fine-Grained Aspects",
    author = "Ni, Jianmo  and
      Li, Jiacheng  and
      McAuley, Julian",
    booktitle = "Proceedings of the 2019 Conference on Empirical Methods in Natural Language Processing and the 9th International Joint Conference on Natural Language Processing (EMNLP-IJCNLP)",
    month = nov,
    year = "2019",
    address = "Hong Kong, China",
    publisher = "Association for Computational Linguistics",
    url = "https://aclanthology.org/D19-1018/",
    doi = "10.18653/v1/D19-1018",
    pages = "188--197"
}

@inproceedings{LLaGA,
title={{LL}a{GA}: Large Language and Graph Assistant},
author={Runjin Chen and Tong Zhao and AJAY KUMAR JAISWAL and Neil Shah and Zhangyang Wang},
booktitle={Forty-first International Conference on Machine Learning},
year={2024},
url={https://openreview.net/forum?id=B48Pzc4oKi}
}

@article{CS-TAG,
  title={A comprehensive study on text-attributed graphs: Benchmarking and rethinking},
  author={Yan, Hao and Li, Chaozhuo and Long, Ruosong and Yan, Chao and Zhao, Jianan and Zhuang, Wenwen and Yin, Jun and Zhang, Peiyan and Han, Weihao and Sun, Hao and others},
  journal={Advances in Neural Information Processing Systems},
  volume={36},
  pages={17238--17264},
  year={2023}
}

@inproceedings{GIANT,
    title={Node Feature Extraction by Self-Supervised Multi-scale Neighborhood Prediction},
    author={Eli Chien and Wei-Cheng Chang and Cho-Jui Hsieh and Hsiang-Fu Yu and Jiong Zhang and Olgica Milenkovic and Inderjit S Dhillon},
    booktitle={International Conference on Learning Representations},
    year={2022}
}

@inproceedings{GLEM,
    title={Learning on Large-scale Text-attributed Graphs via Variational Inference},
    author={Jianan Zhao and Meng Qu and Chaozhuo Li and Hao Yan and Qian Liu and Rui Li and Xing Xie and Jian Tang},
    booktitle={The Eleventh International Conference on Learning Representations },
    year={2023}
}

@inproceedings{GraphNarrator,
    title = "{G}raph{N}arrator: Generating Textual Explanations for Graph Neural Networks",
    author = "Pan, Bo  and
      Xiong, Zhen  and
      Wu, Guanchen  and
      Zhang, Zheng  and
      Zhang, Yifei  and
      Hu, Yuntong  and
      Zhao, Liang",
    booktitle = "Proceedings of the 63rd Annual Meeting of the Association for Computational Linguistics (Volume 1: Long Papers)",
    month = jul,
    year = "2025",
    address = "Vienna, Austria",
    publisher = "Association for Computational Linguistics",
    url = "https://aclanthology.org/2025.acl-long.2/",
    doi = "10.18653/v1/2025.acl-long.2",
    pages = "23--42",
    ISBN = "979-8-89176-251-0"
}

@inproceedings{LensGNN,
    title = "Can Large Language Models Act as Ensembler for Multi-{GNN}s?",
    author = "Duan, Hanqi  and
      Cheng, Yao  and
      Yu, Jianxiang  and
      Liu, Yao  and
      Li, Xiang",
    booktitle = "Proceedings of the 2025 Conference on Empirical Methods in Natural Language Processing",
    month = nov,
    year = "2025",
    address = "Suzhou, China",
    publisher = "Association for Computational Linguistics",
    url = "https://aclanthology.org/2025.emnlp-main.1470/",
    doi = "10.18653/v1/2025.emnlp-main.1470",
    pages = "28863--28882",
    ISBN = "979-8-89176-332-6"
}

@inproceedings{TAPE,
title={Harnessing Explanations: {LLM}-to-{LM} Interpreter for Enhanced Text-Attributed Graph Representation Learning},
author={Xiaoxin He and Xavier Bresson and Thomas Laurent and Adam Perold and Yann LeCun and Bryan Hooi},
booktitle={The Twelfth International Conference on Learning Representations},
year={2024},
url={https://openreview.net/forum?id=RXFVcynVe1}
}

@inproceedings{GAugLLM,
  title={{GAugLLM}: Improving Graph Contrastive Learning for Text-Attributed Graphs with Large Language Models},
  author={Fang, Yi and Fan, Dongzhe and Zha, Daochen and Tan, Qiaoyu},
  booktitle={Proceedings of the 30th ACM SIGKDD Conference on Knowledge Discovery and Data Mining},
  pages={747--758},
  year={2024}
}

@inproceedings{GraphGPT,
  title={GraphGPT: Graph instruction tuning for large language models},
  author={Tang, Jiabin and Yang, Yuhao and Wei, Wei and Shi, Lei and Su, Lixin and Cheng, Suqi and Yin, Dawei and Huang, Chao},
  booktitle={Proceedings of the 47th International ACM SIGIR Conference on Research and Development in Information Retrieval},
  pages={491--500},
  year={2024}
}

@inproceedings{ENGINE,
  title={Efficient tuning and inference for large language models on textual graphs},
  author={Zhu, Yun and Wang, Yaoke and Shi, Haizhou and Tang, Siliang},
  booktitle={Proceedings of the Thirty-Third International Joint Conference on Artificial Intelligence},
  pages={5734--5742},
  year={2024}
}

@inproceedings{SKETCH,
    title = "Taming Language Models for Text-attributed Graph Learning with Decoupled Aggregation",
    author = "Zhou, Chuang  and
      Wang, Zhu  and
      Chen, Shengyuan  and
      Du, Jiahe  and
      Zheng, Qiyuan  and
      Xu, Zhaozhuo  and
      Huang, Xiao",
    booktitle = "Proceedings of the 63rd Annual Meeting of the Association for Computational Linguistics (Volume 1: Long Papers)",
    month = jul,
    year = "2025",
    publisher = "Association for Computational Linguistics",
    url = "https://aclanthology.org/2025.acl-long.173/",
    doi = "10.18653/v1/2025.acl-long.173",
    pages = "3463--3474",
    ISBN = "979-8-89176-251-0"
}

@inproceedings{GRENADE,
    title = "{GRENADE}: Graph-Centric Language Model for Self-Supervised Representation Learning on Text-Attributed Graphs",
    author = "Li, Yichuan  and
      Ding, Kaize  and
      Lee, Kyumin",
    booktitle = "Findings of the Association for Computational Linguistics: EMNLP 2023",
    month = dec,
    year = "2023",
    address = "Singapore",
    publisher = "Association for Computational Linguistics",
    url = "https://aclanthology.org/2023.findings-emnlp.181/",
    doi = "10.18653/v1/2023.findings-emnlp.181",
    pages = "2745--2757"
}

@inproceedings{OFA,
title={One For All: Towards Training One Graph Model For All Classification Tasks},
author={Hao Liu and Jiarui Feng and Lecheng Kong and Ningyue Liang and Dacheng Tao and Yixin Chen and Muhan Zhang},
booktitle={The Twelfth International Conference on Learning Representations},
year={2024}
}

@inproceedings{SentenceBERT,
    title = "Sentence-{BERT}: Sentence Embeddings using {S}iamese {BERT}-Networks",
    author = "Reimers, Nils  and
      Gurevych, Iryna",
    booktitle = "Proceedings of the 2019 Conference on Empirical Methods in Natural Language Processing and the 9th International Joint Conference on Natural Language Processing (EMNLP-IJCNLP)",
    month = nov,
    year = "2019",
    address = "Hong Kong, China",
    publisher = "Association for Computational Linguistics",
    url = "https://aclanthology.org/D19-1410/",
    doi = "10.18653/v1/D19-1410",
    pages = "3982--3992"
}

@article{RoBERTa,
  title={RoBERTa: A robustly optimized bert pretraining approach},
  author={Liu, Yinhan and Ott, Myle and Goyal, Naman and Du, Jingfei and Joshi, Mandar and Chen, Danqi and Levy, Omer and Lewis, Mike and Zettlemoyer, Luke and Stoyanov, Veselin},
  journal={arXiv preprint arXiv:1907.11692},
  year={2019}
}

@inproceedings{HTAG,
  title={Multi-Scale Heterogeneous Text-Attributed Graph Datasets From Diverse Domains},
  author={Liu, Yunhui and Xie, Qizhuo and Shi, Jinwei and Shen, Jiaxu and He, Tieke},
  booktitle={Companion Proceedings of the ACM on Web Conference 2025},
  pages={757--760},
  year={2025}
}

@article{CCC,
  title={The comparison of dendrograms by objective methods},
  author={Sokal, Robert R and Rohlf, F James},
  journal={Taxon},
  pages={33--40},
  year={1962},
  publisher={JSTOR}
}

@article{VEM,
  title={An introduction to variational methods for graphical models},
  author={Jordan, Michael I and Ghahramani, Zoubin and Jaakkola, Tommi S and Saul, Lawrence K},
  journal={Machine learning},
  volume={37},
  number={2},
  pages={183--233},
  year={1999},
  publisher={Springer}
}

@incollection{Hard-EM1,
  title={A view of the EM algorithm that justifies incremental, sparse, and other variants},
  author={Neal, Radford M and Hinton, Geoffrey E},
  booktitle={Learning in graphical models},
  pages={355--368},
  year={1998},
  publisher={Springer}
}

@inproceedings{LAMA,
    title = "Language Models as Knowledge Bases?",
    author = {Petroni, Fabio  and
      Rockt{\"a}schel, Tim  and
      Riedel, Sebastian  and
      Lewis, Patrick  and
      Bakhtin, Anton  and
      Wu, Yuxiang  and
      Miller, Alexander},
    booktitle = "Proceedings of the 2019 Conference on Empirical Methods in Natural Language Processing and the 9th International Joint Conference on Natural Language Processing (EMNLP-IJCNLP)",
    month = nov,
    year = "2019",
    address = "Hong Kong, China",
    publisher = "Association for Computational Linguistics",
    url = "https://aclanthology.org/D19-1250/",
    doi = "10.18653/v1/D19-1250",
    pages = "2463--2473"
}

@article{Fleiss,
  title={Measuring nominal scale agreement among many raters.},
  author={Fleiss and  Joseph, L. },
  journal={Psychological Bulletin},
  volume={76},
  number={5},
  pages={378--382},
  year={1971},
}

@article{PyAGC,
  title={Bridging academia and industry: a comprehensive benchmark for attributed graph clustering},
  author={Liu, Yunhui and Qiu, Pengyu and Xing, Yu and Liu, Yongchao and Du, Peng and Hong, Chuntao and Zheng, Jiajun and Zheng, Tao and He, Tieke},
  journal={arXiv preprint arXiv:2602.08519},
  year={2026}
}

@article{OFAG,
  title={Towards One-for-All Foundation Model for Attributed Graph Clustering},
  author={Liu, Yunhui and Jin, Xudong and Zhang, Kang and An, Danshuo and Xing, Yu and Song, Te and Liu, Jia and He, Tieke},
  journal={arXiv preprint arXiv:2610.07778},
  year={2026}
}

@article{AGCSurvey,
  title={Beyond the academic monoculture: a unified framework and industrial perspective for attributed graph clustering},
  author={Liu, Yunhui and Liu, Yue and Liu, Yongchao and Zheng, Tao and Li, Stan Z and Liu, Xinwang and He, Tieke},
  journal={arXiv preprint arXiv:2603.20829},
  year={2026}
}

@inproceedings{SASE,
  title={Scalable and adaptive spectral embedding for attributed graph clustering},
  author={Liu, Yunhui and He, Tieke and Wu, Qing and Zheng, Tao and Zhao, Jianhua},
  booktitle={Proceedings of the 33rd ACM international conference on information and knowledge management},
  pages={3912--3916},
  year={2024}
}

@article{HG2M,
  title={Teaching mlps to master heterogeneous graph-structured knowledge for efficient and accurate inference},
  author={Liu, Yunhui and Gao, Xinyi and He, Tieke and Zhao, Jianhua and Yin, Hongzhi},
  journal={IEEE Transactions on Knowledge and Data Engineering},
  volume={37},
  number={10},
  pages={6189--6201},
  year={2025},
  publisher={IEEE}
}

@inproceedings{GS-Quant,
  title={GS-Quant: Granular Semantic and Generative Structural Quantization for Knowledge Graph Completion},
  author={Xie, Qizhuo and Liu, Yunhui and Xing, Yu and Hou, Qianzi and Jin, Xudong and Zheng, Tao and He, Tieke},
  booktitle={Proceedings of the 64th Annual Meeting of the Association for Computational Linguistics (Volume 1: Long Papers)},
  pages={16782--16797},
  year={2026}
}

@inproceedings{TIER,
  title={Learning hierarchical knowledge in text-rich networks with taxonomy-informed representation learning},
  author={Liu, Yunhui and Liu, Yongchao and Chen, Yinfeng and Hong, Chuntao and Zheng, Tao and He, Tieke},
  booktitle={Proceedings of the 32nd ACM SIGKDD Conference on Knowledge Discovery and Data Mining V. 1},
  pages={961--972},
  year={2026}
}

@inproceedings{Kiritchenko2006,
  title={Learning and evaluation in the presence of class hierarchies: Application to text categorization},
  author={Kiritchenko, Svetlana and Matwin, Stan and Nock, Richard and Famili, A Fazel},
  booktitle={Conference of the Canadian society for computational studies of intelligence},
  pages={395--406},
  year={2006},
  organization={Springer}
}

@article{Sokal1962,
  title={The comparison of dendrograms by objective methods},
  author={Sokal, Robert R and Rohlf, F James},
  journal={Taxon},
  volume={11},
  number={2},
  pages={33--40},
  year={1962},
  publisher={JSTOR}
}


\appendix

\clearpage
\section*{Appendix Table of Contents}
\label{sec:appendix_toc}

To facilitate navigation, we organize the supplementary materials as follows:

\begin{itemize}[leftmargin=1.5em, itemsep=0.2em]
    \item \textbf{Appendix~\ref{sec:theory_motivation_appendix}: Theoretical Motivation} \\
    \textit{Formal derivation of the information-theoretic superiority of hierarchy-aware embeddings.}
    
    \item \textbf{Appendix~\ref{sec:theory}: Theoretical Analysis} \\
    \textit{Formal derivation of \texttt{SHiFT} as a generalized Variational EM algorithm.}
    
    \item \textbf{Appendix~\ref{app:setup_details}: Detailed Experimental Setup} \\
    \textit{Dataset statistics, baseline descriptions, and implementation details.}
    
    \item \textbf{Appendix~\ref{app:prompts}: List of LLM Prompts} \\
    \textit{Full templates for taxonomy induction, refinement, and diagnosis.}
    
    \item \textbf{Appendix~\ref{app:taxo_eval_protocols}: Taxonomy Evaluation Protocols} \\
    \textit{Methodologies for calculating Concept Coherence (NPMI) and feature augmentation.}

    \item \textbf{Appendix~\ref{app:patent}: Evaluation on a Patent Graph with Verifiable Hierarchy} \\
    \textit{Hierarchical classification and CPC-tree recovery on an examiner-labeled patent graph.}
    
    \item \textbf{Appendix~\ref{app:efficiency}: Efficiency and Scalability Analysis} \\
    \textit{Runtime profiling and token cost analysis on large-scale graphs.}
    
    \item \textbf{Appendix~\ref{app:llm_sensitivity}: Impact of LLM Selection} \\
    \textit{Ablation studies comparing DeepSeek-V3, GPT-4, and LLaMA-3.}
    
    \item \textbf{Appendix~\ref{app:hyperparameter}: Hyperparameter Sensitivity} \\
    \textit{Analysis of taxonomy depth limits and loss balancing weights.}
    
    \item \textbf{Appendix~\ref{app:visualization_analysis}: Visual Analysis of Representations} \\
    \textit{Heatmaps of embedding similarity and geometric alignment.}
    
    \item \textbf{Appendix~\ref{app:qualitative_analysis}: Qualitative Analysis of Taxonomies} \\
    \textit{Case studies comparing PLM-only vs. \texttt{SHiFT}-induced hierarchies.}
\end{itemize}

\section{Theoretical Motivation: Proofs and Derivations}
\label{sec:theory_motivation_appendix}

We provide formal motivation for why hierarchy-aware representation learning is necessary for text-attributed graphs.
A subtle but important point is that the semantic hierarchy $\mathcal{T}^*$ is a global latent structure rather than a per-sample random label.
Therefore, instead of measuring $I(\boldsymbol{Z};\mathcal{T}^*)$ directly, we quantify whether the representation preserves the \emph{hierarchical relations induced by} $\mathcal{T}^*$ among samples.
This leads to a pairwise formulation based on tree distances between latent leaf concepts.

\subsection{Hierarchical Data-Generating Process}
\label{app:hdgp}

\begin{definition}[Hierarchical Data-Generating Process]
\label{def:hdgp}
Let $\mathcal{T}^*=(\mathcal{C},\mathcal{R})$ be a rooted tree with depth $L$ and leaf set 
$\mathcal{Y}=\{1,\dots,K\}$.
For a leaf concept $y\in\mathcal{Y}$, let 
$\operatorname{anc}_{\ell}(y)$ denote its ancestor at level $\ell$, where $\ell=0$ is the root and $\ell=L$ is the leaf itself.
The generative process is:
\begin{enumerate}[leftmargin=1.5em, itemsep=0.2em]
    \item \textbf{Hierarchy:} A latent semantic hierarchy $\mathcal{T}^*$ is drawn from a prior $p(\mathcal{T}^*)$.
    \item \textbf{Concept assignment:} For each data point $x_i$, a leaf concept 
    $Y_i\sim \mathrm{Cat}(\boldsymbol{\pi})$ is sampled, where $\boldsymbol{\pi}$ is a mixing distribution over leaves.
    Its level-$\ell$ semantic label is 
    $Y_i^{(\ell)}=\operatorname{anc}_{\ell}(Y_i)$.
    \item \textbf{Observation generation:} The observed text and graph signal are generated from a conditional distribution 
    $p(x_i \mid Y_i,\mathcal{T}^*)$, whose semantics are determined by the full root-to-leaf path 
    $\{Y_i^{(0)},\dots,Y_i^{(L)}\}$.
\end{enumerate}
For two independently sampled data points $(x_i,x_j)$, define their latent tree distance as $D^*_{ij} = d_{\mathcal{T}^*}(Y_i,Y_j)$,
where $d_{\mathcal{T}^*}(\cdot,\cdot)$ is the shortest-path distance between two leaf concepts in $\mathcal{T}^*$.
Equivalently, $D^*_{ij}$ encodes the depth of the lowest common ancestor between $Y_i$ and $Y_j$.
\end{definition}

The variable $D^*_{ij}$ is the key object of interest: it captures whether two samples are siblings, cousins, or semantically distant concepts.
Thus, a hierarchy-aware representation should not merely separate leaf clusters; it should preserve these pairwise semantic relations.

\subsection{Hierarchy-Preserving Representations}
\label{app:hier_preserving_rep}

Let $f_{\theta}$ be a graph encoder and $\boldsymbol{Z}=f_{\theta}(\mathcal{G},\boldsymbol{X})$ be the learned node embeddings.
For a node whose latent leaf concept is $Y$, write its embedding as $\boldsymbol{z}$.
We measure the amount of hierarchy preserved by $f_{\theta}$ through the mutual information between the latent tree distance and the embedding-pair statistic:
\begin{equation}
    \mathcal{I}_{\mathrm{hier}}(f_{\theta})
    =
    I\!\left(
        D^*_{ij};
        \|\boldsymbol{z}_i-\boldsymbol{z}_j\|_2
        \mid \mathcal{T}^*
    \right).
    \label{eq:hier_info}
\end{equation}
This quantity is well-defined because $D^*_{ij}$ is a random variable induced by randomly sampled data points, while $\mathcal{T}^*$ is treated as the conditioning latent hierarchy.
A large value of $\mathcal{I}_{\mathrm{hier}}(f_{\theta})$ means that distances in the learned embedding space reveal the semantic distance in the underlying hierarchy.

\begin{definition}[Hierarchy-aligned prototype representation]
\label{def:hier_aligned_proto}
Let $\boldsymbol{\mu}_y\in\mathbb{R}^d$ denote the prototype embedding of leaf concept $y$.
A representation is $(\varepsilon,r,\zeta)$-hierarchy-aligned if there exists a strictly increasing function 
$g:\mathcal{D}\rightarrow\mathbb{R}_{+}$ over the finite set 
$\mathcal{D}=\{d_{\mathcal{T}^*}(y,y') : y,y'\in\mathcal{Y}\}$ such that:
\begin{enumerate}[leftmargin=1.5em, itemsep=0.2em]
    \item \textbf{Prototype tree-distance alignment:}
    \begin{equation}
        \left|
        \|\boldsymbol{\mu}_y-\boldsymbol{\mu}_{y'}\|_2
        -
        g\!\left(d_{\mathcal{T}^*}(y,y')\right)
        \right|
        \leq \varepsilon,
        \quad \forall y,y'\in\mathcal{Y}.
        \label{eq:proto_alignment}
    \end{equation}
    \item \textbf{Within-concept concentration:}
    for a sample embedding $\boldsymbol{z}$ with latent leaf $Y=y$,
    \begin{equation}
        \Pr\!\left(
        \|\boldsymbol{z}-\boldsymbol{\mu}_y\|_2 \leq r
        \mid \mathcal{T}^*
        \right)
        \geq 1-\zeta.
        \label{eq:within_concentration}
    \end{equation}
\end{enumerate}
Let 
\begin{equation}
    \Delta_g
    =
    \min_{d\neq d'\in\mathcal{D}}
    |g(d)-g(d')|
\end{equation}
be the minimum separation between distinct tree-distance levels after transformation by $g$.
\end{definition}

\begin{theorem}[Hierarchy alignment preserves tree-distance information]
\label{thm:hier_info_lower_bound}
Suppose a representation is $(\varepsilon,r,\zeta)$-hierarchy-aligned and satisfies
\begin{equation}
    \varepsilon + 2r < \frac{\Delta_g}{2}.
    \label{eq:margin_condition}
\end{equation}
Then the latent tree distance $D^*_{ij}$ can be recovered from the embedding distance 
$\|\boldsymbol{z}_i-\boldsymbol{z}_j\|_2$ with error probability at most $2\zeta$.
Consequently,
\begin{equation}
    \mathcal{I}_{\mathrm{hier}}(f_{\theta})
    \geq
    H(D^*_{ij}\mid \mathcal{T}^*)
    -
    h_2(2\zeta)
    -
    2\zeta\log(|\mathcal{D}|-1),
    \label{eq:hier_info_lower_bound}
\end{equation}
where $h_2(\cdot)$ is the binary entropy function.
In particular, when $\zeta=0$, the representation preserves the full tree-distance information:
\begin{equation}
    \mathcal{I}_{\mathrm{hier}}(f_{\theta})
    =
    H(D^*_{ij}\mid \mathcal{T}^*).
\end{equation}
\end{theorem}

\begin{proof}
Consider two samples with latent leaves $Y_i=y$ and $Y_j=y'$.
When both embeddings satisfy the concentration event in Eq.~\eqref{eq:within_concentration}, the triangle inequality gives
\begin{align}
    \Big|
    \|\boldsymbol{z}_i-\boldsymbol{z}_j\|_2
    -
    \|\boldsymbol{\mu}_y-\boldsymbol{\mu}_{y'}\|_2
    \Big|
    &\leq
    \|\boldsymbol{z}_i-\boldsymbol{\mu}_y\|_2
    +
    \|\boldsymbol{z}_j-\boldsymbol{\mu}_{y'}\|_2 \notag\\
    &\leq 2r.
\end{align}
Combining this with the prototype alignment condition in Eq.~\eqref{eq:proto_alignment}, we obtain
\begin{equation}
    \Big|
    \|\boldsymbol{z}_i-\boldsymbol{z}_j\|_2
    -
    g(D^*_{ij})
    \Big|
    \leq
    \varepsilon + 2r.
\end{equation}
By the margin condition in Eq.~\eqref{eq:margin_condition}, the intervals centered at 
$\{g(d):d\in\mathcal{D}\}$ with radius $\varepsilon+2r$ are disjoint.
Therefore, the nearest-distance decoder
\begin{equation}
    \widehat{D}_{ij}
    =
    \arg\min_{d\in\mathcal{D}}
    \left|
    \|\boldsymbol{z}_i-\boldsymbol{z}_j\|_2 - g(d)
    \right|
\end{equation}
recovers $D^*_{ij}$ whenever both concentration events hold.
By the union bound, the probability that at least one of the two concentration events fails is at most $2\zeta$.
Thus,
\begin{equation}
    \Pr(\widehat{D}_{ij}\neq D^*_{ij})\leq 2\zeta.
\end{equation}

Applying Fano's inequality yields
\begin{equation}
    H(D^*_{ij}\mid \|\boldsymbol{z}_i-\boldsymbol{z}_j\|_2,\mathcal{T}^*)
    \leq
    h_2(2\zeta)
    +
    2\zeta\log(|\mathcal{D}|-1).
\end{equation}
Finally, by the definition of mutual information,
\begin{align}
    I(D^*_{ij};\|\boldsymbol{z}_i-\boldsymbol{z}_j\|_2\mid \mathcal{T}^*)
    &=
    H(D^*_{ij}\mid\mathcal{T}^*) \notag\\
    &\quad -
    H(D^*_{ij}\mid \|\boldsymbol{z}_i-\boldsymbol{z}_j\|_2,\mathcal{T}^*),
\end{align}
which gives Eq.~\eqref{eq:hier_info_lower_bound}.
When $\zeta=0$, the recovery error is zero, and the conditional entropy term vanishes.
\end{proof}

\subsection{Why Flat Partition Objectives Are Insufficient}
\label{app:flat_insufficient}

Theorem~\ref{thm:hier_info_lower_bound} shows that hierarchy-aligned embeddings preserve the latent tree-distance structure.
We next show that a flat partition-preserving representation can perfectly separate leaf concepts while still failing to encode their hierarchical relations.

\begin{theorem}[Non-identifiability of hierarchy under flat partition alignment]
\label{thm:flat_nonidentifiability}
Assume $K\geq 3$ leaf concepts.
There exists a representation that perfectly separates all leaf concepts but contains no information about positive tree distances in its embedding distances.
Specifically, there exist prototypes $\{\boldsymbol{\mu}_y\}_{y=1}^{K}$ such that
\begin{equation}
    \|\boldsymbol{\mu}_y-\boldsymbol{\mu}_{y'}\|_2 = q,
    \quad \forall y\neq y',
    \label{eq:simplex}
\end{equation}
for some constant $q>0$.
For embeddings concentrated around these prototypes, the distance statistic 
$\|\boldsymbol{z}_i-\boldsymbol{z}_j\|_2$ can distinguish whether two samples belong to the same leaf, but cannot distinguish different positive tree distances.
In the noiseless case,
\begin{equation}
    I\!\left(
    D^*_{ij};
    \|\boldsymbol{z}_i-\boldsymbol{z}_j\|_2
    \mid
    D^*_{ij}>0,\mathcal{T}^*
    \right)
    =
    0.
    \label{eq:flat_zero_info}
\end{equation}
\end{theorem}

\begin{proof}
A regular simplex in $\mathbb{R}^{K-1}$ provides $K$ prototypes 
$\{\boldsymbol{\mu}_y\}_{y=1}^{K}$ satisfying Eq.~\eqref{eq:simplex}.
Thus, every pair of distinct leaf concepts has exactly the same prototype distance $q$.
This representation separates leaves perfectly because $\boldsymbol{\mu}_y\neq \boldsymbol{\mu}_{y'}$ for all $y\neq y'$.
Therefore, any objective that only enforces leaf-level partition consistency can be minimized.

However, conditioned on $D^*_{ij}>0$, the two samples come from different leaves.
In the noiseless case, their embedding distance is always $q$, regardless of whether the leaves are siblings, cousins, or semantically distant concepts in $\mathcal{T}^*$.
Hence,
\begin{equation}
    \|\boldsymbol{z}_i-\boldsymbol{z}_j\|_2
    =
    q
    \quad
    \text{for all } D^*_{ij}>0.
\end{equation}
The distance statistic is therefore constant after conditioning on $D^*_{ij}>0$, and a constant random variable has zero mutual information with $D^*_{ij}$.
This proves Eq.~\eqref{eq:flat_zero_info}.
\end{proof}

\paragraph{Implication for \texttt{SHiFT}.}
Theorem~\ref{thm:flat_nonidentifiability} shows that flat partition alignment alone cannot identify the semantic skeleton of the hierarchy: leaf clusters may be perfectly separated while all inter-cluster relations remain geometrically indistinguishable.
Theorem~\ref{thm:hier_info_lower_bound} further shows that if embedding distances are aligned with tree distances, then the representation preserves the hierarchical relations induced by $\mathcal{T}^*$.
This motivates the three complementary objectives in the M-step of \texttt{SHiFT}: TPA enforces local partition consistency, TSA aligns global embedding geometry with the taxonomy skeleton, and TCA anchors the learned space to explicit LLM-generated semantic concepts.

\section{Theoretical Analysis: Proofs and Derivations}\label{sec:theory}
In this section, we provide a probabilistic interpretation of \texttt{SHiFT}. We demonstrate that our iterative framework constitutes a Variational Expectation-Maximization algorithm, where the proposed objectives maximize the Evidence Lower Bound of the log-likelihood of the graph structure.

\subsection{Probabilistic Generative Model}
Following the paradigm of Variational Graph Auto-Encoders~\cite{VGAE}, we posit that the observed graph adjacency $\boldsymbol{A}$ is generated from latent variables.
Let $\boldsymbol{H} = (\mathcal{T}, \psi)$ denote the latent hierarchy variables, where $\mathcal{T}$ represents the taxonomy structure (comprising tree skeleton $\mathcal{R}$ and concept definitions $\mathcal{C}$) and $\psi: \mathcal{V} \rightarrow \text{Leaves}(\mathcal{T})$ denotes the discrete node-to-concept assignment (partition).
Let $\boldsymbol{Z}$ denote the continuous latent node representations.
The generative process is defined as follows:
\begin{enumerate}[leftmargin=1.5em, itemsep=0.2em]
    \item \textbf{Taxonomy Generation:} A taxonomy $\mathcal{T}$ is drawn from a semantic prior $p(\mathcal{T})$, which governs the valid semantic organization of the domain (simulated by the LLM).
    \item \textbf{Partition Generation:} For each node $v_i$, a concept assignment $\psi_i$ is sampled based on the hierarchy $\mathcal{T}$. This determines the semantic identity of the node before its features are instantiated.
    \item \textbf{Embedding Generation:} The continuous representation $\boldsymbol{z}_i$ is generated conditioned on its assigned concept $\psi_i$ and the taxonomy $\mathcal{T}$, following a conditional likelihood $p_\theta(\boldsymbol{z}_i | \psi_i, \mathcal{T})$. For instance, $\boldsymbol{z}_i$ is drawn from a distribution centered at the semantic anchor of concept $\psi_i$.
    \item \textbf{Observation Generation:} Finally, the graph structure $\boldsymbol{A}$ is generated based on node embeddings and their partitions, governed by $p(\boldsymbol{A} | \boldsymbol{Z}, \psi)$~\cite{VGAE}. Consistent with stochastic block models~\cite{SBM} or homophily assumptions~\cite{Homophily}, edges are more likely to form between nodes in the same semantic partition or with high embedding similarity.
\end{enumerate}

Our goal is to maximize the marginal log-likelihood of the observed graph structure:
\begin{equation}
    \log p_\theta(\boldsymbol{A}) = \log
\sum_{\psi,\mathcal{T}}
\int
p(\boldsymbol{A},\boldsymbol{Z},\psi,\mathcal{T})
\,\mathrm{d}\boldsymbol{Z}.
\end{equation}

\subsection{Evidence Lower Bound (ELBO)}
Direct maximization of the marginal likelihood is intractable due to the integration over continuous embeddings $\boldsymbol{Z}$ and the summation over the combinatorial space of hierarchies.
We introduce a variational distribution
$q(\boldsymbol{Z},\psi,\mathcal{T})$ to approximate the true posterior.
Following mean-field variational inference~\cite{VEM}, we factorize it as $q(\boldsymbol{Z},\psi,\mathcal{T})=q_\theta(\boldsymbol{Z})q(\psi,\mathcal{T})$.
Here, $q_\theta(\boldsymbol{Z})$ denotes the variational posterior over continuous node representations, whose mean is parameterized by the GNN encoder.

Starting from the marginal likelihood, we have:
\begin{equation}
\begin{aligned}
\log p(\boldsymbol{A})
&=
\log
\sum_{\psi,\mathcal{T}}
\int
p(\boldsymbol{A},\boldsymbol{Z},\psi,\mathcal{T})
\,\mathrm{d}\boldsymbol{Z} \\
&=
\log
\sum_{\psi,\mathcal{T}}
\int
q_\theta(\boldsymbol{Z})q(\psi,\mathcal{T})
\frac{
p(\boldsymbol{A},\boldsymbol{Z},\psi,\mathcal{T})
}{
q_\theta(\boldsymbol{Z})q(\psi,\mathcal{T})
}
\,\mathrm{d}\boldsymbol{Z}.
\end{aligned}
\end{equation}
By Jensen's inequality, the ELBO is:
\begin{equation}
\begin{aligned}
\mathcal{L}_{\mathrm{ELBO}}
=&\
\mathbb{E}_{q}[\log p(\boldsymbol{A},\boldsymbol{Z},\psi,\mathcal{T})]
-
\mathbb{E}_{q}[\log q_\theta(\boldsymbol{Z})]
-
\mathbb{E}_{q}[\log q(\psi,\mathcal{T})].
\end{aligned}
\end{equation}
Using the factorization
\begin{equation}
p(\boldsymbol{A},\boldsymbol{Z},\psi,\mathcal{T})
=
p(\boldsymbol{A}|\boldsymbol{Z},\psi)
p(\boldsymbol{Z}|\psi,\mathcal{T})
p(\psi|\mathcal{T})
p(\mathcal{T}),
\end{equation}
we obtain:
\begin{equation}
\mathcal{L}_{\mathrm{ELBO}}=\mathbb{E}_{q}[\log p(\boldsymbol{A}|\boldsymbol{Z},\psi)] + \mathbb{E}_{q}[\log p(\boldsymbol{Z}|\psi,\mathcal{T})] - \mathrm{KL}\big(q(\psi,\mathcal{T})\,\|\,p(\psi,\mathcal{T})\big) + H(q_\theta(\boldsymbol{Z})),
\label{eq:elbo}
\end{equation}
where $p(\psi,\mathcal{T})=p(\psi|\mathcal{T})p(\mathcal{T})$ and
$H(q_\theta(\boldsymbol{Z}))=-\mathbb{E}_{q_\theta}[\log q_\theta(\boldsymbol{Z})]$ is the entropy of the continuous variational posterior.

In practice, \texttt{SHiFT} uses a deterministic GNN encoder.
This can be interpreted as using a fixed-variance Gaussian posterior:
\begin{equation}
q_\theta(\boldsymbol{Z})
=
\prod_{i=1}^{n}
\mathcal{N}
\left(
\boldsymbol{z}_i;
f_\theta(\mathcal{G},\boldsymbol{X})_i,
\sigma_q^2\boldsymbol{I}
\right),
\end{equation}
and evaluating the expected terms at the posterior mean.
Since $\sigma_q^2$ is fixed, the entropy term becomes:
\begin{equation}
H(q_\theta(\boldsymbol{Z}))
=
\frac{nd}{2}
\left(
1+\log(2\pi\sigma_q^2)
\right),
\end{equation}
which is constant with respect to the encoder parameters.
Therefore, although the entropy term is part of the formal ELBO, it does not affect the practical optimization objective and can be safely absorbed into the constant term.

\subsection{E-Step: Taxonomy Induction}
In the E-step, we fix the encoder parameters and thus the current variational posterior $q_\theta(\boldsymbol{Z})$.
We then optimize the discrete variational distribution $q(\psi,\mathcal{T})$.
Following the Hard-EM approximation~\cite{Hard-EM1}, we collapse $q(\psi,\mathcal{T})$ to a point estimate:
\begin{equation}
q(\psi,\mathcal{T})
=
\delta_{(\hat{\psi},\hat{\mathcal{T}})}(\psi,\mathcal{T}).
\end{equation}
Under this approximation, the entropy of $q(\psi,\mathcal{T})$ becomes zero, and the hierarchy-related part of the ELBO reduces to finding a taxonomy and node assignment that are both compatible with the current embeddings and likely under the semantic prior:
\begin{equation}
(\hat{\psi},\hat{\mathcal{T}})
\approx
\arg\max_{\psi,\mathcal{T}}
\mathbb{E}_{q_\theta(\boldsymbol{Z})}
[
\log p(\boldsymbol{Z}|\psi,\mathcal{T})
]
+
\log p(\psi,\mathcal{T}).
\end{equation}
The first term encourages the induced hierarchy to explain the current embedding distribution, while the second term enforces semantic plausibility.
Since the semantic prior $p(\mathcal{T})$ is analytically unavailable, \texttt{SHiFT} uses the LLM as a proxy for this prior~\cite{LAMA}.
The Mapping phase finds a provisional assignment based on embedding similarity, while the Diagnosis phase invokes the LLM to refine semantically unstable regions, thereby approximating the discrete E-step.

\subsection{M-Step: Representation Learning}
In the M-step, we fix the hierarchy $\hat{\mathcal{T}}$ and partition $\hat{\psi}$ obtained from the E-step, and optimize the encoder parameters $\theta$.
Since $q_\theta(\boldsymbol{Z})$ is modeled as a fixed-variance Gaussian centered at the deterministic GNN outputs, the entropy term $H(q_\theta(\boldsymbol{Z}))$ is constant.
Therefore, optimizing $\theta$ amounts to maximizing the following ELBO components:
\begin{equation}
\mathbb{E}_{q_\theta}
[
\log p(\boldsymbol{A}|\boldsymbol{Z},\hat{\psi})
]
+
\mathbb{E}_{q_\theta}
[
\log p(\boldsymbol{Z}|\hat{\psi},\hat{\mathcal{T}})
].
\end{equation}
We show that our three alignment objectives naturally correspond to these two components.

\vpara{Partition Alignment ($\mathcal{L}_{tpa}$) $\Leftrightarrow \log p(\boldsymbol{A} | \boldsymbol{Z}, \hat{\psi})$.}
This term maximizes the likelihood of the graph structure.
We assume a generative mechanism where edges arise from both pairwise embedding similarity, corresponding to homophily~\cite{Homophily}, and shared semantic partitions, corresponding to community structure~\cite{SBM}.
Consequently, a high likelihood $p(\boldsymbol{A}|\boldsymbol{Z},\hat{\psi})$ implies that nodes should be close in the embedding space if they are either topologically connected or assigned to the same semantic cluster.
Our $\mathcal{L}_{tpa}$ operationalizes this by constructing an augmented adjacency $\boldsymbol{A}^+$ that includes both observed edges and intra-partition pairs.
By treating these pairs as positives in the contrastive objective, the encoder respects the observed topology while enforcing consistency with the latent partition derived in the E-step.

\vpara{Embedding Prior Matching $\Leftrightarrow \log p(\boldsymbol{Z} | \hat{\psi}, \hat{\mathcal{T}})$.}
The second term constrains node embeddings to follow the hierarchical prior implied by the induced taxonomy.
We decompose this prior into structural and semantic components:
\begin{itemize}[leftmargin=1.5em, itemsep=0.2em]
    \item \textbf{Skeleton Alignment ($\mathcal{L}_{tsa}$):}
    We model the structural prior as preserving the tree metric:
    \begin{equation}
    p(\boldsymbol{Z}|\hat{\mathcal{T}})
    \propto
    \exp
    \left(
    -\mathrm{dist}
    (
    \boldsymbol{D}^{\boldsymbol{Z}},
    \boldsymbol{D}^{\hat{\mathcal{T}}}
    )
    \right).
    \end{equation}
    Maximizing this likelihood corresponds to aligning the embedding geometry with the taxonomy skeleton. 
    $\mathcal{L}_{tsa}$ implements this alignment through the Cophenetic correlation objective.

    \item \textbf{Concept Alignment ($\mathcal{L}_{tca}$):}
    We model the semantic content prior as Gaussian distributions centered at LLM-generated concept anchors:
    \begin{equation}
    p(\boldsymbol{z}_i|\hat{\psi}_i)
    =
    \mathcal{N}
    (
    \boldsymbol{z}_i;
    \boldsymbol{e}_{\hat{\psi}_i},
    \sigma^2\boldsymbol{I}
    ).
    \end{equation}
    Maximizing the corresponding log-likelihood is equivalent to minimizing the Euclidean distance between a node representation and its assigned concept anchor.
    $\mathcal{L}_{tca}$ extends this idea to all concepts along the root-to-leaf path, thereby enforcing multi-level semantic grounding.
\end{itemize}

\subsection{Convergence Discussion}
The above derivation shows that \texttt{SHiFT} can be interpreted as a generalized hard Variational EM algorithm~\cite{Hard-EM1}.
The E-step updates the discrete hierarchy variables $(\hat{\psi},\hat{\mathcal{T}})$, while the M-step updates the continuous encoder parameters $\theta$.
The entropy term $H(q_\theta(\boldsymbol{Z}))$ is part of the formal ELBO, but becomes constant under our fixed-variance posterior approximation and thus does not affect optimization. 
Unlike classical EM, however, \texttt{SHiFT} does not provide a strict monotonic convergence guarantee.
This is because the E-step involves LLM-based discrete taxonomy edits, which approximate rather than exactly solve the posterior update, and the M-step optimizes non-convex neural objectives that serve as tractable surrogates of the likelihood terms.
Therefore, \texttt{SHiFT} should be viewed as a generalized surrogate EM procedure rather than an exact EM algorithm.
Empirically, the alternating updates lead to stable convergence, as the induced hierarchy and learned embeddings progressively become mutually consistent.

\section{Detailed Experimental Setup}\label{app:setup_details}

\subsection{Datasets}\label{app:datasets}
To provide a comprehensive evaluation of our proposed method, we adopt five datasets from the \texttt{LLMNodeBed}~\cite{LLMNodeBed} benchmark, as summarized in Table~\ref{tab:dataset_statistics}. 
These datasets span diverse domains, including academic networks, web graphs, and product co-purchase networks, and cover a wide range of scales, from thousands to hundreds of thousands of nodes. 
Within \texttt{LLMNodeBed}, each dataset is stored in \texttt{.pt} format using PyTorch, which includes shallow embeddings, raw text of nodes, edge indices, labels, and data splits for convenient loading. 
The processed data is publicly available at \url{https://huggingface.co/datasets/xxwu/LLMNodeBed} under the MIT license.
A description of these datasets is provided below:

\begin{itemize}[leftmargin=1.5em, itemsep=0.2em]
    \item \textbf{Academic Networks:} The \textbf{Cora}, \textbf{Citeseer}~\cite{GCN}, and \textbf{ogbn-arXiv} (abbreviated as "ArXiv")~\cite{OGB} datasets consist of nodes representing papers, with edges indicating citation relationships. 
    The associated text attributes include each paper's title and abstract, which were collected by \citet{TAPE} and \citet{LLM4Graph3}, respectively. 
    Within the dataset, each node is labeled according to its category. 
    For example, the ArXiv dataset includes 40 CS sub-categories such as cs.AI (Artificial Intelligence) and cs.DB (Databases).

    \item \textbf{Web Link Network:} In the \textbf{WikiCS} dataset~\cite{WikiCS}, each node represents a Wikipedia page, and edges indicate reference links between pages. 
    The raw text for each node includes the page name and content, which was collected by \citet{OFA}. 
    The classification goal is to categorize each entity into different Wikipedia categories.

    \item \textbf{E-Commerce Networks:} The \textbf{Ele-Computer} (abbreviated as "Computer") dataset is derived from the Amazon Electronics dataset~\cite{Amazon}, where each node represents an item in the Computer category. 
    We utilize the processed datasets released in~\citet{CS-TAG}. 
    In this e-commerce network, edges indicate co-purchase or co-view relationships. The associated text for each item includes descriptions or user reviews. 
    The classification task involves categorizing the products into fine-grained sub-categories.
\end{itemize}

\begin{table}[t!]
\centering
\caption{Dataset statistics including the number of nodes, edges, and classes, and their domains.}\label{tab:dataset_statistics}
\begin{tabular}{lccccc}
\toprule
Dataset & \#Nodes & \#Edges & \#Classes & Domain \\
\midrule 
Cora & 2,708 & 5,429 & 7 & Citation \\
Citeseer & 3,186 & 4,277 & 6 & Citation \\
ArXiv & 169,343 & 1,166,243 & 40 & Citation \\
WikiCS & 11,701 & 216,123 & 10 & Web Link \\
Computer & 87,229 & 721,081 & 10 & E-Commerce \\
\bottomrule
\end{tabular}
\end{table}

These five datasets release flat class labels only.
That does not mean the underlying domains are unstructured: citation networks are organized in practice by taxonomies such as the ACM Computing Classification System, and product graphs by catalogs such as the Amazon product taxonomy, but those hierarchical annotations are not part of the public splits.
Flat accuracy and clustering gains are therefore consistent with hierarchy-aware representations capturing latent structure, yet they do not directly test recovery of a known tree.
Appendix~\ref{app:patent} supplies that test on a patent graph whose CPC labels are assigned by examiners.

\subsection{Baselines}\label{app:baselines}
Following the \texttt{LLMNodeBed}~\cite{LLMNodeBed} benchmark, we compare our method against the following categories of approaches:
\begin{itemize}[leftmargin=1.5em, itemsep=0.2em]
    \item \textbf{Classic GNNs with Shallow Embeddings:} GCN$_{\text{SE}}$~\cite{GCN}, SAGE$_{\text{SE}}$~\cite{SAGE}, and GAT$_{\text{SE}}$~\cite{GAT}, which use traditional GNN architectures combined with shallow text features such as bag-of-words.

    \item \textbf{Hyperbolic GNNs:} HGCN~\cite{HGCN} and HGCL~\cite{HGCL}, which map data into hyperbolic space to capture hierarchy implicitly.
    
    \item \textbf{PLM-based Methods:} Fine-tuned encoder-only pre-trained language models used to obtain text embeddings for classification, including SenBERT (66M)~\cite{SentenceBERT} and RoBERTa (355M)~\cite{RoBERTa}.
    
    \item \textbf{Hybrid PLM + GNN:} Models integrating PLM-derived textual embeddings with GNN-based structural learning. We evaluate GIANT~\cite{GIANT} and GLEM~\cite{GLEM} as representative methods.
    
    \item \textbf{LLM-as-Encoder:} Approaches that use large language models as node feature encoders, injecting LLM-derived embeddings into graph learning pipelines. We include ENGINE~\cite{ENGINE} and GCN$_{\text{LLM}}$~\cite{LLMNodeBed}.
    
    \item \textbf{LLM-as-Explainer:} Methods that leverage LLMs to generate auxiliary semantic explanations to enhance node representations. We consider TAPE~\cite{TAPE} and GAugLLM~\cite{GAugLLM}.
    
    \item \textbf{LLM-as-Predictor:} Approaches where LLMs directly perform node classification via prompts or instruction-tuned generation. We include GraphGPT~\cite{GraphGPT}, LLaGA~\cite{LLaGA}, and instruction-tuned LLMs (LLM$_{\text{IT}}$)~\cite{LLMNodeBed}.
\end{itemize}

\subsection{Implementation Details}
\label{app:implementation}

\subsubsection{Architecture Setup}
To ensure a fair comparison with state-of-the-art methods, we align our architectural choices with the \texttt{LLMNodeBed} benchmark suite~\cite{LLMNodeBed}.
\begin{itemize}[leftmargin=1.5em, itemsep=0.2em]
    \item \textbf{Text Encoder (PLM):} We utilize the frozen \textbf{RoBERTa-Large} (355M parameters)~\cite{RoBERTa} to encode raw node documents into initial feature vectors.
    \item \textbf{Graph Encoder (GNN):} We employ a standard \textbf{GCN}~\cite{GCN} as the backbone encoder. Following the benchmark, we perform a grid search for the optimal architecture: number of layers $L \in \{2, 3\}$, hidden dimension $d \in \{64, 128, 256\}$, with optional batch normalization and residual connections.
    \item \textbf{LLM for Taxonomy Induction:} For the semantic reasoning in the E-step, we access \textbf{DeepSeek-V3}~\cite{DeepSeekV3} via API. To ensure reproducibility, we set the generation temperature to $0.1$.
\end{itemize}

\subsubsection{Algorithm Configuration}
The training process of \texttt{SHiFT} involves a warm-up phase followed by the iterative EM loop. The specific configurations for each stage are as follows:

\vpara{Warm-up: Graph Contrastive Pretraining.}
Before the EM iterations, we initialize the node embeddings via structure-aware contrastive learning (Eq.~\eqref{eq:gcp}).
We pretrain the GNN encoder for up to 50 epochs using the Adam optimizer. The contrastive weight $\gamma$ is set to $1.0$.

\vpara{E-Step: Hierarchical Semantic Induction.}
We construct the taxonomy with a maximum depth of $L_{max}=3$. To ensure statistical significance of the induced concepts, we enforce the following constraints during the recursive clustering and LLM diagnosis:
\begin{itemize}[leftmargin=1.5em, itemsep=0.2em]
    \item \textbf{Minimum Cluster Size:} A valid concept node must contain at least $N_{min}=50$ document samples.
    \item \textbf{Merge Threshold:} Two sibling concepts are merged if their centroid cosine similarity exceeds $\tau_{merge} = 0.9$, indicating semantic redundancy.
    \item \textbf{Split Criterion:} A cluster is flagged for splitting if its internal variance increases by a factor of $1.5\times$ compared to the previous iteration, suggesting low cohesion.
    \item \textbf{Pruning:} Degenerate clusters with fewer than 5 members are pruned as outliers.
\end{itemize}

\vpara{M-Step: Hierarchical Representation Learning.}
Fixing the induced taxonomy, we optimize the GNN encoder to minimize the joint objective $\mathcal{L}_{total}$.
To mitigate the quadratic complexity of contrastive terms, we employ \textbf{negative sampling} for both the pretraining (Eq.~\eqref{eq:gcp}) and partition alignment (Eq.~\eqref{eq:tpa}) objectives.
Specifically, for each positive pair, we sample exactly one negative pair (1:1 ratio).
We train for 10 epochs per iteration.
The balancing hyperparameters are set by default: Taxonomy Partition Alignment weight $\gamma=1.0$, Taxonomy Skeleton Alignment weight $\alpha=0.5$, and Taxonomy Concept Alignment weight $\beta=0.3$.

\vpara{Iterative Protocol.}
The full \texttt{SHiFT} framework executes the EM loop (E-step $\leftrightarrow$ M-step) for a total of 3 iterations, which we empirically found sufficient for convergence.

\subsubsection{Evaluation Protocol}
We strictly follow the standard semi-supervised learning setting defined in \texttt{LLMNodeBed}.
\begin{itemize}[leftmargin=1.5em, itemsep=0.2em]
    \item \textbf{Node Classification:} After the unsupervised representation learning, we freeze the GNN encoder and attach a linear classification head. We fine-tune this head using the official train/validation/test splits provided by the benchmark. We report the mean Accuracy and standard deviation across 5 random seeds.
    \item \textbf{Node Clustering:} Following~\cite{PyAGC, OFAG}, we apply K-Means clustering directly on the learned node embeddings $\boldsymbol{Z}$. The number of clusters $K$ is set to the ground-truth class count of each dataset. Performance is evaluated using Normalized Mutual Information (NMI) and Adjusted Rand Index (ARI).
\end{itemize}

\section{List of LLM Prompts}
\label{app:prompts}

We employ a systematic prompt engineering strategy to guide the LLM in inducing and refining the semantic taxonomy. 
All prompts are designed to output structured JSON objects to ensure parseability.
In the templates below, placeholders enclosed in \texttt{\{\{ \}\}} denote dynamic content inserted during runtime.
To optimize token usage, input document texts are truncated to the first 500 characters.

\subsection{Taxonomy Induction (Cold-Start)}

\subsubsection{Concept Induction Prompt}
\label{prompt:concept_induction}
This prompt corresponds to the E-step's Phase 3 (Section~\ref{sec:taxonomy_initialization}), used to generate a concept name and definition from a cluster of document embeddings.

\begin{tcolorbox}[title=System Prompt]
You are an expert taxonomist assisting in organizing scientific literature.
\end{tcolorbox}

\begin{tcolorbox}[title=User Prompt]
Context: We are currently exploring sub-topics under the parent category: \texttt{\{\{ parent\_info \}\}}.

Task: Below are \texttt{\{\{ num\_docs \}\}} representative papers from a specific cluster of documents within this parent category.
Your goal is to summarize the common research topic shared by these papers into a specific sub-concept.

Additionally, determine whether this sub-concept is semantically coherent and focused enough that it should NOT be further subdivided. 
If the papers clearly belong to a single, narrow topic with no distinct sub-directions, set "concept\_split\_needed" to false. 
If there appear to be multiple distinct sub-themes or methodological directions, set it to true.

Input Papers:
\texttt{\{\{ documents\_list \}\}}

Constraints:
1. The Concept Name should be concise (2-5 words) and standard in the field.
2. The Definition should be a single sentence explaining what this sub-concept focuses on, distinct from other sub-fields.
3. Ensure the concept strictly belongs to \texttt{\{\{ parent\_info \}\}}.

Output Format (JSON):
\{
  "concept\_name": "String",
  "definition": "String",
  "concept\_split\_needed": true or false
\}
\end{tcolorbox}

\subsubsection{Taxonomy Refinement Prompt}
\label{prompt:refinement}
This prompt corresponds to the E-step's Phase 4 (Section~\ref{sec:taxonomy_initialization}), acting as an "Editor" to clean up the over-clustered candidates via MERGE or PRUNE operations.

\begin{tcolorbox}[title=System Prompt]
You are a senior editor optimizing a hierarchical taxonomy.
\end{tcolorbox}

\begin{tcolorbox}[title=User Prompt]
Context: Under the parent category \texttt{\{\{ parent\_info \}\}}, we identified \texttt{\{\{ num\_candidates \}\}} candidate sub-concepts based on clustering. However, the clustering algorithm may have produced:
1. Duplicates/Synonyms (e.g., "GNN" and "Graph Neural Nets" should be MERGED).
2. Overly fragmented topics (e.g., "LSTM applications" and "RNN models" might be MERGED into "Recurrent Neural Networks").
3. Outliers/Irrelevant topics (e.g., a Biology topic appearing in a CS list should be PRUNED).

Candidate Sub-concepts:
ID | Name | Definition
\texttt{\{\{ concept\_list \}\}}

Task: Review the list and output a set of operations to refine this level of the taxonomy.
Allowed Operations:
- MERGE: Combine IDs [id1, id2, ...] into a new, broader concept. Provide a new Name and Definition.
- PRUNE: Remove ID [id] if it is clearly off-topic, too vague, or low quality.
- KEEP: Keep ID [id] as is (or rename it slightly for clarity).

Output Format (JSON list):
[
  \{"op": "MERGE", "source\_ids": [1, 5, 8], "new\_name": "Graph Neural Networks", "new\_def": "...", "reason": "Synonyms"\},
  \{"op": "PRUNE", "target\_id": 3, "reason": "Off-topic physics paper"\},
  \{"op": "KEEP", "target\_id": 2, "new\_name": "Transformers"\}
]
\end{tcolorbox}

\subsection{Taxonomy Maintenance (Incremental Updates)}

\subsubsection{Merge Diagnosis Prompt}
\label{prompt:merge_diagnosis}
Used in the "Surgical Diagnosis" phase (Section~\ref{sec:taxonomy_updates}) when sibling concepts exhibit high cosine similarity. It supports multi-way merging strategies.

\begin{tcolorbox}[title=System Prompt]
You are an expert taxonomist optimizing a document taxonomy.
\end{tcolorbox}

\begin{tcolorbox}[title=User Prompt]
Context: Under the parent category \texttt{\{\{ parent\_info \}\}}, we identified \texttt{\{\{ num\_siblings \}\}} sibling concepts that appear semantically similar based on their vector representations.

Similar Concepts:
\texttt{\{\{ concept\_details \}\}}

Task: Analyze these concepts and determine the best merging strategy.

You can output ONE of the following strategies:
Strategy A: MERGE ALL (Use when they are all synonyms or very closely related)
Strategy B: PARTIAL MERGE (Use when there are subgroups of similar concepts)
Strategy C: KEEP ALL SEPARATE (Use when they are genuinely distinct despite high similarity)

Consider:
1. Are they synonyms? (e.g., "GNN", "Graph Neural Nets", "Graph NN")
2. Are they closely related subtopics? (e.g., "LSTM", "GRU", "Recurrent Nets")
3. Are they distinct but related? (e.g., "Image Classification", "Object Detection")

Output Format (JSON):
\{
  "strategy": "MERGE\_ALL" or "PARTIAL\_MERGE" or "KEEP\_SEPARATE",
  "reasoning": "Brief explanation of your decision",
  "operations": [ ... ]
\}
\end{tcolorbox}

\subsubsection{Definition Refinement Prompt}
\label{prompt:definition_refinement}
Triggered when a cluster experiences significant member turnover (Concept Drift), asking the LLM to update the definition based on new core-set documents.

\begin{tcolorbox}[title=System Prompt]
You are an expert taxonomist refining concept definitions.
\end{tcolorbox}

\begin{tcolorbox}[title=User Prompt]
Context: We have an existing concept whose member documents have changed significantly after embedding updates.

Current Concept:
- Name: \texttt{\{\{ current\_name \}\}}
- Definition: \texttt{\{\{ current\_def \}\}}

New Representative Papers:
\texttt{\{\{ new\_docs \}\}}

Task: Determine if the current definition still accurately describes these papers, or if it needs refinement.

Consider:
1. Does the old definition still capture the main theme of these papers?
2. Has the focus shifted slightly (e.g., from theory to applications)?
3. Should we keep the name but refine the definition?

Output Format (JSON):
\{
  "should\_update": true or false,
  "reason": "Brief explanation",
  "new\_name": "Refined name (can be same as old)",
  "new\_definition": "Refined definition"
\}
\end{tcolorbox}

\subsubsection{Outlier Assignment Prompt}
\label{prompt:outlier}
Used to re-assign "unassigned bucket" nodes to the most relevant existing concept based on semantic content.

\begin{tcolorbox}[title=System Prompt]
You are an expert taxonomist assigning documents to concepts.
\end{tcolorbox}

\begin{tcolorbox}[title=User Prompt]
Context: We have a document that doesn't clearly belong to any existing cluster. 
We need to assign it to the most appropriate concept.

Document:
\texttt{\{\{ document\_text \}\}}

Candidate Concepts:
\texttt{\{\{ candidate\_list \}\}}

Task: Which concept does this document best fit into? Consider the core topic and methodology.

Output Format (JSON):
\{
  "best\_match\_id": 1 or 2 ... (or 0 if none fit),
  "reason": "Brief explanation"
\}
\end{tcolorbox}

\section{Taxonomy Evaluation Protocols}\label{app:taxo_eval_protocols}

\subsection{Concept Coherence Calculation}
\label{app:concept_coherence_calculation}
To strictly quantify the semantic coherence of the induced taxonomy, we employ Normalized Pointwise Mutual Information (NPMI). The calculation proceeds in two steps:

\textbf{1. Keyword Extraction via c-TF-IDF.}
We first aggregate all documents assigned to a concept $c$ into a single "super-document". We then calculate the Class-based TF-IDF score for each word $w$ in concept $c$:
\begin{equation}
    W_{w, c} = \text{tf}_{w, c} \times \log\left(1 + \frac{N}{\text{df}_w}\right),
\end{equation}
where $tf_{w,c}$ is the frequency of word $w$ in concept $c$, $N$ is the total number of concepts, and $df_w$ is the number of concepts containing $w$. We select the top-10 words with the highest scores as the representative keyword set $W_c$.

\textbf{2. NPMI Coherence Score.}
We measure the pairwise co-occurrence of words in $W_c$ across the entire corpus using a sliding window. The NPMI for a pair $(w_i, w_j)$ is defined as:
\begin{equation}
    \text{NPMI}(w_i, w_j) = \frac{\log \frac{P(w_i, w_j)}{P(w_i) P(w_j)}}{-\log P(w_i, w_j)},
\end{equation}
where $P(w)$ is the probability of word occurrence. The final score for the taxonomy is the average NPMI over all pairs in all concepts.

\subsection{Human Evaluation Protocol}
\label{app:human_eval}

We recruited three CS graduate students (with backgrounds in NLP and graph learning) as annotators.
The evaluation was conducted on the induced taxonomies of Cora, covering three taxonomy variants: the text-only baseline ($\mathcal{T}_{\text{w/o }\mathcal{G}}$), the cold-start initialization ($\mathcal{T}_{\text{init}}$), and the final \texttt{SHiFT} taxonomy ($\mathcal{T}_{\text{final}}$).

For each dataset and each taxonomy variant, we randomly sampled 20 concept nodes.
Each annotator was presented with:
\begin{itemize}[leftmargin=1.5em, itemsep=0.2em]
    \item The concept name and definition generated by the LLM.
    \item Five representative documents assigned to that concept (selected by the dual-level core-set sampling described in Sec.~\ref{sec:taxonomy_initialization}).
    \item The concept's parent node name and definition (to enable assessment of the parent-child relation).
\end{itemize}
Annotators were asked to rate each concept on the following three criteria using a 1--5 Likert scale (1 = strongly disagree, 5 = strongly agree):

\begin{itemize}[leftmargin=1.5em, itemsep=0.2em]
    \item \textbf{Readability}: The concept name and definition are clear, concise, and well-defined.
    \item \textbf{Coherence}: The representative documents genuinely and consistently belong to this concept.
    \item \textbf{Hierarchy Validity}: The parent-child relationship between this concept and its parent is logically sound and follows an is-a relation.
\end{itemize}

The three taxonomy variants were presented in randomized order, and annotators were blind to which variant they were evaluating.
The annotation interface displayed the concept information in a unified format to prevent visual bias.

\subsection{Taxonomy as Features Details}
\label{app:taxonomy_as_features}
In the extrinsic evaluation (Sec~\ref{sec:taxonomy_evaluation}), we construct the augmented feature input as follows:
\begin{enumerate}[leftmargin=1.5em, itemsep=0.2em]
    \item \textbf{Path Extraction:} For a node $v$ assigned to a leaf concept, we trace the path from the root $c_{root} \to \dots \to c_{leaf}$.
    \item \textbf{Text Linearization:} We concatenate the names and definitions of all concepts along the path to form $\mathcal{P}(v) = \text{"[ROOT] } name_{root}: def_{root} \dots \text{ [LEAF] } name_{leaf}: def_{leaf}"$.
    \item \textbf{Input Construction:} The final input is $S_{new} = S_{raw} \oplus \texttt{[SEP]} \oplus \mathcal{P}(v)$.
    \item \textbf{Encoding:} We encode $S_{new}$ using the frozen PLM to obtain $X_{aug} \in \mathbb{R}^{d}$.
\end{enumerate}
We then train a standard 2-layer MLP and a 2-layer GCN on $X_{aug}$ using the cross-entropy loss. 

\section{Evaluation on a Patent Graph with Verifiable Hierarchy}
\label{app:patent}

This section tests whether an induced taxonomy recovers a hierarchy that was assigned independently of the model.
The flat-label benchmarks cannot answer that question on their own (Appendix~\ref{app:datasets}).

\subsection{Dataset}
\label{app:patent_data}

We construct \textbf{Patent}, a text-attributed graph of 10{,}090 patents.
Node text (title and abstract) and examiner citation links are taken from the public USPTO collection at \url{https://github.com/opendata-ai/tr}.
Each node is annotated with its primary Cooperative Patent Classification (CPC) code from Google Patents Public Data (CC BY 4.0).
CPC codes are assigned by USPTO patent examiners.
Restricting to the primary code yields a strict 3-level tree on this graph: \textbf{L1} Section (7 classes), \textbf{L2} Class (55 classes), and \textbf{L3} Subclass (162 classes).
Gold concept names are read from the official CPC Title List (USPTO/EPO, May 2026).\footnote{\url{https://www.cooperativepatentclassification.org}}
We use a temporal split, train $\le 2013$, validation $2014$--$2015$, and test $\ge 2016$, and release the processed graph with our code.

\subsection{Hierarchical Classification}
\label{app:patent_clf}

We compare \texttt{SHiFT} with MLP, GCN, and TAPE under the same encoder and training protocol as Section~\ref{sec:experimental_setup}, using the temporal split above.
Each method is scored at all three CPC levels.
Hierarchical F-score follows Kiritchenko et al.~\cite{Kiritchenko2006}.
For a test patent $i$, let $Y_i$ be the gold subclass together with its ancestors, excluding the root, and let $\hat{Y}_i$ be the analogous set for the prediction.
Micro-averaged hierarchical precision, recall, and F-score are
\begin{equation}
hP = \frac{\sum_i |\hat{Y}_i \cap Y_i|}{\sum_i |\hat{Y}_i|}, \qquad
hR = \frac{\sum_i |\hat{Y}_i \cap Y_i|}{\sum_i |Y_i|}, \qquad
hF = \frac{2\, hP\, hR}{hP+hR}.
\label{eq:hf}
\end{equation}
A subclass error that still matches the true class or section receives partial credit, so $hF$ is not a restatement of Acc@L3.
Results over five seeds are in Table~\ref{tab:patent_clf}.

\begin{table}[t]
\centering
\caption{Hierarchical classification on Patent (\%). Best results are in bold.}
\label{tab:patent_clf}
\begin{tabular}{lcccc}
\toprule
\textbf{Method} & \textbf{Acc@L3} & \textbf{Acc@L2} & \textbf{Acc@L1} & \textbf{hF} \\
\midrule
MLP  & 64.32 \scriptsize{$\pm$0.58} & 73.65 \scriptsize{$\pm$0.65} & 81.32 \scriptsize{$\pm$0.49} & 73.45 \scriptsize{$\pm$0.42} \\
GCN  & 70.91 \scriptsize{$\pm$0.52} & 78.52 \scriptsize{$\pm$0.33} & 83.84 \scriptsize{$\pm$0.18} & 77.85 \scriptsize{$\pm$0.39} \\
TAPE & 71.54 \scriptsize{$\pm$0.41} & 79.43 \scriptsize{$\pm$0.28} & 84.41 \scriptsize{$\pm$0.31} & 78.96 \scriptsize{$\pm$0.33} \\
\textbf{\texttt{SHiFT}} & \textbf{74.18 \scriptsize{$\pm$0.44}} & \textbf{81.27 \scriptsize{$\pm$0.31}} & \textbf{85.73 \scriptsize{$\pm$0.25}} & \textbf{81.34 \scriptsize{$\pm$0.28}} \\
\bottomrule
\end{tabular}
\end{table}

The margin over the strongest flat baseline, TAPE, increases with granularity: $+1.3$ at L1, $+1.8$ at L2, and $+2.6$ at L3, with $+2.4$ on $hF$.
Coarse section labels are easy for a strong text model; subclass labels are where coarse-to-fine structure helps.
The $hF$ gain is the same phenomenon measured along the ancestor chain rather than at a single cut of the tree.
This is the pattern expected when the encoder uses the hierarchy, and it is not explained by graph smoothing alone: GCN already has the citation graph, and TAPE already has LLM-enhanced text.

\subsection{Alignment with the CPC Tree}
\label{app:patent_align}

Classification shows that hierarchy-aware representations help when the label tree is the target.
We next ask whether the taxonomy \texttt{SHiFT} induces, without reading CPC symbols, agrees with that tree.
We compare the three taxonomies already used in Section~\ref{sec:taxonomy_evaluation}: text-only $\mathcal{T}_{\text{w/o }\mathcal{G}}$ (PLM embeddings), cold-start $\mathcal{T}_{\text{init}}$ (after graph-contrastive warm-up), and $\mathcal{T}_{\text{final}}$ (after EM refinement).

\vpara{Node-level semantic alignment.}
For an induced concept at depth $d \in \{1,2,3\}$, we mean-pool frozen RoBERTa embeddings of the title and abstract of patents assigned to that concept.
CPC classes at the same level are embedded in the same way, using the patents that carry each code.
Each induced concept retrieves its five nearest CPC classes by cosine similarity.
We count a match when the nearest neighbor exceeds a threshold $\tau$ fixed before the comparison and shared by all three taxonomies.
Precision is the fraction of induced concepts with a match.
Recall is the fraction of CPC classes matched by at least one induced concept.

\vpara{Document-level hierarchy recovery.}
Semantic match of concept names can succeed even when the tree geometry is wrong.
For every pair of patents we compute the cophenetic distance under the induced tree and under the CPC tree, using the depth of their lowest common ancestor at the leaf level~\cite{Sokal1962}, and report the Pearson correlation between the two distance vectors.
The score is high only when sibling and cousin relations agree, not merely when individual nodes have a nearby CPC title.

\begin{table}[t]
\centering
\caption{Agreement between induced taxonomies and the official CPC tree on Patent.}
\label{tab:patent_align}
\begin{tabular}{lccc}
\toprule
\textbf{Taxonomy} & \textbf{Precision} & \textbf{Recall} & \textbf{Cophenetic $r$} \\
\midrule
$\mathcal{T}_{\text{w/o }\mathcal{G}}$ (text-only) & 0.7969 & 0.5160 & 0.2852 \\
$\mathcal{T}_{\text{init}}$ (cold-start) & 0.8814 & 0.5863 & 0.5603 \\
$\mathcal{T}_{\text{final}}$ (Ours) & \textbf{0.9145} & \textbf{0.6399} & \textbf{0.6423} \\
\bottomrule
\end{tabular}
\end{table}

Table~\ref{tab:patent_align} supports three observations.
(1) \emph{The citation graph carries hierarchical signal that text alone misses.}
$\mathcal{T}_{\text{w/o }\mathcal{G}}$ already matches many CPC titles (precision $0.80$) but barely preserves pairwise tree distance ($r = 0.29$).
Adding the patent graph in $\mathcal{T}_{\text{init}}$ nearly doubles cophenetic correlation ($0.29 \to 0.56$).
Concepts induced from text can look locally coherent and still scramble which documents are siblings.
(2) \emph{EM refinement improves coverage and geometry together.}
From $\mathcal{T}_{\text{init}}$ to $\mathcal{T}_{\text{final}}$, precision rises by $0.033$, recall by $0.054$, and cophenetic correlation by $0.082$.
The same ordering, text-only $<$ cold-start $<$ final, appears in NPMI and in human Hierarchy Validity on Cora, Citeseer, and WikiCS (Section~\ref{sec:taxonomy_evaluation}).
Those proxy metrics were not an artifact of missing labels: the same refinement moves the induced tree toward an external gold hierarchy.
(3) \emph{Full recovery of CPC is not the objective.}
Recall $0.64$ and $r = 0.64$ leave a real gap.
\texttt{SHiFT} names concepts from the documents and chooses its own granularity; it does not emit USPTO symbols.
The monotonic gain across all three metrics is still evidence that this graph contains a recoverable hierarchy, and that hierarchy-aware induction tracks it more closely than a graph-agnostic clustering of the same text.

\section{Efficiency and Scalability Analysis}
\label{app:efficiency}

We analyze the efficiency of \texttt{SHiFT} from three perspectives: theoretical complexity, empirical runtime comparison against baselines, and a detailed breakdown of LLM inference costs.

\subsection{Complexity Analysis}
\label{app:complexity}

\vpara{E-Step (Taxonomy Induction).}
The cold-start initialization performs hierarchical K-Means clustering over $n$ nodes with embedding dimension $d$.
At each recursion level, clustering $n$ nodes into $K$ micro-clusters costs $\mathcal{O}(n K d)$ per K-Means iteration.
Core-set sampling via CCPR requires computing PageRank on induced subgraphs, which costs $\mathcal{O}(|\mathcal{E}|)$ in total across all clusters.
LLM calls scale with the number of micro-clusters $K_{\text{total}}$ across all recursion levels; since $K_{\text{total}} \ll n$ in practice, this step is decoupled from graph scale.
For the incremental update (Mapping \& Diagnosis), node-to-centroid assignment costs $\mathcal{O}(n K d)$, and LLM calls are triggered only for anomalous clusters, reducing to a small fraction of $K_{\text{total}}$.

\vpara{M-Step (Representation Learning).}
The GNN forward pass costs $\mathcal{O}(|\mathcal{E}| d + n d^2)$ per layer, consistent with standard GCN complexity~\cite{GCN}.
Constructing the partition-augmented adjacency $\boldsymbol{A}^+$ for TPA adds at most $\mathcal{O}(n^2 / K)$ intra-cluster pairs on average, which is manageable for moderate $K$.
TSA computes pairwise distances among $K$ cluster centroids, costing $\mathcal{O}(K^2 d)$.
TCA evaluates path-wise distances for each node over $L$ levels, costing $\mathcal{O}(nLd)$.
Since $K \ll n$ and $L$ is small (typically $\leq 4$), both TSA and TCA introduce negligible overhead relative to the GNN backbone.

Overall, the dominant cost per EM iteration is $\mathcal{O}(|\mathcal{E}| d + nKd)$, which is comparable to a standard graph contrastive learning pipeline.
The LLM cost is bounded by $\mathcal{O}(K_{\text{total}})$ calls per E-step, which decreases substantially after the cold-start phase.

\subsection{Empirical Efficiency Comparison}
\label{app:empirical_efficiency}

We conduct a comprehensive efficiency analysis on the large-scale ArXiv dataset (169,343 nodes; 1,166,243 edges) and compare \texttt{SHiFT} against representative baselines from the \texttt{LLMNodeBed}~\cite{LLMNodeBed}.

\vpara{Hardware Setup.}
All baseline results are reported on a single NVIDIA H100 (80GB) GPU.
\texttt{SHiFT} is evaluated on a single NVIDIA Tesla V100 (32GB) GPU, which has considerably lower memory bandwidth and compute capacity.
This hardware asymmetry makes the comparison conservative and favorable to the baselines.

Table~\ref{tab:efficiency_comparison} summarizes GPU memory usage and total training time.
Since \texttt{SHiFT} uses LLM through an external API, its reported GPU memory reflects only the local graph-side computation, whereas some LLM-based baselines include the memory footprint of locally hosted LLMs.
Therefore, Table~\ref{tab:efficiency_comparison} should be interpreted as a practical deployment-cost comparison under each method's original execution setting, rather than a hardware-parity comparison of total LLM serving cost.
Under this interpretation, \texttt{SHiFT} has the following practical efficiency properties:

\begin{itemize}[leftmargin=1.5em, itemsep=0.2em]
    \item \textbf{Low Local Memory Footprint.}
    \texttt{SHiFT} consumes only 16.76 GB of local GPU memory, significantly less than LLM-heavy methods such as GraphGPT (60.41 GB) and LLM$_{\text{IT}}$ (69.13 GB).
    This makes \texttt{SHiFT} practical to run on widely available mid-range GPUs (e.g., V100, RTX 3090/4090).

    \item \textbf{Fast Wall-Clock Training.}
    \texttt{SHiFT} completes the full pipeline, including pretraining, taxonomy induction, and EM iterations, in approximately 5.5 hours.
    This is $6.8\times$ faster than TAPE (37.4 h) and $6.6\times$ faster than LLM$_{\text{IT}}$ (36.3 h), validating our key design principle of decoupling heavy LLM reasoning (invoked sparingly) from lightweight GNN training (performed repeatedly).
\end{itemize}

\begin{table*}[t]
\centering
\caption{Efficiency comparison on ArXiv. Baseline results are from \texttt{LLMNodeBed} running on an \textbf{H100 (80 GB)}; \texttt{SHiFT} runs on a \textbf{V100 (32 GB)}. Time is reported in hours (h), minutes (m), or seconds (s).}
\label{tab:efficiency_comparison}
\resizebox{\textwidth}{!}{
\begin{tabular}{lccccccccccc}
\toprule
\textbf{Metric} & \textbf{GCN$_{\text{SE}}$} & \textbf{SenBERT} & \textbf{RoBERTa} & \textbf{ENGINE} & \textbf{GCN$_{\text{LLM}}$} & \textbf{TAPE} & \textbf{GraphGPT} & \textbf{LLaGA} & \textbf{LLM$_{\text{IT}}$} & \textbf{\texttt{SHiFT} (Ours)} \\
\midrule
\textbf{GPU Mem.} (GB) & 4.79 & 9.98 & 46.10 & 7.08 & 7.36 & 46.10 & 60.41 & 41.94 & 69.13 & \textbf{16.76}$^\dagger$ \\
\textbf{Total Time}     & 51s  & 7m   & 41m   & 2.6h & 1.4h & 37.4h & 7.8h & 7.7h  & 36.3h & \textbf{5.5h} \\
\bottomrule
\multicolumn{11}{l}{\footnotesize $^\dagger$ \texttt{SHiFT} runs on a 32 GB V100; methods with $>$40 GB memory require 80 GB A100/H100 GPUs.}
\end{tabular}
}
\end{table*}

\subsection{Time Breakdown and LLM Cost Analysis}
\label{app:llm_cost}

We further decompose the total runtime of \texttt{SHiFT} to quantify the benefit of our Mapping \& Diagnosis strategy.

\vpara{Training Time Composition.}
The total runtime ($\approx$ 5 h 29 m) is distributed as follows:
\begin{itemize}[leftmargin=1.5em, itemsep=0.2em]
    \item \textbf{E-Step (Taxonomy Induction):} The cold-start initialization takes $\approx$ 1 h 19 m. Thanks to the incremental Mapping \& Diagnosis design, subsequent E-steps are progressively cheaper: Update 1 takes $\approx$ 1 h 01 m, and Update 2 takes $\approx$ 59 m.
    \item \textbf{M-Step and Warm-up:} The remaining $\approx$ 2 h 10 m is spent on GCN training and contrastive pretraining, both of which are highly efficient.
\end{itemize}

\vpara{Reduction in LLM API Calls.}
A key concern for LLM-integrated graph learning is inference cost.
Table~\ref{tab:llm_calls} tracks LLM API calls across E-step iterations.
During cold-start initialization, 1,217 calls are required to construct the taxonomy from scratch.
In subsequent updates, the Mapping \& Diagnosis strategy (Section~\ref{sec:taxonomy_updates}) restricts LLM invocation to anomalous clusters only, reducing calls to 285 in Update 1 (\textbf{76.6\% reduction}) and 270 in Update 2.
This confirms that the semantic structure stabilizes rapidly after initialization, incurring minimal marginal LLM cost in later iterations.

\begin{table}[h]
    \centering
    \caption{Number of LLM API calls during E-steps on ArXiv.}
    \label{tab:llm_calls}
    \begin{tabular}{lcc}
    \toprule
    \textbf{Phase} & \textbf{LLM Calls} & \textbf{Reduction vs.\ Init.} \\
    \midrule
    Taxonomy Initialization & 1,217 & -- \\
    Incremental Update 1    &   285 & $\downarrow$ 76.6\% \\
    Incremental Update 2    &   270 & $\downarrow$ 77.8\% \\
    \bottomrule
    \end{tabular}
\end{table}

\section{Impact of LLM Selection}
\label{app:llm_sensitivity}

The core premise of \texttt{SHiFT} is leveraging the semantic reasoning capabilities of LLMs to induce interpretable taxonomies. 
To assess whether our framework relies on specific proprietary models or is robust across different backbones, we evaluate \texttt{SHiFT} using three different LLMs varying in scale and architecture:
(1) \textbf{DeepSeek-V3}~\cite{DeepSeekV3}: The default large-scale model used in our main experiments;
(2) \textbf{Qwen3-32B}~\cite{Qwen3}: A powerful open-weights model with strong reasoning capabilities;
(3) \textbf{Qwen3-8B}~\cite{Qwen3}: A lightweight model suitable for consumer-grade GPUs. 
We access these LLMs via the API provided by Alibaba Cloud. 
We compare these against a \textbf{w/o LLM} baseline, which represents the standard GCN trained on PLM features without any taxonomy-induced constraints.

\begin{table}[h]
    \centering
    \caption{Node classification accuracy (\%) with different LLM backbones.}
    \label{tab:llm_sensitivity}
    \begin{tabular}{l c c c}
    \toprule
    \textbf{LLM Used} & \textbf{Cora} & \textbf{Citeseer} & \textbf{WikiCS} \\
    \midrule
    \textbf{DeepSeek-V3 (Default)} & \textbf{84.83 \scriptsize{±0.58}} & \textbf{73.96 \scriptsize{±0.92}} & \textbf{82.40 \scriptsize{±0.48}} \\
    Qwen3-32B & 82.44 \scriptsize{±0.37} & 73.24 \scriptsize{±1.03} & 82.04 \scriptsize{±0.41} \\
    Qwen3-8B (Lightweight) & 84.14 \scriptsize{±0.60} & 72.97 \scriptsize{±0.27} & 81.65 \scriptsize{±0.18} \\
    \midrule
    \textit{w/o LLM (GCN Backbone)} & 81.70 \scriptsize{±1.01} & 69.71 \scriptsize{±0.64} & 79.95 \scriptsize{±0.15} \\
    \bottomrule
    \end{tabular}
\end{table}

The results are presented in Table~\ref{tab:llm_sensitivity}. We observe three key trends:

\vpara{1. Robustness Across Architectures.}
\texttt{SHiFT} consistently outperforms the \textbf{w/o LLM} baseline regardless of the LLM backend used. 
Even with the smallest model (Qwen3-8B), we observe substantial gains (e.g., $+3.26\%$ on Citeseer compared to baseline). 
This demonstrates that the effectiveness of \texttt{SHiFT} stems from the methodology of aligning embeddings with a semantic hierarchy, rather than solely from the "knowledge injection" of a specific super-model.

\vpara{2. Benefit of Stronger Reasoners.}
DeepSeek-V3 consistently yields the highest accuracy across all datasets (84.83\% on Cora). 
This aligns with intuition: a more capable taxonomist (LLM) induces a cleaner, more logically consistent hierarchy (E-step), which provides higher-quality supervision signals for the GCN (M-step). 
For complex datasets like Citeseer, where community structures are sparse, the stronger reasoning capability of DeepSeek-V3 ($73.96\%$) provides a noticeable edge over Qwen3-32B ($73.24\%$) and 8B ($72.97\%$).

\vpara{3. Efficiency-Performance Trade-off.}
Notably, the lightweight Qwen3-8B delivers surprisingly competitive performance, coming within $\approx 0.7\%$ of DeepSeek-V3 on Cora and WikiCS. 
It even marginally outperforms the larger 32B model on Cora ($84.14\%$ vs $82.44\%$), likely because the taxonomy of Cora is relatively simple, and the 8B model is less prone to over-complicating concept definitions.
This finding is particularly encouraging for practical deployment: it suggests that \texttt{SHiFT} can achieve state-of-the-art results using locally deployable, open-source LLMs, significantly reducing inference costs and latency.

\section{Hyperparameter Sensitivity}\label{app:hyperparameter}

\subsection{Sensitivity to the Level Weighting Factor}\label{app:lambda_sensitivity}

In $\mathcal{L}_{tca}$, the weight $w_l = \lambda^{l}$ controls the relative importance assigned to each level of the hierarchy. We clarify two points here.

\vpara{Motivation for $\lambda > 1$.}
When $\lambda = 1$, all hierarchy levels receive equal weight. This is suboptimal because coarse-grained concepts (e.g., ``Computer Science'') are shared across many nodes and provide limited discriminative signal, while fine-grained concepts (e.g., ``Value Function Approximation'') are far more informative for distinguishing individual nodes. Setting $\lambda > 1$ progressively emphasizes the deeper, more discriminative levels, reflecting this inherent coarse-to-fine structure of the taxonomy.

\vpara{Sensitivity Study.}
To empirically justify our default choice of $\lambda = 1.5$, we conduct a sensitivity 
\begin{wraptable}{r}{0.3\textwidth}
\vspace{-1em}
\centering
\caption{Sensitivity of node classification accuracy (\%) to the level weighting factor $\lambda$.}
\label{tab:lambda_sensitivity}
\resizebox{0.3\textwidth}{!}{
\begin{tabular}{l c c}
\toprule
\textbf{$\lambda$} & \textbf{Cora} & \textbf{Citeseer} \\
\midrule
1.0 (equal) & 84.21 & 73.18 \\
1.25        & 84.52 & 73.67 \\
\textbf{1.5 (default)} & \textbf{84.83} & \textbf{73.96} \\
2.0         & 84.45 & 73.40 \\
\bottomrule
\end{tabular}
}
\vspace{-0.5em}
\end{wraptable}
analysis by varying $\lambda \in \{1.0, 1.25, 1.5, 2.0\}$ on Cora and Citeseer. Results are reported in Table~\ref{tab:lambda_sensitivity}.
Equal weighting ($\lambda = 1$) underperforms because the uniformly distributed loss allows the loose, less discriminative upper-level constraints to dilute the fine-grained alignment signal. Overly aggressive weighting ($\lambda = 2$) also degrades performance, as it disproportionately focuses on leaf-level concepts and neglects the coarse-level regularization that stabilizes the global geometry of the embedding space. $\lambda = 1.5$ strikes an effective balance between these two extremes and is adopted as the default throughout our experiments.

\subsection{Sensitivity to the Taxonomy Depth}\label{app:depth_sensitivity}
We investigate the impact of the maximum taxonomy depth $L_{max}$ on node classification performance.
As shown in Table~\ref{tab:depth_sensitivity}, we observe a consistent trend where performance peaks at Depth 3 across the evaluated datasets (Cora, Citeseer, WikiCS).

We attribute this to a trade-off between semantic granularity and structural noise:
\begin{itemize}[leftmargin=1.5em, itemsep=0.2em]
    \item \textbf{Depth 2 (Too Coarse):} The hierarchy lacks sufficient depth to capture fine-grained semantic distinctions (e.g., specific methods within a sub-field), limiting the discriminative power of the embeddings.
    \item \textbf{Depth 4 (Too Fine):} For these medium-scale academic graphs, Depth 4 often leads to over-fragmentation, where coherent concepts are artificially split into minute sub-clusters, introducing noise into the alignment objectives.
\end{itemize}
Therefore, we set $L_{max}=3$ as the default configuration for these benchmarks.

\vpara{Note on Scalability.}
It is worth noting that the optimal depth is likely correlated with the scale and semantic diversity of the graph. 
While $L_{max}=3$ suffices for the datasets evaluated here (2k-10k nodes), significantly larger or broader datasets (e.g., full-scale \texttt{ogbn-papers100M}~\cite{OGB} with millions of nodes) may inherently possess a deeper semantic structure. 
In such cases, increasing $L_{max}$ (e.g., to 5 or 6) might be necessary to fully disentangle the complex topic hierarchy without over-fragmentation.

\begin{table}[h]
    \centering
    \caption{Impact of maximum taxonomy depth ($L_{max}$) on node classification accuracy (\%).}
    \label{tab:depth_sensitivity}
    \begin{tabular}{l c c c}
    \toprule
    \textbf{Depth} & \textbf{Cora} & \textbf{Citeseer} & \textbf{WikiCS} \\
    \midrule
    2 & 84.40 \scriptsize{±0.30} & 73.57 \scriptsize{±0.63} & 81.63 \scriptsize{±0.17} \\
    \textbf{3 (Default)} & \textbf{84.83 \scriptsize{±0.58}} & \textbf{73.96 \scriptsize{±0.92}} & \textbf{82.40 \scriptsize{±0.48}} \\
    4 & 84.20 \scriptsize{±0.20} & 72.70 \scriptsize{±0.82} & 81.85 \scriptsize{±0.65} \\
    \bottomrule
    \end{tabular}
\end{table}

\section{Visual Analysis of Representations}\label{app:visualization_analysis}

\begin{figure*}[t]
  \centering
  \subfigure[GCN Baseline]{\includegraphics[width=0.48\linewidth]{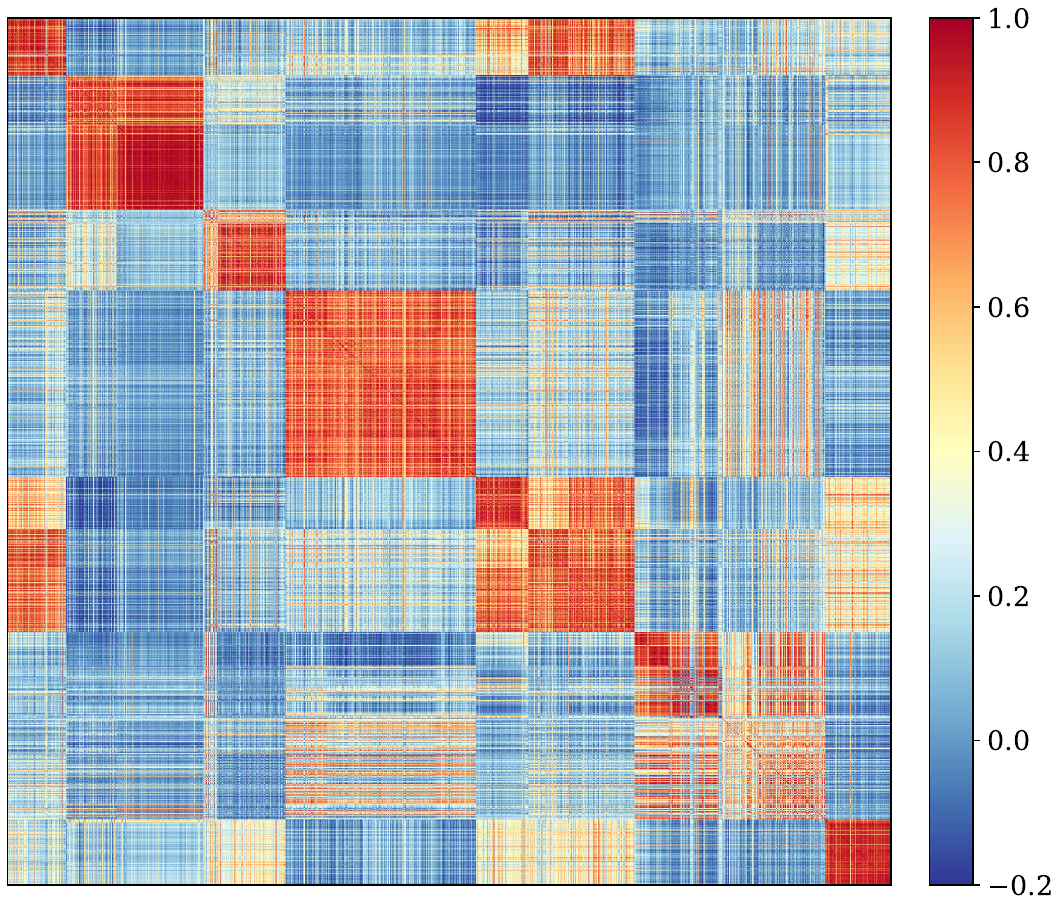}\label{fig:heatmap_gcn}}
  \subfigure[\texttt{SHiFT} (Ours)]{\includegraphics[width=0.48\linewidth]{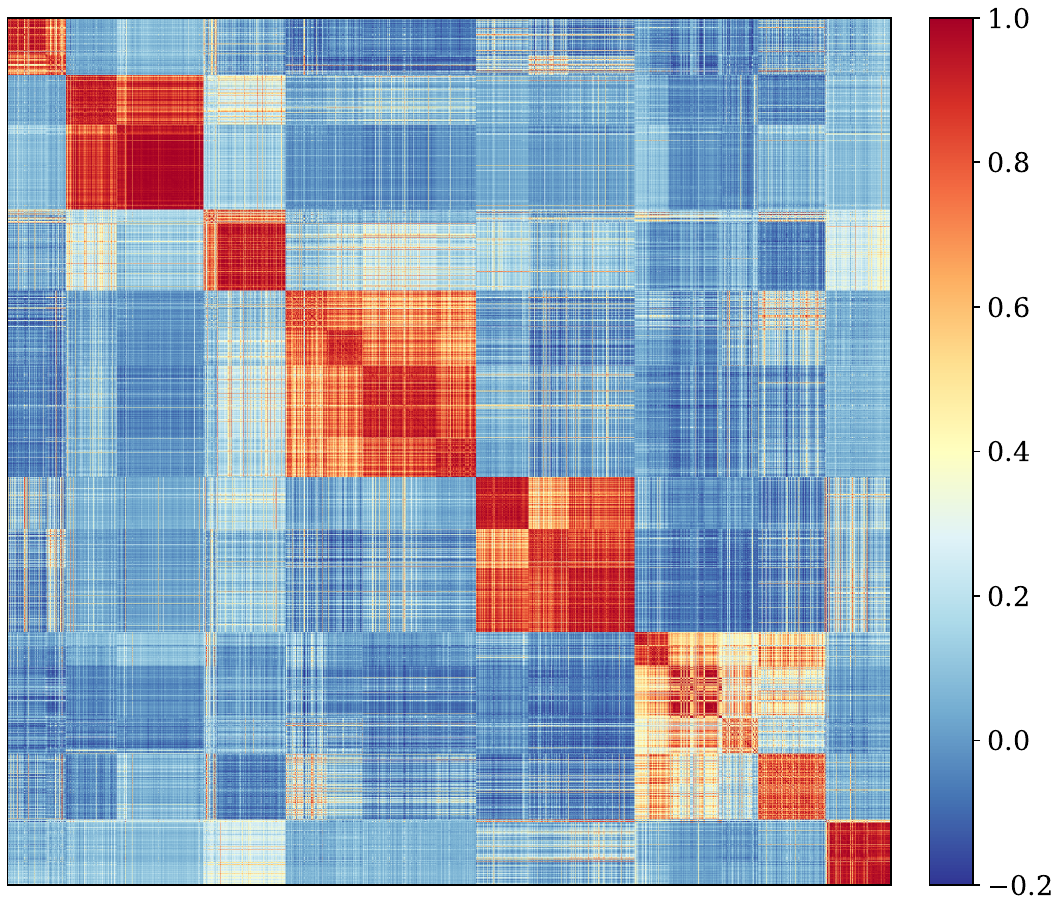}\label{fig:heatmap_ours}}
  \caption{Visualization of the pairwise cosine similarity matrix of node embeddings on the Cora dataset.}
  \label{fig:heatmap_comparison}
\end{figure*}

To further investigate the geometry of the learned embedding space, we plot the heatmap of the pairwise cosine similarity matrix in Figure~\ref{fig:heatmap_comparison}. 
The nodes are ordered based on the Depth-First Search traversal of the taxonomy $\mathcal{T}$ induced by \texttt{SHiFT}. 
This ordering ensures that semantically related nodes (siblings and cousins) are placed adjacently.

\vpara{Baseline Analysis.} 
As observed in Figure~\ref{fig:heatmap_comparison}(a), the baseline GCN produces a diffuse block pattern. 
While it captures some local community structures (indicated by reddish regions), it fails to reveal a clear hierarchical organization. 
The significant noise in the off-diagonal areas (blue regions containing yellow/orange streaks) suggests poor semantic disentanglement, where topologically connected but semantically distinct nodes are pulled too close due to the aggregation mechanism.

\vpara{\texttt{SHiFT} Analysis.} 
In contrast, \texttt{SHiFT} (Figure~\ref{fig:heatmap_comparison}(b)) demonstrates a sharp hierarchical structure. 
We observe a "blocks-within-blocks" pattern:
\begin{itemize}[leftmargin=1.5em, itemsep=0.2em]
    \item \textbf{Fine-grained Cohesion:} The intense red squares along the diagonal correspond to leaf concepts. The sharp boundaries indicate that the Taxonomy Partition Alignment successfully enforces strict local clustering.
    \item \textbf{Coarse-grained Structure:} Crucially, these small squares are grouped into larger, lighter red blocks. This reflects the Taxonomy Skeleton Alignment, which ensures that sibling concepts remain geometrically closer to each other than to concepts in distant branches.
    \item \textbf{Semantic Disentanglement:} The off-diagonal regions are cleanly separated (deep blue), confirming that \texttt{SHiFT} effectively distinguishes between disparate semantic branches (e.g., separating "Neural Networks" from "Probabilistic Methods").
\end{itemize}

\section{Qualitative Analysis of Taxonomies}
\label{app:qualitative_analysis}

To understand how our method uncovers the semantic hierarchy of text-attributed graphs, we conduct a detailed case study on the Cora dataset.
We compare three variations of the induced taxonomy:
(1) \textbf{PLM-only ($\mathcal{T}_{\text{w/o } \mathcal{G}}$):} Induced solely by clustering RoBERTa embeddings without graph structure;
(2) \textbf{Structure-Aware Initialization ($\mathcal{T}_{\text{init}}$):} Induced after the graph contrastive pretraining warm-up;
(3) \textbf{\texttt{SHiFT} Final ($\mathcal{T}_{\text{final}}$):} The final result after iterative EM updates.
The visualizations of these taxonomies are presented in Figure~\ref{fig:taxonomy_plm}, Figure~\ref{fig:taxonomy_init}, and Figure~\ref{fig:taxonomy_final}, respectively.

\subsection{Structural Disentanglement vs. Semantic Redundancy}
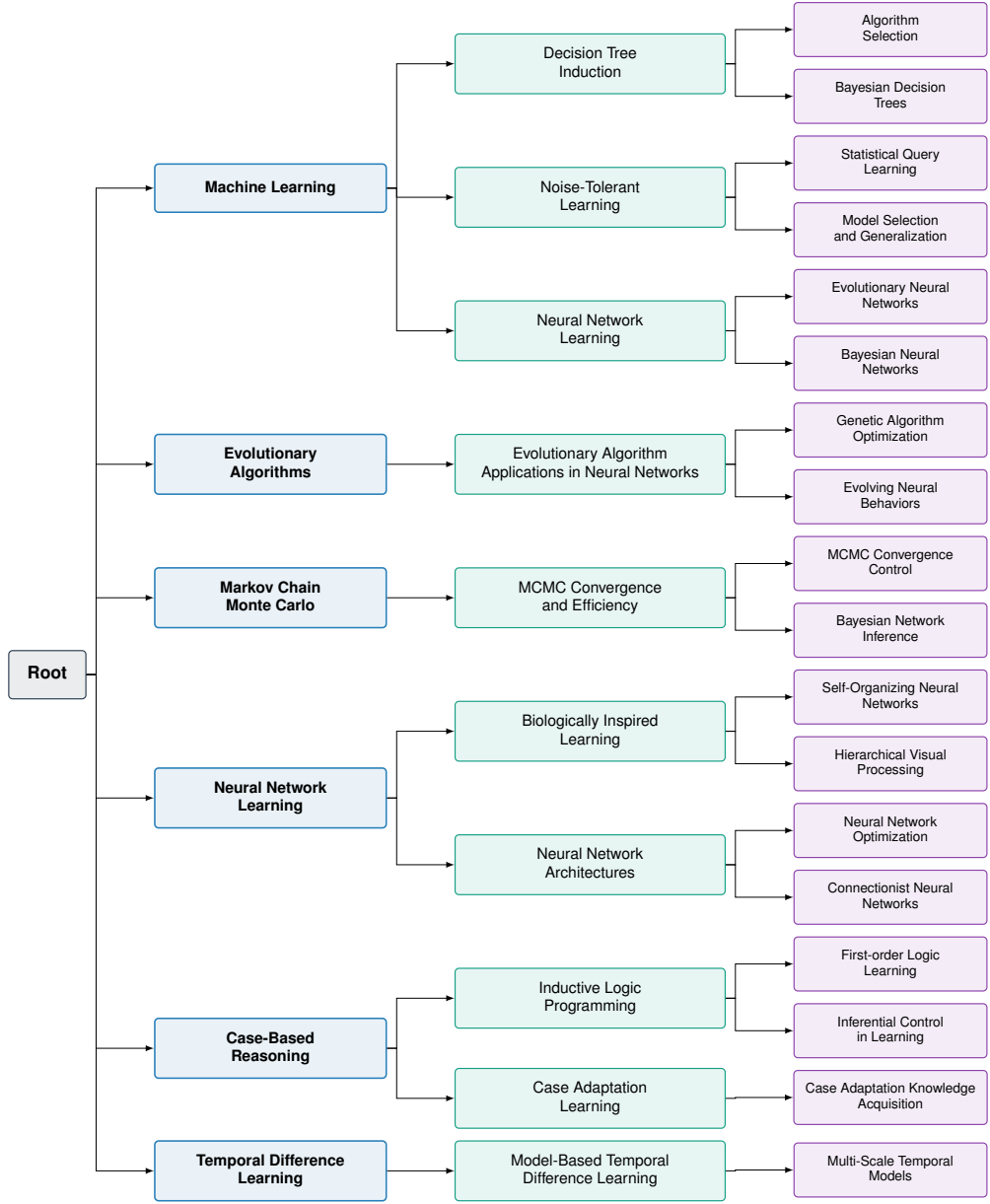
\begin{figure*}[ht]
\centering
\resizebox{0.95\textwidth}{!}{
\begin{forest}
  for tree={
    grow'=0,
    font=\footnotesize\sffamily,
    draw,
    rounded corners=2pt,
    edge={-latex, line width=0.5pt},
    align=center,
    inner sep=6pt,
    l sep=35pt,
    s sep=6pt,
    line width=0.6pt,
    tier/.option=level,
    if level=0{
      fill=level0!10, draw=level0, font=\small\bfseries\sffamily,
      minimum width=40pt, minimum height=25pt
    }{},
    if level=1{
      fill=level1!10, draw=level1, thick, font=\footnotesize\bfseries\sffamily,
      minimum width=120pt, minimum height=25pt
    }{},
    if level=2{
      fill=level2!10, draw=level2,
      minimum width=140pt, minimum height=25pt
    }{},
    if level=3{
      fill=level3!10, draw=level3, font=\scriptsize\sffamily,
      minimum width=100pt, minimum height=25pt
    }{},
  },
  forked edges,
  [Root
    [Machine Learning
      [Decision Tree\\Induction
        [Algorithm\\Selection]
        [Bayesian Decision\\Trees]
      ]
      [Noise-Tolerant\\Learning
        [Statistical Query\\Learning]
        [Model Selection\\and Generalization]
      ]
      [Neural Network\\Learning
        [Evolutionary Neural\\Networks]
        [Bayesian Neural\\Networks]
      ]
    ]
    [Evolutionary\\Algorithms
      [Evolutionary Algorithm\\Applications in Neural Networks
        [Genetic Algorithm\\Optimization]
        [Evolving Neural\\Behaviors]
      ]
    ]
    [Markov Chain\\Monte Carlo
      [MCMC Convergence\\and Efficiency
        [MCMC Convergence\\Control]
        [Bayesian Network\\Inference]
      ]
    ]
    [Neural Network\\Learning
      [Biologically Inspired\\Learning
        [Self-Organizing Neural\\Networks]
        [Hierarchical Visual\\Processing]
      ]
      [Neural Network\\Architectures
        [Neural Network\\Optimization]
        [Connectionist Neural\\Networks]
      ]
    ]
    [Case-Based\\Reasoning
      [Inductive Logic\\Programming
        [First-order Logic\\Learning]
        [Inferential Control\\in Learning]
      ]
      [Case Adaptation\\Learning
        [Case Adaptation Knowledge\\Acquisition]
      ]
    ]
    [Temporal Difference\\Learning
      [Model-Based Temporal\\Difference Learning
        [Multi-Scale Temporal\\Models]
      ]
    ]
  ]
\end{forest}
}
\caption{Cora taxonomy induced solely by PLM embeddings without graph structure ($\mathcal{T}_{\text{w/o } \mathcal{G}}$).}
\label{fig:taxonomy_plm}
\end{figure*}

The most striking difference lies in the global topology.
As shown in Figure~\ref{fig:taxonomy_plm}, the PLM-only baseline fails to distinguish citation communities from lexical topics.
It generates a generic "Machine Learning" child node under the root, which is uninformative for a dataset entirely composed of ML papers.
More critically, it exhibits \textit{structural hallucination}: "Neural Network Learning" appears twice, once as a child of "Machine Learning" and once as an independent root child.
This confirms that without topological guidance, pure semantic similarity is insufficient to separate distinct research sub-fields (e.g., distinguishing "Rule Learning" papers from "Neural Network" papers, which use similar vocabulary but rarely cite each other).

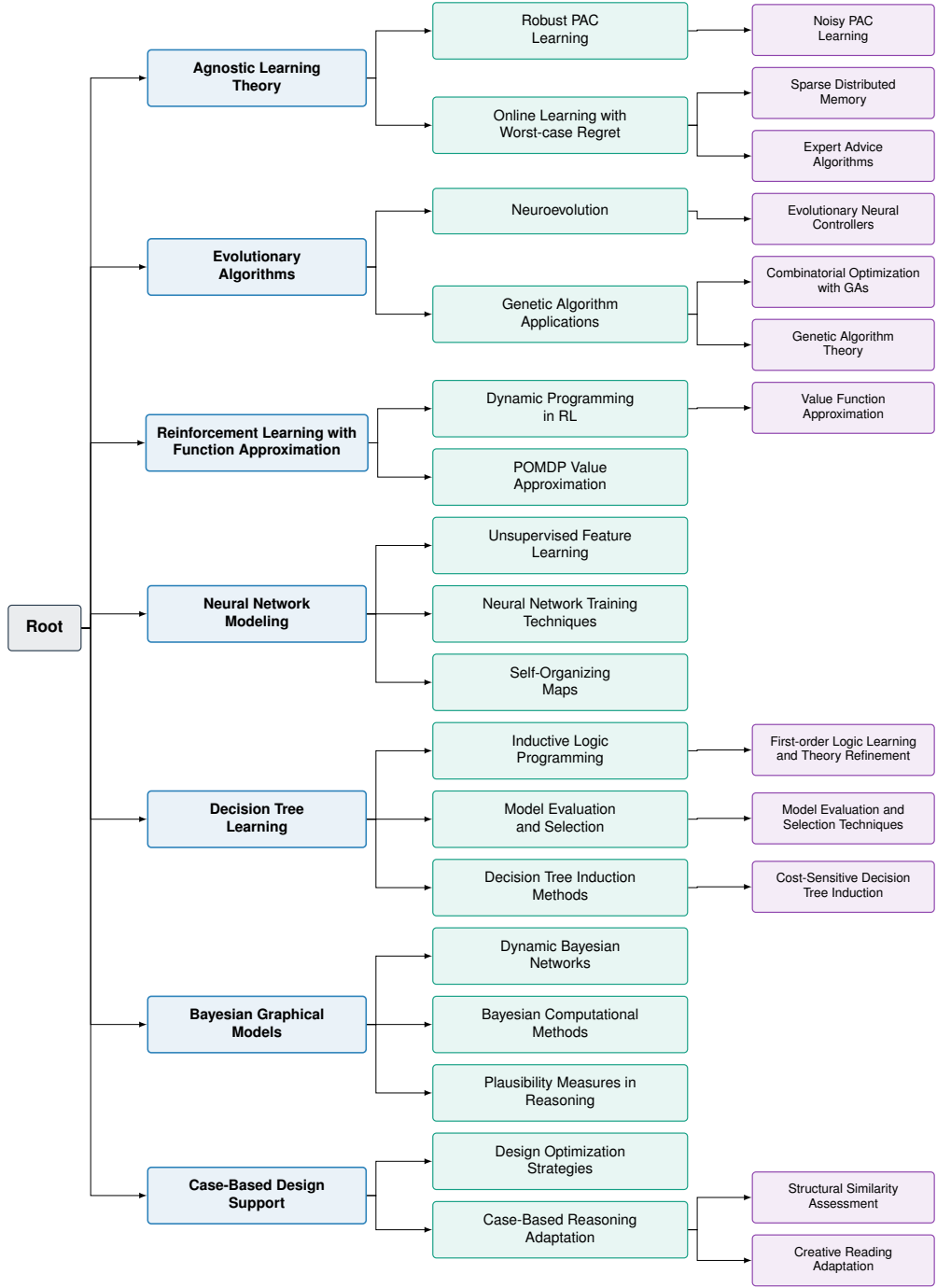
\begin{figure*}[ht]
\centering
\resizebox{0.95\textwidth}{!}{
\begin{forest}
  for tree={
    grow'=0,
    font=\footnotesize\sffamily,
    draw,
    rounded corners=2pt,
    edge={-latex, line width=0.5pt},
    align=center,
    inner sep=6pt,
    l sep=35pt,
    s sep=6pt,
    line width=0.6pt,
    tier/.option=level,
    if level=0{
      fill=level0!10, draw=level0, font=\small\bfseries\sffamily,
      minimum width=40pt, minimum height=25pt
    }{},
    if level=1{
      fill=level1!10, draw=level1, thick, font=\footnotesize\bfseries\sffamily,
      minimum width=120pt, minimum height=25pt
    }{},
    if level=2{
      fill=level2!10, draw=level2,
      minimum width=140pt, minimum height=25pt
    }{},
    if level=3{
      fill=level3!10, draw=level3, font=\scriptsize\sffamily,
      minimum width=100pt, minimum height=25pt
    }{},
  },
  forked edges,
  [Root
    [Agnostic Learning\\Theory
      [Robust PAC\\Learning
        [Noisy PAC\\Learning]
      ]
      [Online Learning with\\Worst-case Regret
        [Sparse Distributed\\Memory]
        [Expert Advice\\Algorithms]
      ]
    ]
    [Evolutionary\\Algorithms
      [Neuroevolution
        [Evolutionary Neural\\Controllers]
      ]
      [Genetic Algorithm\\Applications
        [Combinatorial Optimization\\with GAs]
        [Genetic Algorithm\\Theory]
      ]
    ]
    [Reinforcement Learning with\\Function Approximation
      [Dynamic Programming\\in RL
        [Value Function\\Approximation]
      ]
      [POMDP Value\\Approximation]
    ]
    [Neural Network\\Modeling
      [Unsupervised Feature\\Learning]
      [Neural Network Training\\Techniques]
      [Self-Organizing\\Maps]
    ]
    [Decision Tree\\Learning
      [Inductive Logic\\Programming
        [First-order Logic Learning\\and Theory Refinement]
      ]
      [Model Evaluation\\and Selection
        [Model Evaluation and\\Selection Techniques]
      ]
      [Decision Tree Induction\\Methods
        [Cost-Sensitive Decision\\Tree Induction]
      ]
    ]
    [Bayesian Graphical\\Models
      [Dynamic Bayesian\\Networks]
      [Bayesian Computational\\Methods]
      [Plausibility Measures in\\Reasoning]
    ]
    [Case-Based Design\\Support
      [Design Optimization\\Strategies]
      [Case-Based Reasoning\\Adaptation
        [Structural Similarity\\Assessment]
        [Creative Reading\\Adaptation]
      ]
    ]
  ]
\end{forest}
}
\caption{Cora taxonomy initialized after Graph Contrastive Pretraining ($\mathcal{T}_{\text{init}}$).}
\label{fig:taxonomy_init}
\end{figure*}

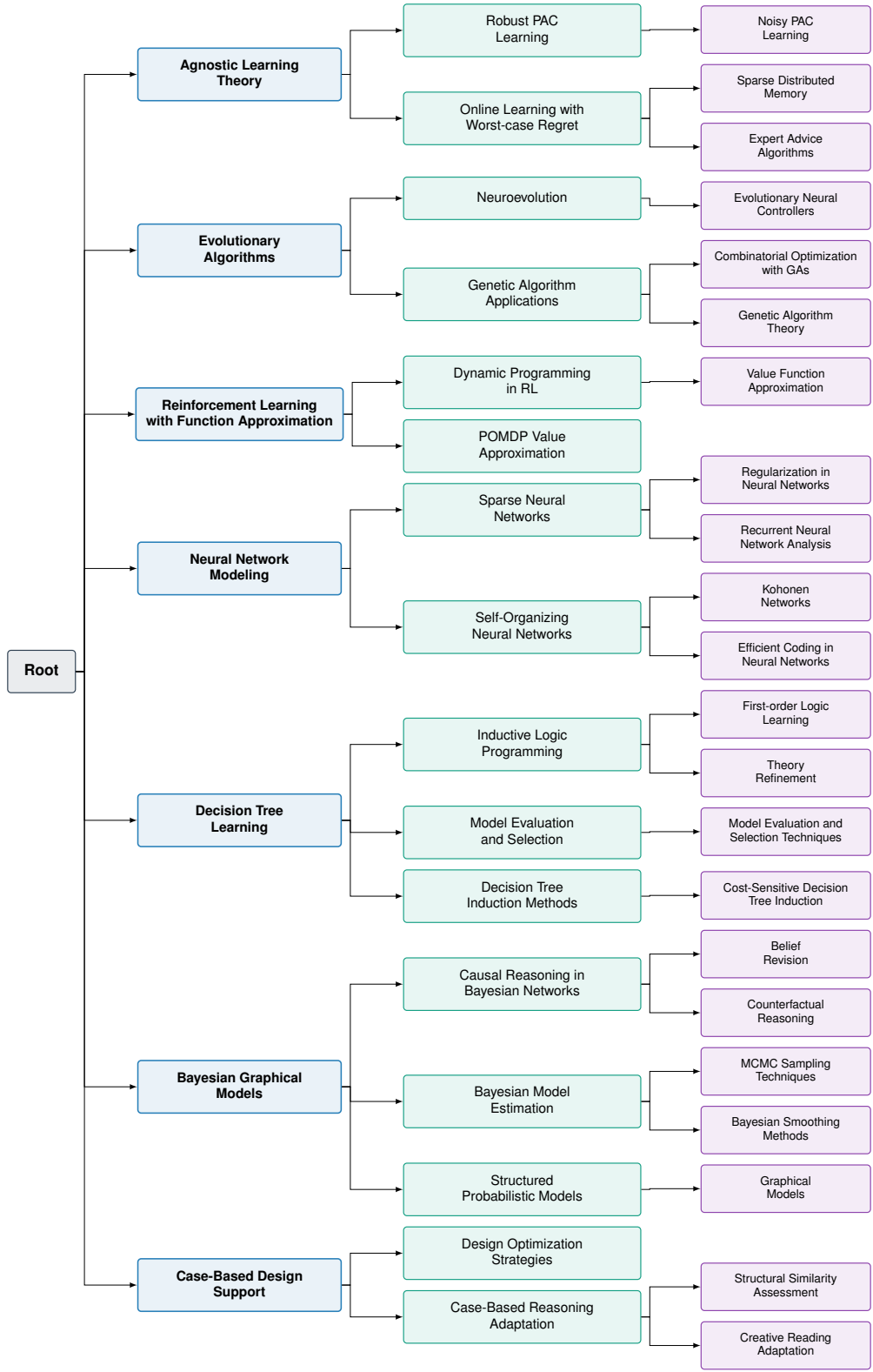
\begin{figure*}[ht]
\centering
\resizebox{0.95\textwidth}{!}{
\begin{forest}
  for tree={
    grow'=0,
    font=\footnotesize\sffamily,
    draw,
    rounded corners=2pt,
    edge={-latex, line width=0.5pt},
    align=center,
    inner sep=6pt,
    l sep=35pt,
    s sep=6pt,
    line width=0.6pt,
    tier/.option=level,
    if level=0{
      fill=level0!10, draw=level0, font=\small\bfseries\sffamily,
      minimum width=40pt, minimum height=25pt
    }{},
    if level=1{
      fill=level1!10, draw=level1, thick, font=\footnotesize\bfseries\sffamily,
      minimum width=120pt, minimum height=25pt
    }{},
    if level=2{
      fill=level2!10, draw=level2,
      minimum width=140pt, minimum height=25pt
    }{},
    if level=3{
      fill=level3!10, draw=level3, font=\scriptsize\sffamily,
      minimum width=100pt, minimum height=25pt
    }{},
  },
  forked edges,
  [Root
    [Agnostic Learning\\Theory
      [Robust PAC\\Learning
        [Noisy PAC\\Learning]
      ]
      [Online Learning with\\Worst-case Regret
        [Sparse Distributed\\Memory]
        [Expert Advice\\Algorithms]
      ]
    ]
    [Evolutionary\\Algorithms
      [Neuroevolution
        [Evolutionary Neural\\Controllers]
      ]
      [Genetic Algorithm\\Applications
        [Combinatorial Optimization\\with GAs]
        [Genetic Algorithm\\Theory]
      ]
    ]
    [Reinforcement Learning\\with Function Approximation
      [Dynamic Programming\\in RL
        [Value Function\\Approximation]
      ]
      [POMDP Value\\Approximation]
    ]
    [Neural Network\\Modeling
      [Sparse Neural\\Networks
        [Regularization in\\Neural Networks]
        [Recurrent Neural\\Network Analysis]
      ]
      [Self-Organizing\\Neural Networks
        [Kohonen\\Networks]
        [Efficient Coding in\\Neural Networks]
      ]
    ]
    [Decision Tree\\Learning
      [Inductive Logic\\Programming
        [First-order Logic\\Learning]
        [Theory\\Refinement]
      ]
      [Model Evaluation\\and Selection
        [Model Evaluation and\\Selection Techniques]
      ]
      [Decision Tree\\Induction Methods
        [Cost-Sensitive Decision\\Tree Induction]
      ]
    ]
    [Bayesian Graphical\\Models
      [Causal Reasoning in\\Bayesian Networks
        [Belief\\Revision]
        [Counterfactual\\Reasoning]
      ]
      [Bayesian Model\\Estimation
        [MCMC Sampling\\Techniques]
        [Bayesian Smoothing\\Methods]
      ]
      [Structured\\Probabilistic Models
        [Graphical\\Models]
      ]
    ]
    [Case-Based Design\\Support
      [Design Optimization\\Strategies]
      [Case-Based Reasoning\\Adaptation
        [Structural Similarity\\Assessment]
        [Creative Reading\\Adaptation]
      ]
    ]
  ]
\end{forest}
}
\caption{Final Cora taxonomy induced by \texttt{SHiFT} ($\mathcal{T}_{\text{final}}$).}
\label{fig:taxonomy_final}
\end{figure*}

In contrast, both $\mathcal{T}_{\text{init}}$ (Figure~\ref{fig:taxonomy_init}) and $\mathcal{T}_{\text{final}}$ (Figure~\ref{fig:taxonomy_final}) utilize the graph structure to pull the embedding space into topologically coherent regions.
\texttt{SHiFT} successfully identifies exactly 7 top-level concepts, which map near-perfectly to the ground-truth classes of Cora, as detailed in Table~\ref{tab:mapping}.
For instance, the vague "Machine Learning" cluster is replaced by specific communities like "Decision Tree Learning" (Rule Learning) and "Neural Network Modeling", effectively resolving the redundancy issue.

\begin{table}[h]
\centering
\caption{Mapping between Ground-Truth classes and \texttt{SHiFT}-induced concepts.}
\label{tab:mapping}
\begin{tabular}{l l}
\toprule
\textbf{Cora Ground Truth} & \textbf{\texttt{SHiFT} Induced Concept ($\mathcal{T}_{\text{final}}$)} \\
\midrule
Neural Networks & Neural Network Modeling \\
Rule Learning & Decision Tree Learning \\
Reinforcement Learning & RL with Function Approximation \\
Probabilistic Methods & Bayesian Graphical Models \\
Theory & Agnostic Learning Theory \\
Genetic Algorithms & Evolutionary Algorithms \\
Case Based & Case-Based Design Support \\
\bottomrule
\end{tabular}
\end{table}

\subsection{Hierarchy Depth and Concept Refinement}
While the initialization step ($\mathcal{T}_{\text{init}}$) captures the skeleton, the iterative EM process ($\mathcal{T}_{\text{final}}$) significantly refines the semantic granularity and corrects logical inconsistencies.

\vpara{Correction of Concept Drift.}
In the PLM-only tree, "Markov Chain Monte Carlo" is placed at Level 1, parallel to broad fields like "Evolutionary Algorithms".
\texttt{SHiFT} corrects this by recognizing MCMC as a methodological tool.
In $\mathcal{T}_{\text{final}}$, MCMC is correctly placed as a Level 3 sub-concept under "Bayesian Model Estimation", which itself sits under the "Bayesian Graphical Models" umbrella.
This restores the logical \textit{Field $\rightarrow$ Method $\rightarrow$ Technique} hierarchy.

\vpara{Coarse-to-Fine Granularity.}
The final taxonomy demonstrates a sophisticated understanding of sub-fields.
Consider the "Reinforcement Learning" branch:
\begin{itemize}[leftmargin=1.5em, itemsep=0.2em]
    \item \textbf{Level 1 (Field):} \textit{Reinforcement Learning with Function Approximation}, which correctly identifies the specific flavor of RL dominant in the Cora dataset.
    \item \textbf{Level 2 (Methodology):} \textit{Dynamic Programming in RL}, which distinguishes model-based approaches from others.
    \item \textbf{Level 3 (Technique):} \textit{Value Function Approximation}, a specific optimization technique within DP.
\end{itemize}
This deep hierarchy contrasts sharply with flat clustering methods, providing an interpretable roadmap of the scientific domain.

\vpara{Semantic Precision.}
The LLM's role in definition generation shines in the "Theory" class.
While the ground truth label is simply "Theory", \texttt{SHiFT} generates the concept "Agnostic Learning Theory", with a definition emphasizing "minimal assumptions about target functions".
This accurately reflects the \textit{Computational Learning Theory} (COLT) literature (e.g., PAC learning) prevalent in Cora, offering higher semantic density than human-annotated labels.

\clearpage
\section{Miscellaneous}\label{app:misc}

\subsection{Limitations}\label{app:limitation}

Despite the promising results achieved by our framework in uncovering semantic hierarchies for text-attributed graphs, we acknowledge two primary limitations that define the scope of our current work and point toward future research directions.

\vpara{Resource Consumption and Latency.}
First, the integration of LLMs introduces non-negligible computational overhead compared to traditional GNNs.
Although our Mapping \& Diagnosis strategy significantly reduces the frequency of LLM invocation, and our sensitivity analysis with Qwen3-8B (Appendix~\ref{app:llm_sensitivity}) demonstrates that smaller, locally deployable models can yield competitive performance without relying on expensive proprietary APIs, the fundamental inference latency of LLMs remains a bottleneck.
On extremely large-scale graphs with millions of nodes or in latency-sensitive applications, the cost and time required for the LLM to recursively cluster and define concepts may be prohibitive.
Future work could explore distilling the taxonomic reasoning capabilities of large models into smaller, locally deployable student models to mitigate this bottleneck.

\vpara{Scope of Hierarchical Evaluation.}
Second, most public TAG benchmarks still provide only flat labels.
On Cora, Citeseer, ArXiv, WikiCS, and Computer we therefore rely on NPMI, human judgments, and downstream accuracy as proxies.
Appendix~\ref{app:patent} adds one direct test: a patent graph with a 3-level CPC taxonomy.
Even there, \texttt{SHiFT} is not designed to reproduce the USPTO symbol inventory verbatim.
It induces data-driven concepts, so CPC recall remains partial ($0.64$ for $\mathcal{T}_{\text{final}}$).
A multi-domain suite of hierarchy-annotated TAGs is still needed before one can claim that explicit hierarchy modeling is necessary for every benchmark.

\subsection{Ethical Considerations}\label{app:ethical}
To the best of our knowledge, the research presented in this paper conforms with the NeurIPS Code of Ethics. The work uses standard public benchmark datasets and a patent graph built from public USPTO records. It does not involve the collection of new personal data. It includes a small-scale, low-risk human evaluation for assessing taxonomy quality; this evaluation did not involve sensitive personal data and was conducted following the authors’ institutional norms for such assessments. The proposed method is intended as a general-purpose research contribution and not for harmful applications such as surveillance, deception, or weaponization. We are not aware of any violations regarding privacy, consent, copyright, or responsible release, while acknowledging that pretrained language models and benchmark datasets may inherit existing biases that should be considered in downstream use.

\subsection{Broader Impacts}\label{app:broader_impacts}
This work is a foundational method for hierarchy-aware representation learning on text-attributed graphs. 
On the positive side, inducing explicit and human-readable taxonomies may improve the interpretability of graph learning systems and support applications such as scientific literature organization, knowledge discovery, and recommendation. 

Potential negative impacts also exist. 
Because the induced hierarchy depends on input texts, graph structure, and LLM judgments, the method may inherit or amplify existing biases, leading to misleading or unfair semantic groupings. 
Moreover, when applied to sensitive relational data, explicit hierarchy induction may expose latent group structure and raise privacy concerns. 

We therefore recommend human oversight, bias auditing of induced concepts, and cautious use on sensitive or high-stakes applications. 
The induced taxonomy should be treated as an assistive abstraction rather than ground truth.

\subsection{Declaration of LLM Usage}\label{app:llm_declaration}
Large Language Models (LLMs) are an essential component of our method. 
Specifically, in the E-step of our EM framework, we use an external LLM as a semantic reasoner to induce and refine the hierarchical taxonomy. 
The LLM is prompted to (i) generate concept names and definitions from representative document sets, (ii) decide whether clusters require further splitting, and (iii) perform editing operations such as merging, pruning, and redefining concepts during incremental updates.

Importantly, the LLM is not used to generate node features, labels, or evaluation results. 
Instead, it serves as a structured semantic oracle whose outputs are explicitly integrated into the learning process via taxonomy construction and alignment objectives. 
All prompts, usage details, and implementation settings are provided in Appendix~\ref{app:prompts} to ensure transparency and reproducibility.

\end{document}